\documentclass[a4paper,12pt,times,numbered,print,index, custom, authoryear]{Classes/PhDThesisPSnPDF}

\input{Preamble/preamble}

\title{Speeding Up MACE: Low-Precision Tricks for Equivarient Force Fields}

\author{Alexandre Benoit}

\dept{Department of Engineering}

\university{University of Cambridge}
\crest{\includegraphics[width=0.2\textwidth]{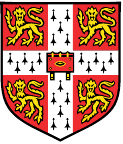}}

\degreetitle{Master of Philosophy in Machine Learning and Machine Intelligence}

\college{Clare College}

\subject{LaTeX} \keywords{{LaTeX} {PhD Thesis} {Engineering} {University of
Cambridge}}

\ifdefineAbstract
\fi

\ifdefineChapter
\fi

\begin{document}

\frontmatter

\maketitle


\begin{dedication} 

\textit{Cheers to the things that should make life worth living, but only serve to remind me how much harder it is to exist.}

\end{dedication}

\begin{declaration}

I, Alexandre Benoit of Clare College, candidate for the MPhil in Machine Learning and Machine Intelligence, hereby declare that this dissertation are original and have not been  submitted in whole or in part for consideration for any other degree or 
qualification in this, or any other university. This dissertation is my own 
work and contains nothing which is the outcome of work done in collaboration 
with others, except as specified in the text and here, in the declaration.

This dissertation contains around 12,500 words including appendices, bibliography, footnotes, tables and equations. The code related to this work is available on GitHub and can be accessed via the reference \citep{alexandrebenoitcode}.

This work builds on the ACE/MACE suite \citep{Batatia2022Design}, \citep{Batatia2022mace}, which I have adapted to investigate low-precision techniques within the MACE architecture.

I also acknowledge that limitations in access to the High Performance Computing (HPC) service significantly affected the results and workflow of this thesis. Frequent interruptions and extended queue times for server allocation resulted in considerable delays, challenges and frustration. To mitigate these disruptions, I relied on CBL GPU machines (NVIDIA GPU RTX 2080 Ti ) with lower specifications than the HPC system (NVIDIA GPU A100 80GB SXM), which, while enabling progress, inevitably introduced compromises in performance and results. Experiments were executed on an RTX 2080 Ti; however, the methodology and resulting outcomes are hardware-agnostic and fully reproducible on contemporary GPUs, including the A100 and H200. Thus, the findings contribute to the current body of literature without being rendered obsolete by the use of an earlier-generation graphics card.


\end{declaration}
\begin{abstract}
Machine‐learning force fields can deliver accurate molecular dynamics (MD) at high computational cost. For SO(3)-equivariant models such as MACE, there is little systematic evidence on whether reduced-precision arithmetic and GPU-optimized kernels can cut this cost without harming physical fidelity.

This thesis aims to make MACE cheaper and faster while preserving accuracy by identifying computational bottlenecks and evaluating low-precision execution policies.

We profile MACE end-to-end and per block, compare the e3nn and NVIDIA cuEquivariance backends, and assess FP64/FP32/BF16/FP16 settings (with FP32 accumulation) for inference, short NVT and a long NPT water simulations, and toy training runs under reproducible, steady-state timing.

cuEquivariance reduces inference latency by about $3\times$. Casting only linear layers to BF16/FP16 within an FP32 model yields roughly $4\times$ additional speedups, while energies and thermodynamic observables in NVT/NPT MD remain within run-to-run variability. Half-precision weights during training degrade force RMSE. Mixing e3nn and cuEq modules without explicit adapters causes representation mismatches.

Fused equivariant kernels and mixed-precision \emph{inference} can substantially accelerate state-of-the-art force fields with negligible impact on downstream MD. A practical policy is to use cuEquivariance with FP32 by default and enable BF16/FP16 for linear layers (keeping FP32 accumulations) for maximum throughput, while training remains in FP32. Further gains are expected on Ampere/Hopper GPUs (TF32/BF16) and from kernel-level FP16/BF16 paths and pipeline fusion.
\end{abstract}



\tableofcontents

\listoffigures

\listoftables


\printnomenclature

\mainmatter


\chapter{Introduction}  

\ifpdf
    \graphicspath{{Chapter1/Figs/Raster/}{Chapter1/Figs/PDF/}{Chapter1/Figs/}}
\else
    \graphicspath{{Chapter1/Figs/Vector/}{Chapter1/Figs/}}
\fi

\textit{This chapter provides a general overview of the broader context in which this thesis is situated, beginning with the big picture and overall research goals in accessible terms. It outlines the motivation, approach, and main contributions of the work, while also situating the thesis within its strategic problem framework. The chapter concludes with a roadmap of the thesis, describing its organisation and structure, and highlighting any published papers by the author on which parts of the thesis are based.}

\section{Overview}

Modern machine learning models have achieved remarkable success across domains, but this progress often comes with steep computational costs. Training and deploying large models like advanced language models can require enormous computing power and energy, raising concerns about economic and environmental sustainability \citep{Legrand_2025}. A recent example is DeepSeek, a ChatGPT-like AI that matched GPT-4’s performance at a fraction of the cost – roughly \$5.6M vs \$3B+ – while using far fewer hardware resources and about 90\% less energy. This paradigm shift toward efficiency over brute-force computation highlights a broader trend in AI: the race to make models cheaper and greener without sacrificing capability and accuracy where environmental impact can be drastically reduced as mentioned in the example above where it achieved 90\% less energy consumption and 92\% lower carbon footprint.

A similar challenge exists in the field of molecular dynamics (MD) simulations. Predicting interatomic forces with high fidelity is computationally intensive because of large and iterative nature making training and inference computationally expensive to calculating energies and forces \citep{bukharin2023machinelearningforcefields}, which has motivated the development of machine learning force fields. MACE (Higher Order Equivariant Message Passing Neural Networks for Fast and Accurate Force Fields) \citep{Batatia2022mace} is a state-of-the-art equivariant neural network for atomic force prediction that achieves excellent accuracy on benchmark systems. MACE addresses some limitations of earlier models – many message-passing neural networks (MPNNs) suffered from high computational cost and poor scalability – by using higher-order equivariant interactions (four-body message passing) to improve efficiency. In particular, this design makes MACE a fast and highly parallelizable model while reaching or exceeding the accuracy of previous approaches. However, even with such innovations, training and running MACE for long MD trajectories or large systems can still demand significant time and energy. Reducing the computational expense of these models is key to enabling affordable large-scale simulations.

Strategies to solve this problem spans global architectural improvements, fine-grained profiling and bottleneck elimination, mathematical optimizations of equivariant operations, hardware-specific enhancements (particularly on NVIDIA A100 GPUs), and broader scaling techniques. One promising direction for accelerating MACE --- particularly its core architectural and mathematical components --- is to exploit lower numerical precision during computations while leveraging the state-of-the-art capabilities of modern GPUs such as NVIDIA’s A100 or H200. In modern deep learning, mixed-precision arithmetic (using 16-bit or improved 32-bit floats) has proven to greatly accelerate training and inference on GPUs. This approach yields multiple benefits – smaller memory footprint, faster throughput, and lower energy usage – while typically maintaining model accuracy within acceptable bounds \citep{NVIDIA_Developer}. For example, hardware optimized for half-precision can deliver higher FLOP/s and reduce power consumption, making training more cost-effective and sustainable.

In this thesis, we apply mixed-precision techniques to the MACE force field model. We investigate how low-precision tricks (such as 16-bit floating point operations or dynamic numerical precision scaling) can accelerate MACE’s training and inference, and what impact these have on the model’s predictive precision. By doing so, we aim to enable molecular dynamics simulations that retain MACE’s state-of-the-art accuracy while running significantly faster and more efficiently.

We will try to answer the following research questions:
\begin{itemize}
    \item \textbf{RQ 1 - Blockwise precision limits}: For each MACE block, what is the lowest arithmetic precision (FP64/FP32/BF16/FP16, with/without FP32 accumulation) that preserves each mathematical tensor outputs numerical accuracy?
    \item \textbf{RQ 2 - End-to-end speed/memory/energy trade-offs}: What end-to-end gains do mixed-precision configurations deliver on NVIDIA powerful GPUs for inference?
    \item \textbf{RQ 3 - Kernel's effect on E3NN and cuEquivariance}: How do different kernel backends interact with precision (FP32/BF16/FP16) for the dominant MACE's blocks, and what is the accuracy delta between e3nn and cuEquivariance across batch sizes and systems sizes?
    \item \textbf{RQ 4 - Long-horizon MD stability shaping the numerical policy design}:  With low-precision inference, how do long MD trajectories compare to high-precision baselines in terms of energy drift (NVE), thermostat bias (NVT), and structural observables (e.g., RDF, MSD) and how does it guides us for choosing the perfect policies taking into account trade-off between stability and speed?
    \item \textbf{RQ5 - Sensitivity to low-precision training noise.} How sensitive is MACE to rounding errors introduced by FP16/BF16 (or mixed-precision) training, and to what extent can the model adapt to - or compensate for the noise propagated by low-precision arithmetic?
\end{itemize}

\section{Research gaps and Contributions} 

    Although numerous acceleration techniques have been proposed for MACE and related ML force fields, evidence is fragmented and often tied to narrow experimental settings. Existing studies typically report accuracy benchmarks or isolated micro-optimizations, but they rarely provide a systematic, end-to-end view of \emph{where} MACE spends time and memory, \emph{why}, and \emph{how} those costs trade off against predictive fidelity. In particular, there is no comprehensive profiling of MACE’s execution pipeline that quantifies dominant bottlenecks and evaluates low-precision regimes (fp64/fp32/bf16/fp16) under controlled conditions. As a result, adopting any single published trick is uncertain outside of the authors’ setup, and it is unclear which combinations actually deliver reliable speedups for training and inference without degrading physical metrics (energies, forces, MD stability).

The major contributions of this thesis are the following:
\begin{enumerate}
    \item \textbf{Profiling \& cost model:} End-to-end profiling of MACE and a compact FLOP/memory model highlighting MACE's block computational expensiveness.
    \item \textbf{Mixed-precision mapping:} Block-wise mixed precision study of numerical precision speedup and manufacturing a “when to downcast” rules on specific sub-blocks of MACE between cuEquivariance and E3NN.
    \item \textbf{MD case study:} Water-box experiments assessing density and energy-conservation impacts under reduced precision.
    \item \textbf{Low-precision training:} Empirical analysis of FP16/BF16 training and its effect on convergence and $(\mathrm{RMSE}_E,\mathrm{RMSE}_F)$.
\end{enumerate}

\section{Outlines} 

This thesis is organised as follows:
\begin{itemize}
\item Chapter~\ref{chapter:background} reviews the MACE architecture, its core mathematical components, and recent advances in SO(3)-equivariant methods. It further discusses state-of-the-art approaches and methods for evaluating and accelerating machine learning architectures, alongside an overview of numerical precision theory and its impact on tensor computations and results.
\item Chapter~\ref{chapter:methodology} details the experimental setup, including the hardware environment, datasets, and procedures adopted to carry out the benchmarking and evaluations.
\item Chapter~\ref{chapter:results_and_discussion} presents the benchmarking results across different MACE blocks, examines inference under reduced numerical precision, and analyses the behaviour of molecular dynamics simulations under varying precision levels.
\item Chapter~\ref{chapter:conclusion} presents the main takeaways of this thesis, reflecting on the key findings and their broader implications. It also outlines promising directions for future research and highlights potential avenues for further improving and extending this study.  
\end{itemize}


\chapter{Background}
\label{chapter:background}

\ifpdf
    \graphicspath{{Chapter2/Figs/Raster/}{Chapter2/Figs/PDF/}{Chapter2/Figs/}}
\else
    \graphicspath{{Chapter2/Figs/Vector/}{Chapter2/Figs/}}
\fi

\textit{This background surveys acceleration strategies for MACE and related ML force fields with a performance-oriented scope. We synthesize methods across: global architectural changes (Section~\ref{sec:pmo}) where we introduce MACE and motivate its suitability as a target force field; fine-grained profiling and bottleneck removal (Section~\ref{sec:clpb}); numerical precision methods for speed optimization (Section~\ref{sec:paa}); mathematical optimizations of equivariant operations (Section~\ref{sec:se}); hardware-aware implementations (Section~\ref{sec:gco}); several model compression applied to force field models (Section~\ref{sec:mc}) and broader scaling techniques (Section~\ref{sec:cmhas}-~\ref{sec:bp}). 
We focus on methods that increase training/inference throughput and reduce latency while preserving fidelity in energies, forces, and MD stability. This is not a primer on atomistic simulation or a history of equivariance; mathematical details are introduced only when they inform performance-critical design in MACE or guide our comprehension on expensiveness of the model's calculations. Our aim is to motivate the methodological choices developed later and to define clear criteria for efficiency-accuracy trade-offs. Because few works target low-precision specifically for this architecture (actually nearly 0 \citep{Zhou_2025}), we adopt a systems perspective, synthesizing developer-led techniques and toolchain reports that enable mixed/low-precision execution in practice. The emphasis is on actionable, hardware-aware strategies rather than a broad census, providing a concise roadmap and open directions for future students.}

\section{Problem and MACE overview}
\label{sec:pmo}

   \begin{figure}[H]
        \centering
        \includegraphics[width=\linewidth]{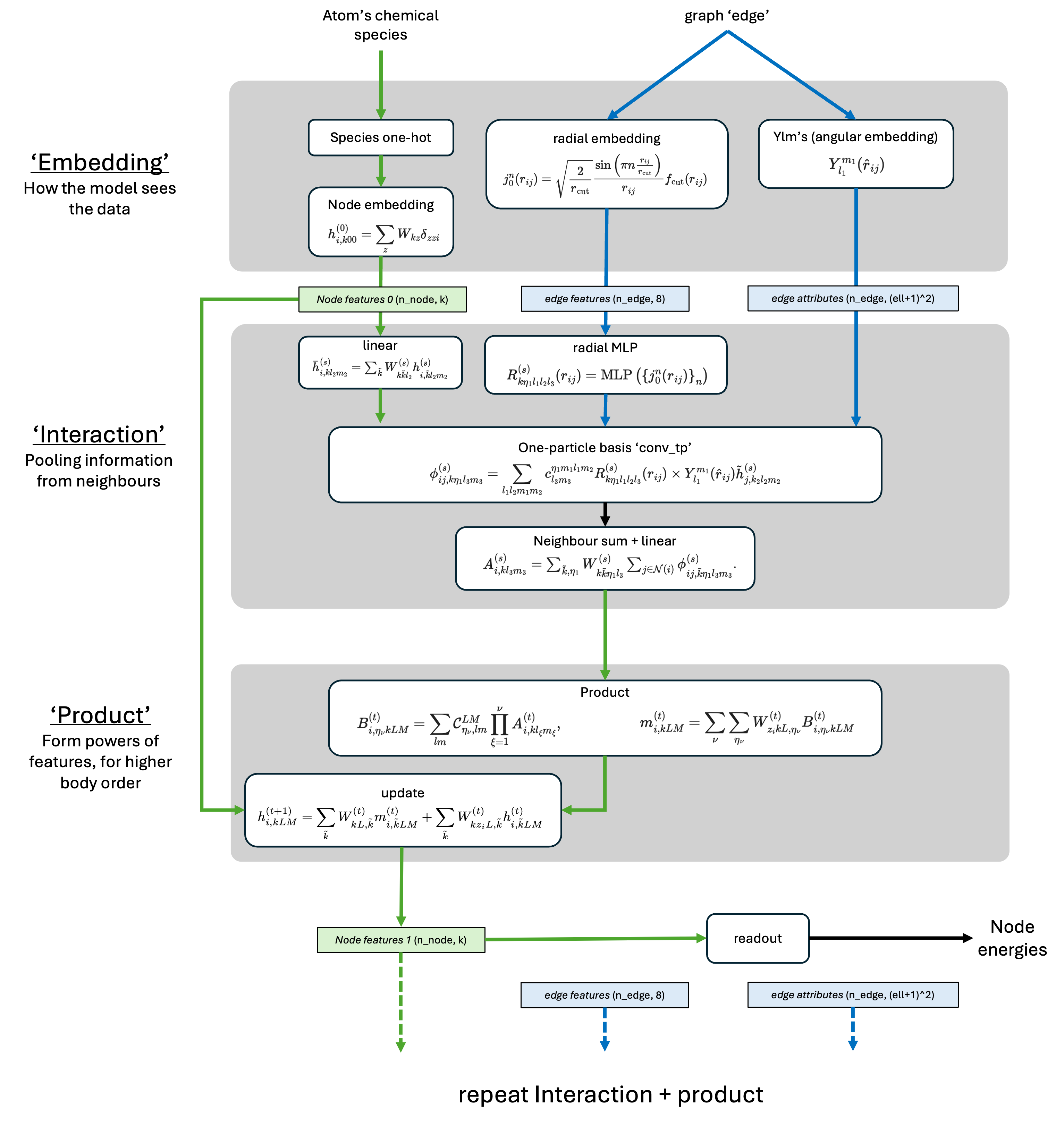}
        \caption{MACE Schematic \citep{Batatia2022mace}}
        \label{fig:maceschematic}
    \end{figure}

\subsection{Graph representation of atomic configurations}
\label{sec:grac}
We model an atomistic system as a graph embedded in $\mathbb{R}^3$: nodes are atoms and edges connect atoms within a cutoff. At layer $t$, the state of atom $i$ is
$\sigma_i^{(t)}=(\mathbf r_i, z_i, \mathbf h_i^{(t)})$, where $\mathbf r_i\in\mathbb R^3$ are coordinates, $z_i$ is the chemical species, and $\mathbf h_i^{(t)}$ are learnable features. A standard MPNN forward pass alternates \emph{message}, \emph{update}, and \emph{readout} steps:
\begin{equation}
    \mathbf m_i^{(t)}=\bigoplus_{j\in\mathcal N(i)} M^{(t)}\!\big(\sigma_i^{(t)},\sigma_j^{(t)}\big)
\end{equation}
\begin{equation}
    \mathbf h_i^{(t+1)}=U^{(t)}\!\big(\sigma_i^{(t)},\mathbf m_i^{(t)}\big),
\end{equation}
\begin{equation}
    E_i=\sum_{t=1}^{T} R^{(t)}\!\big(\sigma_i^{(t)}\big)
\end{equation}

Equivariance is expressed by labeling features with irreps $(L,M)$ so that under $Q\in O(3)$,
\begin{equation}
    h^{(t)}_{i,kLM}(Q\cdot \{\mathbf r\})=\sum_{M'} D^L_{M'M}(Q)\,h^{(t)}_{i,kLM'}(\{\mathbf r\}),
\end{equation}

where $D^L$ is the Wigner $D$-matrix . Scalars have $L{=}0$; $L{>}0$ are equivariant tensors.

\subsection{Higher-order Message Passing in MACE}
\label{sec:homp}
MACE expands each message in a hierarchical body-order series up to correlation order $\nu$:
\begin{equation}
    \mathbf m_i^{(t)}=\sum_{j} u_1 + \sum_{j_1,j_2} u_2 + \cdots + \sum_{j_1,\dots,j_\nu} u_\nu
\end{equation}

with sums over neighbors of $i$. Writing the sums with self-interactions enables a tensor-product parameterization that avoids the exponential cost of explicit triplet/quadruplet enumeration (contrast with many-body $\sum_{j_1<\cdots<j_\nu}$).

\paragraph{Two-body edge embedding.}
For each layer, MACE first builds \emph{edge-pooled} two-body features $A_i^{(t)}$ by combining a learned radial basis $R^{(t)}$, spherical harmonics $Y_{\ell m}(\hat{\mathbf r}_{ji})$, and species-mixed neighbor features via Clebsch–Gordan (CG) coupling:
\begin{equation}
    A^{(t)}_{i,kl_3m_3}=\!\!\sum_{l_1m_1,l_2m_2}\! C^{l_3m_3}_{l_1m_1,l_2m_2}\!\!\sum_{j\in\mathcal N(i)}\! R^{(t)}_{k l_1 l_2 l_3}(r_{ji})\,Y_{l_1 m_1}(\hat{\mathbf r}_{ji})
\sum_{\tilde k} W^{(t)}_{k\tilde k l_2}\,h^{(t)}_{j,\tilde k l_2 m_2}
\label{eq:26}
\end{equation}

In the first layer, $h^{(1)}_j$ encodes only $z_j$, yielding a cheaper specialization.

\paragraph{Node-local higher-order tensor products.}
Higher-order features $B_i^{(t)}$ are then formed on \emph{nodes} by taking $\nu$-fold tensor products of $A_i^{(t)}$ (mixing channels with weights $w$) and symmetrizing with generalized CG coefficients $C^{LM}_{\eta_\nu,lm}$ :
\begin{equation}
    B^{(t)}_{i,\eta_\nu k LM}=\sum_{lm} C^{LM}_{\eta_\nu,lm}\prod_{\xi=1}^{\nu}\Big(\sum_{\tilde k} w^{(t)}_{k\tilde k \ell_\xi}\,A^{(t)}_{i,\tilde k \ell_\xi m_\xi}\Big)
    \label{eq:27}
\end{equation}
These $C$ tensors are very sparse and can be precomputed, making  efficient.

\paragraph{Message, update, readout.}
Messages are linear combinations of $B_i^{(t)}$ across body orders and couplings:
\begin{equation}
    m^{(t)}_{i,kLM}=\sum_{\nu}\sum_{\eta_\nu} W^{(t)}_{z_i\, k L,\eta_\nu}\, B^{(t)}_{i,\eta_\nu k LM}.
\end{equation}

Updates are linear with a residual, and the readout maps the invariant ($L{=}0$) features to site energies with linear heads for all but the final layer.

\paragraph{MACE inference: energy and forces}
Let $\mathbf{R}=(\mathbf r_1,\dots,\mathbf r_N)$ be atomic positions and $\mathbf{Z}=(Z_1,\dots,Z_N)$ the species. From the final layer’s equivariant features, take the invariant ($L{=}0$) scalars $\mathrm{Inv}\!\big(A_i^{(T)}\big)$ and map them to per-atom energies with a readout $\rho$:
\begin{equation}
\varepsilon_i \;=\; \rho\!\Big(\mathrm{Inv}\!\big(A_i^{(T)}\big),\,Z_i\Big), 
\qquad
E(\mathbf R,\mathbf Z) \;=\; \sum_{i=1}^{N} \varepsilon_i.
\end{equation}
Forces follow from energy conservation as the negative gradient of $E$:
\begin{equation}
\mathbf F_i \;=\; -\,\nabla_{\mathbf r_i}\, E(\mathbf R,\mathbf Z)
\;=\; -\,\sum_{j=1}^{N} \frac{\partial \varepsilon_j}{\partial \mathbf r_i}.
\end{equation}

    \subsection{Architecture details and current improvements}
    \label{sec:ad}

    A minimal \textsc{MACE} comprises a species embedding, two message–passing layers that build edge then node-local equivariant features and update hidden states, and a hierarchical readout that sums site energies. By forming higher-order correlations primarily via \emph{node-local} tensor products rather than many stacked layers, standard \textsc{MACE} keeps depth small, receptive fields compact, and evaluation highly parallelizable.

    For speed at scale, alternative designs relax some geometric constraints and favor GPU-friendly dense kernels. \textsc{EScAIP} inserts multi-head attention at the neighborhood level and reports $\!10\times$ faster inference with $5\times$ lower memory at state-of-the-art accuracy, indicating that scalable attention can rival stricter symmetry encoding in practice \citep{qu2024importancescalableimprovingspeed}. Complementary global strategies include sparsifying message passing (focusing on the most informative neighbors) and compressing models via knowledge distillation; notably, Hessian-augmented distillation for MACE-like architectures yields compact students up to $20\times$ faster while matching or slightly improving accuracy \citep{amin2025fastspecializedmachinelearning}.
    
    \subsection{Computational cost model}
    \label{sec:ccm}

    \paragraph{Dominant costs.}
    In MACE-like architectures, the principal hotspot is the \emph{edge-level} SO(3)-equivariant tensor product in Eq~\ref{eq:26}, where Clebsch–Gordan (CG) coupling combines spherical harmonics $Y_{\ell m}$, a radial basis, and neighbor features. The MACE design mitigates this by evaluating the expensive edge tensor product once (typically in the second layer) and subsequently building higher-order correlations with the \emph{node-local} tensor product of Eq~\ref{eq:27}, which scales with the number of nodes rather than edges.

\paragraph{FLOPs and memory trade-offs.}
    Operator-level profiling quantifies per-layer FLOPs and memory, enabling targeted speed–accuracy tuning. Raising the message body order $\nu$ or angular bandwidth $L$ ($=\ell_{\max}$) rapidly inflates parameters and FLOPs with diminishing accuracy gains. On the memory side, high-$\ell$ feature maps and long neighbor lists dominate activations; once intermediates saturate device memory, throughput becomes bandwidth-bound --- alleviated by activation compression, recomputation/checkpointing, or reduced precision \citep{qu2024importancescalableimprovingspeed}.
    
    \paragraph{Back-of-the-envelope FLOP scaling.}
    Let $d_n \equiv |\mathcal N(i)|$ denote the average number of neighbors per node, $K$ the number of feature channels, and $\ell_{\max}$ the maximal angular order. Denote by $N_{\text{rad}}$ the size of the radial basis and by $N_{\text{path}}$ the number of retained coupling paths after CG sparsification. The leading asymptotics for the principal building blocks are:
    \begin{itemize}
        \item \textbf{Spherical harmonics and radial basis (per edge):} 
            \begin{equation}
                \mathcal O\!\big(d_n\,\ell_{\max}^2\big) \quad \text{to evaluate } \{Y_{\ell m}\}_{\ell \le \ell_{\max}}, 
            \end{equation}
            \begin{equation}
                \mathcal O\!\big(d_n\,N_{\text{rad}}\big) \quad \text{for the radial MLP basis.}
            \end{equation}
        \item \textbf{Eq.~\ref{eq:26} Clebsch–Gordan contraction (per edge):}  

            \begin{equation}
            \mathcal O\!\Big(d_n\,K^2 \sum_{\text{allowed }(\ell_1,\ell_2,\ell_3)} 
            (2\ell_1{+}1)(2\ell_2{+}1)(2\ell_3{+}1)\Big).
            \end{equation}

        \item \textbf{Eq.~\ref{eq:27} node tensor product (per node):}
        \begin{equation}
        \mathcal O\!\big(K\,N_{\text{path}}\,(2L{+}1)\big),
        \end{equation}
    \end{itemize}

    after precontracting Clebsch–Gordan coefficients with weights.
    While pruning to irreducible, allowed CG paths reduces constants, the edge-proportional scaling of Eq.~\ref{eq:26} remains the chief driver of wall time becoming target for mixed-precision execution, kernel fusion, and neighbor-wise batching

    \paragraph{Memory traffic and tensor shapes.}
    Key activations have the following tensor shapes:
    
    \begin{equation}
    A_i^{(t)} \in \mathbb{R}^{N_{\mathrm{atoms}} \times K \times (\ell_{\max}+1)^2}.
    \end{equation}
    
    \begin{equation}
    B_i^{(t)} \in \mathbb{R}^{N_{\mathrm{atoms}} \times K \times N_{\mathrm{path}} \times (2L+1)}.
    \end{equation}
    
    \begin{equation}
    m_i^{(t)} \in \mathbb{R}^{N_{\mathrm{atoms}} \times K \times (2L+1)}.
    \end{equation}
    
    Reducing $\ell_{\max}$ and $N_{\mathrm{path}}$ directly shrinks activation size and alleviates bandwidth pressure.

\section{Component-Level Profiling and Bottlenecks}
    \label{sec:clpb}
    
    \subsection{Profiling Methodologies}
    \label{sec:pm}

    The PyTorch Profiler (with TensorBoard) instruments training and inference at the operator level, reporting per-op wall time, FLOP estimates, memory allocation/peaks, and underlying CUDA kernel names \citep{paszke2019pytorchimperativestylehighperformance}. End-to-end traces of MACE forward/backward passes localize runtime to specific components and typically identify tensor-product contractions, neighbor-list construction, and scatter/gather as dominant hotspots.
    
    For hardware-level analysis, Nsight Systems and Nsight Compute provide complementary views \citep{Institute}. Nsight Systems reconstructs CPU–GPU timelines (overlap, kernel-launch patterns, longest-running kernels), while Nsight Compute exposes kernel micro-metrics (achieved occupancy, memory coalescing, cache hit rates, warp execution efficiency, divergence, DRAM throughput). Together, they map each MACE subroutine to a quantified cost and guide optimizations such as kernel fusion, mixed-precision execution, and Tensor-Core–aware reformulations.
    
    \subsection{Bottlenecks study}
    \label{sec:be}
    
    Profiling of SO(3)-equivariant, MACE-like models consistently pinpoints the Clebsch–Gordan (tensor-product) blocks as the dominant hotspot, accounting for roughly $75$–$90\%$ of end-to-end runtime and memory due to high-order dense contractions that produce large intermediates and stress memory bandwidth \citep{lee2025flashtp}. 
    
    A second major limiter is spherical-harmonic evaluation on every atom–neighbor edge; its cost grows with $\ell_{\max}$ and system size, and standard kernels exhibit poor time/memory behavior at scale --- motivating specialized, more efficient SH implementations \citep{lee2024scaling}.

\section{Precision-aware acceleration}
\label{sec:paa}
    \subsection{Numerical precision theory}
    \label{sec:npt}
    
\begin{figure}[h]
        \centering
        \includegraphics[width=\linewidth]{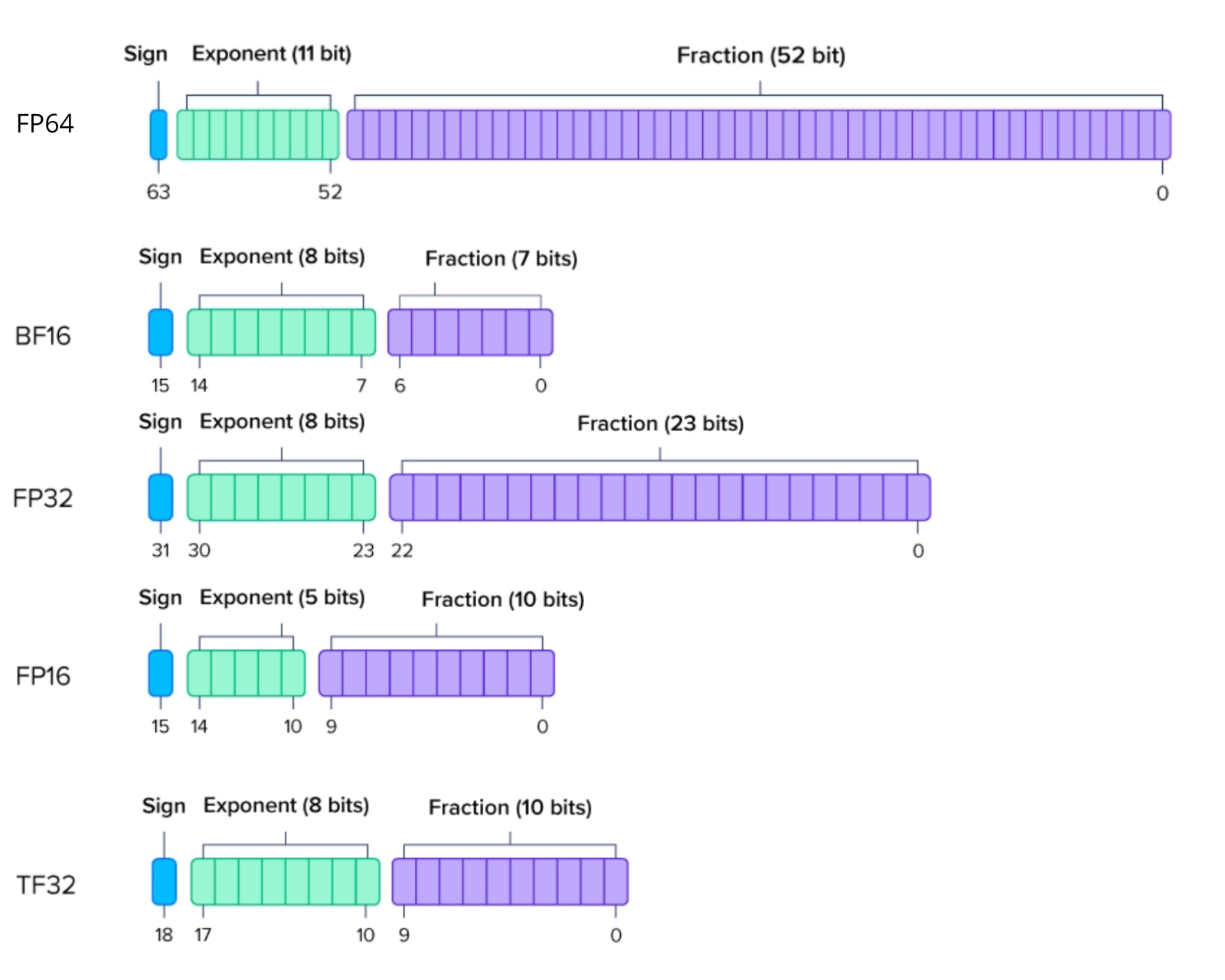}
        \caption[Numerical precision]{Bit allocations for the floating-point formats used in this work. FP64 offers the highest precision; FP32 is the single-precision baseline. FP16 reduces both exponent and fraction widths; BF16 trades mantissa precision for FP32-like range; TF32 preserves FP32 range with a 10-bit mantissa to enable Tensor-Core acceleration for FP32-coded kernels. The mantissa width governs machine epsilon and hence rounding granularity \citep{Rshravan_2025}.}
        \label{fig:numerical_prec}
    \end{figure}

    Floating-point formats allocate bits to sign, exponent, and fraction (mantissa), which jointly determine dynamic range and resolution. Figure~\ref{fig:numerical_prec} depicts the bit layouts for FP64 $(1{+}11{+}52)$, FP32 $(1{+}8{+}23)$, FP16 $(1{+}5{+}10)$, BF16 $(1{+}8{+}7)$, and TF32 $(1{+}8{+}10)$. The fraction width $p$ sets the machine epsilon $\,\varepsilon\approx 2^{-p}\,$ near unity and the unit-in-the-last-place (ULP) at magnitude $x$ as $\mathrm{ulp}(x)\approx |x|\,\varepsilon$ \citep{Goldberg1991}. Consequently,
    $\varepsilon_{\text{FP64}}\!\approx\!2^{-52}$,
    $\varepsilon_{\text{FP32}}\!\approx\!2^{-23}$,
    $\varepsilon_{\text{TF32}}\!\approx\!2^{-10}$,
    $\varepsilon_{\text{FP16}}\!\approx\!2^{-10}$,
    and $\varepsilon_{\text{BF16}}\!\approx\!2^{-7}$.
    Because BF16 and TF32 retain FP32’s 8-bit exponent, they preserve FP32-like dynamic range; FP16’s 5-bit exponent narrows the range and is therefore more prone to overflow/underflow in training.
    
    \paragraph{Rounding and when to keep FP32 accumulations.}
    Rounding error grows with the number of terms in reductions and dot products; a length–$k$ inner product accumulated in precision $\varepsilon_{\text{acc}}$ incurs a forward error of order $k\,\varepsilon_{\text{acc}}$ times the sum of magnitudes. Accordingly, mixed-precision policies execute GEMMs in FP16/BF16 on Tensor Cores but \emph{accumulate} in FP32 and promote numerically sensitive reductions (means/variances, softmax, large summations) to FP32 \citep{PyTorch_2024}. This design curbs the error budget while retaining low-precision throughput. TF32 offers a compatibility path for FP32-coded matmuls with FP32 outputs but a 10-bit mantissa internally on Ampere Tensor Cores \citep{Krashinsky_Giroux_Jones_Stam_Ramaswamy_2020}.
    
    \paragraph{Machine epsilon and catastrophic cancellation.}
    Subtraction of nearly equal quantities destroys leading mantissa bits (“catastrophic cancellation”), leaving roughly $\log_2(|x-y|/|x|)$ effective bits \citep{Goldberg1991}. Practical mitigations include (i) performing differences and variance computations in FP32, (ii) using cancellation-robust formulas (e.g., Welford’s method), and (iii) relying on FP32 accumulators in mixed-precision GEMMs (with FP32 accumulation).

    \subsection{Quantization}
    \label{sec:q}
    
    Quantization reduces the numeric precision used to represent weights and activations in order to cut memory footprint and increase arithmetic throughput. The benefit is twofold: model size and bandwidth scale down roughly in proportion to bit-width, and modern accelerators can execute many more low-precision operations per cycle. The principal risk is degradation from rounding and clipping, which can accumulate and especially in iterative generators such as diffusion samplers or long physics processes making careful control of numerical error important \citep{fan2025sqdmacceleratingdiffusionmodels}. In practice, FP16 and BF16 often preserve inference accuracy for vision, language, and physical models with minimal changes, but integer quantization typically yields the largest memory and latency gains and therefore receives most attention in deployment.
    
    Post-training quantization (PTQ) converts a trained FP32 model to lower precision without retraining. Parameters and intermediate tensors are mapped to a discrete set via an affine transform determined during a short calibration pass on representative data (estimating scale and, for asymmetric schemes, zero-point). Simple policies per-tensor and symmetric ranges are fast but may be sensitive to outliers; robust PTQ commonly combines per-channel weight quantization with outlier-aware calibration such as percentile clipping to stabilize activation ranges \citep{Cheng_Ho_Xin_2024}. When the target is FP16/BF16, deployment is often as simple as casting the model and enabling half-precision execution; because these formats remain floating-point, well-trained networks usually require little or no calibration, provided intermediate values remain within range \citep{Cheng_Ho_Xin_2024}.
    
    Tooling support is mature on NVIDIA hardware. Developers frequently use TensorRT for highly optimized FP16 deployment, or framework bridges such as ONNX Runtime or OpenVINO for graph-level conversion and calibration. For graph neural networks, PTQ remains feasible but can be complicated by custom message-passing kernels; a pragmatic approach is to quantize dense linear submodules while leaving numerically delicate reductions and aggregations in higher precision, or to implement specialized kernels where needed. Empirically, well designed PTQ retains accuracy close to full precision: for example, GNN studies report that 8-bit inference can achieve “similar accuracy to full precision” when PTQ is coupled with minor fine-tuning \citep{tailor2021degreequantquantizationawaretraininggraph}. Given typical 4$\times$ reductions in model size from FP32 to INT8 and frequent 2$\times$ speedups, quantization reduces the bit-width of MACE’s weights and activations, shrinking memory traffic and enabling higher-throughput low-precision kernels (e.g., INT8/FP16 tensor cores) for its contraction-dominated blocks, thereby cutting inference latency with minimal accuracy loss under calibrated PTQ \citep{Cheng_Ho_Xin_2024,tailor2021degreequantquantizationawaretraininggraph}.

    \subsection{Mixed-precision training and inference}
    \label{sec:mpti}
    
        Lower-precision floating point substantially accelerates MACE inference on NVIDIA GPUs by unlocking Tensor Core throughput. On A100 (Ampere), half-precision tensor math reaches on the order of $\sim$312 TFLOPS for FP16/BF16 which is roughly an order of magnitude higher than FP32 while even Turing-class GPUs such as the RTX 2080 Ti expose FP16 Tensor Cores at $\sim$107.6 TFLOPS \citep{Krashinsky_Giroux_Jones_Stam_Ramaswamy_2020}. Casting MACE’s dense contractions to FP16/BF16 enables these kernels for matrix–multiply and convolution-like workloads, yielding large throughput gains with limited accuracy loss by design \citep{StackGpu_2025}. Empirically, switching MACE from FP64 to FP32 already provides a $\sim$2$\times$ speedup with only minor accuracy impact; recent evaluations report FP32 energies and forces deviating from FP64 by $\mathcal{O}(1)$ meV and $\mathcal{O}(10^{-2})$ meV/\AA{}, respectively \citep{Cho_Yoo_Yi_Yang_Lee_Jeong_Choo_2025}. This robustness motivates further reductions to FP16/BF16 at inference time. In particular, BF16 on A100 retains FP32’s 8-bit exponent, mitigating overflow and underflow, while halving memory and activating Tensor Cores. Mixed-precision execution that keeps the bulk of linear algebra in 16-bit and upcasts only numerically sensitive accumulations to FP32 therefore offers a high-throughput path on both A100 and 2080 Ti with minimal loss in force-field fidelity \cite{Cho_Yoo_Yi_Yang_Lee_Jeong_Choo_2025}. However, the use of lower numerical precision in ML force fields is largely uncharted territory where traditional ML potentials often use double precision (DeepMD uses FP64, others FP32) due to stability requirements \citep{Zhou_2025}. The FastCHGNet project explicitly states that “to the best of our knowledge, there are currently no atomic potential models trained using half-precision” sharing the sensitivity and high accuracy demands of force-field training as obstacles to quantization. This reinforces the perspective that focusing on low-precision is an open research gap.
    
    Operationally, mixed precision denotes assigning different numeric formats to different parts of the computation to balance speed, memory, and stability. On NVIDIA GPUs, the practical inference regime runs tensor-core–eligible kernels (e.g., \texttt{matmul}/\texttt{addmm}/\texttt{einsum}) in FP16 or BF16 with FP32 accumulation, while promoting numerically delicate steps like softmax, layer-normalization statistics, and large reductions to FP32. Framework-level automatic mixed precision implements these dispatch rules transparently via “autocast,” so most models obtain tensor-core acceleration and memory savings without manual kernel selection. For deployment, FP32/FP16/FP16 (storage/compute) are used in practice. Casting pretrained FP32 weights to FP16 typically perturbs parameters well within model tolerance, whereas retaining FP32 weights and casting on the fly trades extra bandwidth for conservative storage. BF16 as both storage and compute format is increasingly common on Ampere/Hopper because it delivers FP16-like speed with a wider dynamic range.
    
    TF32, introduced on Ampere, offers a compatibility path by running FP32-coded GEMMs on tensor cores using a 10-bit mantissa while preserving FP32 interfaces. It is most relevant when legacy FP32 pipelines must remain unchanged; for new deployments, full FP16/BF16 generally yields higher throughput and better memory compression. When enabled, TF32 can provide substantial speedups for GEMM-dominated graphs with an effective precision close to FP16, while emitting FP32 outputs \citep{Krashinsky_Giroux_Jones_Stam_Ramaswamy_2020}.
    
    The same principles extend to training, with additional safeguards for optimization dynamics. A standard FP16 recipe performs forward and backward in half precision, maintains FP32 master weights, converts gradients to FP32 for the optimizer step, and applies dynamic loss scaling to prevent gradient underflow; PyTorch’s \texttt{GradScaler} (and earlier NVIDIA Apex) implement this pattern reliably. Training in BF16 typically removes the need for loss scaling due to its larger exponent range, while retaining near-FP16 throughput on recent GPUs. Empirically, both FP16 and loss scaling and BF16 reach FP32-comparable convergence for large models, reduce activation/optimizer memory, and enable larger effective batch sizes. For equivariant force-field architectures like MACE, these policies align naturally: tensor-product contractions and message-passing GEMMs benefit from half-precision tensor cores with FP32 accumulation, while normalizations, long-range reductions, and global energy/force aggregations are kept in FP32 via autocast. Training under mixed precision also conditions weight distributions to low-precision arithmetic, making FP16/BF16 inference the default high-throughput choice; if strict FP32 interfaces must persist, TF32 remains a viable compatibility acceleration \citealp{Krashinsky_Giroux_Jones_Stam_Ramaswamy_2020}).

    \subsection{Precision‐loss metrics}
    \label{sec:plm}
    
    \paragraph{Static precision–error metrics (vs.\ FP64).}
    To quantify numerical differences from a numerical precision speed up process between a reduced-precision configuration and a double-precision (FP64) baseline on a test set $\mathcal S$, the literature would denote for structure $s$ with $N_s$ atoms, let $(E^{64}_s,\mathbf F^{64}_s)$ be FP64 energy/forces and $(\hat E_s,\hat{\mathbf F}_s)$ the accelerated predictions. Absolute and relative \emph{per-structure} errors are
    \begin{equation}
    \mathrm{AE}_E(s)=\frac{1}{N_s}\,\bigl|\hat E_s-E^{64}_s\bigr|
    \end{equation}
    \begin{equation}
            \mathrm{RAE}_E(s)=\frac{\bigl|\hat E_s-E^{64}_s\bigr|}{\bigl|E^{64}_s\bigr|}
    \end{equation}
    \begin{equation}
    \mathrm{AE}_F(s)=\frac{1}{N_s}\sum_{a=1}^{N_s}\bigl\|\hat{\mathbf f}_{s,a}-\mathbf f^{64}_{s,a}\bigr\|_2
    \end{equation}
    \begin{equation}
        \mathrm{RAE}_F(s)=\frac{\bigl\|\hat{\mathbf F}_s-\mathbf F^{64}_s\bigr\|_F}{\bigl\|\mathbf F^{64}_s\bigr\|_F}.
    \end{equation}
    where the dataset aggregates as the mean and the $95^{\mathrm{th}}$ percentile (P95) across $s\in\mathcal S$:
    
    This protocol follows prior work validating that carefully engineered single/mixed precision can match FP64 force accuracy for held-out molecules \citep{kovács2025maceofftransferableshortrange} and is consistent with precision-stability assessments used in MLFF acceleration studies \citep{amin2025fastspecializedmachinelearning}.
    
    \paragraph{Constant-Temperature, Constant-Pressure ensemble (NPT) MD metrics}
    For an NPT trajectory, progress of the simulation is monitored with observables recorded in the driver: instantaneous density $\rho(t)$, pressure $p(t)$, and temperature $T(t)$. With total mass $M=\sum_i m_i$ and simulation volume $V(t)$,
    \begin{equation}
    \rho(t)=\frac{M}{V(t)}\times c_{\rho}
    \end{equation}
    \begin{equation}
       \bar{\rho}=\frac{1}{K}\sum_{k=1}^{K}\rho(t_k), 
    \end{equation}
    \begin{equation}
        \sigma_{\rho}=\sqrt{\frac{1}{K-1}\sum_{k=1}^{K}\bigl(\rho(t_k)-\bar{\rho}\bigr)^2},
    \end{equation}
    where $c_{\rho}$ is the unit-conversion factor used in the code. Pressure is obtained from the virial/stress as
    \begin{equation}
    p(t)=-\tfrac{1}{3}\,\mathrm{tr}\,\boldsymbol{\sigma}(t)
    \end{equation}
    \begin{equation}
        \bar{p}=\frac{1}{K}\sum_{k=1}^{K}p(t_k)
    \end{equation}
    \begin{equation}
        \Delta p=\bar{p}-p_{\mathrm{target}}.
    \end{equation}
    By using normal static precision mentioned above, $\bar T$ to $T_{\mathrm{target}}$ is compared. For reporting, trajectory means, standard deviations, and P95 deviations (e.g., $\mathrm{P95}\{|\,\rho(t)-\rho_{\mathrm{target}}|\}$) is often used. These NPT metrics complement the static FP64 comparisons by testing whether reduced precision preserves equilibrium thermophysical properties under barostat/thermostat control \citep{amin2025fastspecializedmachinelearning}.

\section{SO(3) efficiency}
\label{sec:se}

    \subsection{Sparse structure in CG/tensor products}
    \label{sec:sscp}

    The tensor product of spherical harmonics in $\mathrm{SO}(3)$–equivariant networks couples feature channels of angular momenta $\ell$ via Clebsch–Gordan coefficients (equivalently, Wigner $3j$ symbols). Owing to selection rules (e.g., triangle inequalities and $m$-sum constraints), many of these coefficients vanish, implying that only a subset of irrep triplets actually interact. Recent implementations exploit this structured sparsity to reduce computational cost. In particular, the MACE-OFF framework accelerates higher-order message passing by omitting all couplings corresponding to zero Clebsch–Gordan entries and evaluating only the nonzero coupling paths in the tensor–product basis, thereby cutting both arithmetic and memory traffic \citep{kovács2025maceofftransferableshortrange}. Furthermore, a very recent study by \citep{xie2025pricefreedomexploringexpressivity} analyzes the cost of equivariant tensor product operations and their expressivity–speed tradeoffs where they discuss the Gaunt Tensor Product (GTP), proposed by \citep{luo2024enablingefficientequivariantoperations}, which significantly speeds up high-order tensor contractions by using precomputed Gaunt coefficients (spherical harmonic integrals) in place of Clebsch–Gordan coupling. However, such speedups "come at the cost of expressivity" in the model even though they found a way to simplify GTP via a spherical grid method, achieving about 30\% faster training of a MACE model with virtually no loss in asymptotic complexity. In addition, the pipeline of \emph{spherical-harmonic evaluation} to \emph{basis projection} to \emph{message update} is fused into custom kernels so that intermediates reside in registers and shared memory, eliminating redundant loops and global-memory round trips. Taken together, sparsity-aware coupling and kernel fusion yield substantial speedups over naïve implementations without altering model semantics \citep{kovács2025maceofftransferableshortrange}.
    
    \subsection{Optimizing spherical harmonics kernels}
    \label{sec:oshk}

    Evaluating spherical harmonics $Y_{\ell m}(\theta,\phi)$ for each neighbor constitutes a significant computational burden, especially as the maximal angular order $\ell$ increases. This cost can be mitigated through specialized numerical schemes and low-level kernel optimizations. A notable line of work employs the Triton GPU compiler to implement custom kernels for high-order spherical harmonics, attaining up to $5\times$ speedups and reducing memory usage to roughly one third of standard e3nn-based implementations. Because Triton generates device-specific code, these kernels are effectively hardware-agnostic, running efficiently across NVIDIA, AMD, and Intel GPUs \citep{lee2024scaling}. Complementarily, the \texttt{sphericart} library provides highly optimized C++ routines for real spherical harmonics --- leveraging stable recurrence relations and SIMD vectorization -- and has been integrated into MACE-OFF. Substituting \texttt{sphericart} for baseline implementations substantially lowers the cost of computing $Y_{\ell m}$ over all neighbors \citep{kovács2025maceofftransferableshortrange}. Taken together, these advances enable the inclusion of higher-order $\ell$ terms --- often beneficial for accuracy --- while substantially curbing the associated performance penalties and the speed
    
    \subsection{Fusion of tensor products}
    \label{sec:ftp}

    A recent contribution to accelerating equivariant force-field models is \textsc{FlashTP}, a specialized library that fuses and optimizes tensor–product (TP) operations for equivariant MLIPs \citep{lee2025flashtp}. In baseline frameworks such as e3nn and standard MACE implementations, an equivariant layer typically issues many small matrix multiplications --- one per admissible pair of input/output irreducible representations --- incurring substantial kernel–launch and memory–traffic overhead. \textsc{FlashTP} replaces this pattern with a single, fused GPU kernel that evaluates all required tensor products in one pass, reusing on-chip data and minimizing global-memory round trips. The library further implements \emph{path aggregation}, in which multiple interaction paths (sequences of irrep couplings) are combined into a larger batched contraction that better saturates the GPU. By eliding zero-multiply work implied by selection rules and consolidating epilogues, \textsc{FlashTP} reduces redundant data movement and improves arithmetic intensity. Empirically, the authors report up to $41\times$ per-kernel speedups over the e3nn baseline, end-to-end inference accelerations of approximately $4.2\times$, and about $6\times$ lower memory usage. Given the tensor product’s dominance in the runtime profile of MACE-like architectures, these improvements translate directly into multi-fold model speedups. \textsc{FlashTP} is released as a plug-in that can replace the native e3nn tensor-product operator with minimal code changes, yielding immediate performance gains \citep{lee2025flashtp}.

\section{GPU-centric optimization}
\label{sec:gco}

\subsection{Tensor Core mapping and mixed precision}
\label{sec:tcmmp}

Modern NVIDIA Tensor Cores deliver disproportionately higher dense-GEMM throughput at reduced precision, making FP16/BF16 the natural targets for MACE’s contraction-heavy blocks (tensor products, channel mixing). On A100, FP16/BF16 Tensor Core ops reach $\mathcal{O}(10^2)$ TFLOPS versus $\sim\!20$ TFLOPS for FP32; H100 extends this advantage further \citep{Krashinsky_Giroux_Jones_Stam_Ramaswamy_2020, Van_den_Berghe_2024}. Realizing these gains requires (i) casting core contractions to FP16/BF16 (or INT8 when calibrated) with FP32 accumulation, (ii) batching and tiling equivariant contractions as dense or batched GEMMs that match TC tile shapes (e.g., $16{\times}16$), and (iii) using cuBLAS/cublasLt or CUTLASS/WMMA epilogues to fuse bias/activation and avoid extra memory traffic \citep{StackGpu_2025}. In equivariant message passing, spherical-harmonic bases combined with learned weights can be aggregated into block-dense multiplies; packing neighbor features contiguously enables TC execution, while asynchronous copies and shared-memory staging keep TCs fed. Complementarily, FTC-GNN shows that offloading dense algebra to Tensor Cores and handling sparse gather/scatter on CUDA cores yields $\sim\!5\times$ speedups via sparse-to-dense transformations \citep{wu2025acceleratingsparsegraphneural}. Ampere’s $2{:}4$ structured sparsity can further double TC throughput when weights are pruned and fine-tuned to the supported pattern, compounding mixed-precision gains \citep{Krashinsky_Giroux_Jones_Stam_Ramaswamy_2020}.

\paragraph{Throughput and footprint motivation.}
Half-precision formats also halve memory and bandwidth relative to FP32, enabling larger batches/models or better on-chip caching—crucial for bandwidth-bound GNNs \citep{Krashinsky_Giroux_Jones_Stam_Ramaswamy_2020, Van_den_Berghe_2024}. The table summarizes precision, memory footprint, and peak dense GEMM rates:

\begin{table}[h!]
\centering
\renewcommand{\arraystretch}{1.2}
\resizebox{\textwidth}{!}{%
\begin{tabular}{lccc}
\toprule
\textbf{Precision (bits)} & 
\textbf{Memory Footprint (vs FP32)} & 
\textbf{Peak Compute (A100)} & 
\textbf{Peak Compute (H100)} \\
\midrule
FP32 (32-bit float) & 
1× (4 bytes) & 
19.5 TFLOPS & 
67 TFLOPS \\

FP16 (half) & 
0.5× (2 bytes) & 
312 TFLOPS ($\sim$16×) & 
1,979 TFLOPS ($\sim$30×) \\

BF16 (bfloat16) & 
0.5× (2 bytes) & 
312 TFLOPS ($\sim$16×) & 
1,979 TFLOPS (same as FP16) \\
\bottomrule
\end{tabular}
}
\caption[Precision formats on NVIDIA A100/H100]{Precision formats on NVIDIA A100/H100 with relative memory use and peak dense GEMM throughput. Factors indicate theoretical speedup vs FP32 FMAs \citep{Van_den_Berghe_2024}.}
\end{table}

\subsection{Memory layout \& kernel fusion}
\label{sec:mlkf}

MACE’s gather–scatter pattern induces scattered reads/writes and many small kernels, stressing DRAM and launch overhead \citep{Wang_Yu_Zhao_Li_Zhang}. Use contiguous neighbor/edge layouts for coalesced access; tile and stage hot data (neighbor features, $Y_{\ell m}$) in registers/shared memory to cut refetches. Fuse message formation with accumulation to avoid materializing intermediates --- DGL’s fused message passing reports up to $19\times$ higher throughput with much lower memory use \citep{Wang_Yu_Zhao_Li_Zhang}. Reorder work into cache-sized tiles and employ warp shuffles to improve locality; poor locality and excess traffic are the dominant bottlenecks \citep{huang}. Storing activations/weights in FP16/BF16 halves traffic and improves cache residency while retaining FP32 accumulations for numerical safety \citep{Krashinsky_Giroux_Jones_Stam_Ramaswamy_2020}.

\subsection{A100-specific features, RTX 2080 Ti notes}
\label{sec:asf}

\paragraph{Structured sparsity in tensor products.}
Ampere-class A100 GPUs expose \emph{structured} $2{:}4$ sparsity: when weights are pruned so that two of every four elements are zero in a prescribed pattern, Tensor Cores can execute the corresponding matrix multiplications at roughly double throughput. Applying this to MACE implies magnitude-based pruning (and subsequent fine-tuning) of dense linear submodules and/or block–tensor-product mixing layers to respect the $2{:}4$ constraint, after which sparse-TC kernels can be dispatched for the dense contractions that dominate runtime. While this has yet to be demonstrated for SO(3)-equivariant GNNs, it provides a hardware-aware compression route that can compound with mixed precision to yield additional $\sim 2\times$ inference speedups, with the usual caveat that accuracy must be recovered via careful retraining. More broadly, accelerator-aware optimization has proven decisive for scientific ML workloads, motivating similar co-design for equivariant force fields \citep{musaelian2023scalingleadingaccuracydeep}.

\paragraph{Asynchronous compute and memory copy.}
A100 supports asynchronous data movement (e.g., \texttt{cp.async}/LDGSTS) and overlapping compute with memory transfers, which is particularly valuable for irregular message passing. Tiling edge blocks to fit in on-chip memory, prefetching spherical-harmonic and radial-basis tiles, and scheduling Tensor Core GEMMs to overlap with neighborhood staging can hide DRAM latency. These techniques are complementary to TF32/FP16/BF16 execution and structured sparsity, and they improve utilization without altering model semantics \citep{musaelian2023scalingleadingaccuracydeep}. For RTX~2080~Ti, Tensor Cores natively accelerate FP16 but lack BF16 and $2{:}4$ sparsity support; mixed-precision FP16 with FP32 accumulation remains the primary path on this class of hardware.

\subsection{CUDA kernel optimization}
\label{sec:ckofc}

\paragraph{Custom CUDA for equivariant operators.}
Hand-optimized CUDA C++ kernels can outstrip generic operator stacks when tailored to the dataflow of message passing. The MACE-OFF implementation provides an illustrative precedent: neighbor summation, spherical-harmonic evaluation, and equivariant linear layers were rewritten as fused, inference-oriented kernels, with careful management of registers/shared memory and launch geometry. These kernels were further integrated with LAMMPS and OpenMM to enable zero-copy interoperability, eliminating CPU–GPU round trips by sharing device-resident coordinates and features across the MD code and the ML potential. In aggregate, this co-design improved memory bandwidth utilization and delivered substantial end-to-end speedups for production inference \citep{kovács2025maceofftransferableshortrange}.

\paragraph{Frameworks and compilers (cuBLAS/cublasLt, CUTLASS, cuEquivariance, Triton, \texttt{torch.compile}, TensorRT).}
High-performance libraries and compilers provide a scalable alternative to bespoke kernels while still mapping computations onto Tensor Core–friendly primitives. cuBLAS/cublasLt and CUTLASS expose batched GEMM and epilogue fusion that match the tensor-product contractions prevalent in MACE; NVIDIA’s cuEquivariance supplies custom kernels for equivariant operators that can be enabled directly from model code. Compiler stacks --- \texttt{torch.compile} in PyTorch~2.x, and Triton-backed code generation --- perform graph capture, kernel fusion, layout specialization, and low-level scheduling to approach hand-tuned performance while preserving Python-level ergonomics \citep{PyTorch_2024}. Notably, Triton generates device-specific kernels and has demonstrated cross-vendor portability, enabling efficient spherical-harmonic and related primitives on NVIDIA, AMD, and Intel GPUs with minimal code changes \citep{lee2024scaling}. For deployment, TensorRT offers calibrated FP16/INT8 execution and graph-level optimizations, complementing training-time compilation. Together, these systems reduce kernel launch overheads, raise arithmetic intensity, and unlock Tensor Core execution paths with limited engineering effort \citep{PyTorch_2024, kovács2025maceofftransferableshortrange}.
     
\subsection{CuEquivariance for speeding up matrix–tensor multiplication}
\label{sec:cfsmtm}

NVIDIA’s \texttt{cuEquivariance} provides optimized GPU primitives for equivariant neural operators, exposing fused and batched contractions that map SO(3) tensor products and related reductions onto Tensor Cores with minimal memory traffic. As documented in the official announcement, these kernels reduce kernel–launch overhead, improve arithmetic intensity, and enable mixed-precision execution (FP16/BF16/TF32) while preserving model semantics --- yielding substantial speedups for atomistic ML workloads that rely on dense linear algebra inside message passing \citep{Geiger_Kucukbenli_Zandstein_Tretina_2025}. In practice, replacing generic per-irrep micro-kernels with \texttt{cuEquivariance}’s fused GEMM pathways amortizes indexing and epilogue costs, which is particularly effective for MACE-style higher-order tensor products. Developer guidance, supported dtypes, and integration details are available in the library manual (\href{https://docs.nvidia.com/cuda/cuequivariance/}{NVIDIA cuEquivariance Documentation}).

\section{Model compression}
\label{sec:mc}
    \subsection{Pruning: structured sparsity for A100}
    \label{sec:p}

    Reducing model complexity after training is a general strategy to speed up inference. By pruning less important weights or entire channels, one can obtain a sparser, smaller model that computes faster. While we haven’t yet seen MACE-specific pruning studies, analogous works on other GNNs show that structured pruning can accelerate inference with minimal accuracy drop \citep{Neo_2024} \citep{jing2025autosculptpatternbasedmodelautopruning}. For example, removing redundant graph edges or message pathways can sparsify computations which is also called graph rewiring techniques \citep{bechlerspeicher2024graphneuralnetworksuse}. The last articles shows that GNNs actually tend to overfit the given graph-structure in where better solution can be obtained while ignoring the original graph. In MACE, one might prune high-order latent channels or irreducible representations that contribute little to final accuracy. Any such pruning would need to be balanced carefully to avoid degrading the force field’s precision, but it remains a promising avenue for model compression.

    \subsection{Knowledge distillation}
    \label{sec:kd}

    An effective route to accelerate inference is to train a compact \emph{student} model to emulate a larger \emph{teacher}. \citep{amin2025fastspecializedmachinelearning} (2025) propose a Hessian-augmented distillation scheme for equivariant force fields in which a small MACE-like student is supervised not only on scalar energies $E$ and forces $\mathbf{F}=-\nabla_{\mathbf{x}}E$ but also on the teacher’s energy Hessian $\nabla_{\mathbf{x}}^{2}E$ over a curated subset, thereby conveying local curvature information and stabilizing generalization. This enriched target set yields students that closely track the teacher’s behavior while being substantially lighter. Empirically, the distilled MLFFs attain up to $20\times$ faster energy/force predictions than the original large model, with accuracy maintained or marginally improved within the intended chemical domain. In practice, this enables a deployment pattern in which a broad, expressive MACE foundation model provides supervisory signals, followed by distillation to a lean, domain-specialized student used for production molecular simulations.

\section{Cross-model and benchmarking protocol}
\label{sec:cmhas}

\subsection{Distributed \& parallel training}
\label{sec:dpt}
MACE’s moderate parameter count makes it well suited to data-parallel (DDP) training, in which replicas process disjoint minibatches and synchronize gradients each step. For very large systems or long-range workloads, spatial domain decomposition distributes atoms and halo neighbors across devices, exchanging boundary data asynchronously to overlap communication with computation. This strategy has enabled excellent strong- and weak-scaling for closely related E(3)-equivariant local potentials integrated into LAMMPS on leadership systems, demonstrating that locality-preserving designs can exploit thousands of GPUs efficiently \citep{musaelian2023scalingleadingaccuracydeep}. Inference-oriented multi-GPU evaluators following the same principles achieve parallel speedups for production MD while exposing load-balance and communication bottlenecks that matter at scale \citep{kovács2025maceofftransferableshortrange}. Practical enhancements include gradient accumulation to amortize latency, mixed-precision gradient, and graph partitioning that minimizes inter-GPU traffic. While MACE is suspected to be trained on multiple GPUs in parallel, some parallel researched on scaling these models across multiple GPUs for large systems have been emerging. In particular, \citep{Park_2024} introduced SevenNet, a distributed parallel scheme for equivariant GNN potentials based on NequIP. They address the challenge of integrating message-passing networks with the spatial domain decomposition used in MD codes. SevenNet achieves over 80\% parallel efficiency on 32 GPUs (weak scaling) and was used to run MD simulations with 100,000+ atoms using an equivariant potential. While this is about strong-scaling rather than lower precision, it’s relevant in a systems sense --- it shows another way to speed up force-field models and to deploy them in production MD engines.

\subsection{Ensuring comparability in acceleration processes}
\label{sec:bp}

Benchmarking on standardized kernels and software paths used for MACE acceleration. Enabling \texttt{--enable$_{cueq}$=True} offloads equivariant operators to cuEquivariance’s optimized CUDA kernels and yields up to $5\times$ speedups for large models in training and inference; the same gains carry through LAMMPS/OpenMM integrations. Algorithmically, replacing conventional tensor products with the spherical-grid formulation reduces training runtime by $\sim 30\%$ without accuracy loss \citep{xie2025pricefreedomexploringexpressivity}. Industry benchmarks further show that FP16/INT8 execution on GPUs substantially lowers latency and cost \citep{Gupta_Kiely_2024}, and PTQ can deliver 8-bit inference with accuracy close to full precision \cite{tailor2021degreequantquantizationawaretraininggraph}.

For fair comparisons, it is commong to fix the hardware/software stack (GPU, driver, CUDA/cuDNN/cuBLAS, cuEquivariance), standardize precision regimes per run (FP64, FP32, FP16, and BF16  with identical calibration), and evaluate identical MACE architectures and datasets.

\section{Scope and Selection Criteria}
    Our design choices are deliberately pragmatic and empirical. We prioritize (i) reproducibility on available NVIDIA hardware, (ii) measurable impact on end-to-end runtime and memory, (iii) fidelity safeguards tied to energies, forces, and MD observables, and (iv) methods whose effects can be isolated via profiling. We do not attempt to reformulate MACE’s mathematical foundations; instead, we focus on profiling, precision management, and backend implementations that can be evaluated within the project timeline.
    
\section{Chapter's summary}
    Within the project’s scope, we concentrate on profiling and precision-aware acceleration rather than revisiting the full equivariant framework. Many published speedups rely on bespoke setups or incomplete reporting, limiting reproducibility and generalization. By profiling both training and inference and by rigorously testing low-precision regimes favored by modern GPUs, we provide a realistic account of MACE’s computational expense and a disciplined assessment of where efficiency can be gained without sacrificing predictive fidelity. The next section details the experimental setup used to quantify MACE’s computational cost, validate the preceding analysis, and evaluate acceleration strategies for the full model and its constituent blocks.

\chapter{Methodology}
\label{chapter:methodology}

\ifpdf
    \graphicspath{{Chapter_Methodology/Figs/Raster/}{Chapter_Methodology/Figs/PDF/}{Chapter_Methodology/Figs/}}
\else
    \graphicspath{{Chapter_Methodology/Figs/Vector/}{Chapter_Methodology/Figs/}}
\fi

\section{Hardware Setup}
\label{sec:hardware}

\subsection{Compute Nodes and GPUs}
This work was executed on University of Cambridge high-performance computing resources equipped with NVIDIA A100 80\,GB SXM GPUs, and --- where access constraints or technical issues required --- on Computational and Biological Learning (CBL) lab workstations hosting NVIDIA RTX 2080~Ti GPUs. The A100 80\,GB SXM provides substantially higher memory capacity and bandwidth than prior-generation accelerators, along with Tensor Core support for reduced-precision arithmetic (e.g., TF32/BF16) that is relevant to mixed-precision experiments reported later. As a result, throughput and energy efficiency on tensor-dense kernels are markedly improved relative to older GPUs. Detailed device characteristics for the A100 are provided in Table~\ref{tab:a100nvidia} (Appendix~\ref{appendix:nvidia}), and for the RTX~2080~Ti in Appendix~\ref{appendix:rtx}. Because our focus is methodological rather than architectural hardware design, we treat hardware as a fixed factor and note that further improvements in GPU capability would predictably translate into proportional speedups for the same software stack and models.

\subsection{Performance Measurement and Utilisation Considerations}
All timing was conducted under steady-state conditions with attention to arithmetic (compute) saturation and GPU utilisation. We monitored achieved utilisation to ensure that the GPU’s math units were consistently exercised; in practice, measurements were only retained when sustained utilisation fell within a high-load band (approximately 80–100\%), avoiding misleading results caused by intermittent stalls or CPU-side bottlenecks. It is important to emphasise that memory footprint (instantaneous VRAM allocated) does not by itself predict speed: kernels may be compute-bound or bandwidth-bound, and identical memory usage can correspond to very different instruction mixes and achieved FLOPs. Consequently, our analysis interprets latency and speedup in the context of observed utilisation rather than memory consumption alone.

\subsection{Software Environment}
Experiments were run on Linux with CUDA~12.6 drivers and PyTorch~2.8.0. We used the cuEquivariance library (v0.4.0) for optimised equivariant tensor operations where indicated, and the e3nn backend (v0.4.4) as the baseline implementation for comparison. Atomistic simulations and property queries were orchestrated through ASE, and the MACE (v0.3.14) models correspond to the configurations specified in Section~\ref{sec:experiments}.

\subsection{Determinism and Reproducibility}

To ensure reproducibility, we seed all relevant libraries such as Python, NumPy, and PyTorch CPU and CUDA to be enable on deterministic algorithms mode, accepting a small throughput penalty in exchange for like-for-like comparisons across runs and backends. For molecular dynamics, initial velocities are drawn with a fixed seed so trajectories remain comparable under different numerical precisions. We document seeds, determinism settings, and build options to allow exact replication of timings and numerics. In practice, the accompanying snippet sets seeds, enforces deterministic kernels, disables autotuning that could change algorithm selection, and --- for GEMM determinism --- requires setting the \texttt{CUBLAS\_WORKSPACE\_CONFIG} environment variable before launching Python.

\begin{lstlisting}[language=Python,caption={Deterministic seeding and backend settings for reproducible benchmarks.}]

def set_deterministic(seed: int = 42) -> None:
    # 1) Seed all relevant RNGs
    random.seed(seed)
    np.random.seed(seed)
    torch.manual_seed(seed)
    torch.cuda.manual_seed(seed)
    torch.cuda.manual_seed_all(seed)

    # 2) Enforce deterministic algorithm choices
    torch.use_deterministic_algorithms(True, warn_only=False)

    # 3) cuDNN: disable autotuner and force deterministic kernels
    torch.backends.cudnn.benchmark = False
    torch.backends.cudnn.deterministic = True

    # 4) (Optional) Disable TF32 to stabilise numerics across GPUs
    torch.backends.cuda.matmul.allow_tf32 = False
    torch.backends.cudnn.allow_tf32 = False

    # 5) cuBLAS determinism (set before Python starts):
    export CUBLAS_WORKSPACE_CONFIG=":4096:8"
    # or ":16:8" on older drivers

set_deterministic(42)
\end{lstlisting}

\subsection{Warm-Up, JIT, and Cache Effects}
All benchmarks include an explicit warm-up phase that is discarded from reported statistics. The first iterations pay one-time costs --- CUDA context initialisation, cuBLAS handle creation and algorithm-selection heuristics, host/device allocator ramp-up, and kernel-launch caching in the driver. In addition, when using cuEquivariance, just-in-time specialisation and packing for segmented-polynomial and tensor-product operators occur on first use for each distinct tensor shape. After these caches are populated and shapes repeat, subsequent iterations reflect steady-state performance with substantially lower variance. Unless otherwise stated, we therefore report means and dispersion measures from post–warm-up iterations only.

\section{Dataset and Pre-trained Models}

\subsection{Scope and Rationale}
The datasets employed in this work were chosen to stress the computational kernels of the MACE architecture rather than to target a particular application domain. In particular, we favored systems that produce dense local interaction graphs so as to drive high GPU utilization. This pragmatic choice ensured that block-level profiling and backend comparisons are conducted under load conditions representative of demanding molecular simulations and kernel optimization, while remaining compact enough for controlled micro-benchmarks.

\subsection{Pre-trained Force Field}
All inference and molecular-dynamics (MD) studies use the \emph{MACE-OFF24(M)} pre-trained model (the 2024 medium variant of MACE-OFF). Relative to the earlier OFF23 series, OFF24(M) extends the layerwise cutoff to $r_{\text{max}} = 6.0\,\text{\AA}$ and incorporates additional configurations from SPICE version~2. The medium setting employs 128 chemical channels with a maximum message equivariance of $L_{\text{max}} = 1$. The corresponding learnable parameter count for OFF24(M) is approximately $9.1 \times 10^{6}$ weights.

\subsection{Benchmark and Evaluation Systems}
\begin{figure}[h]
  \centering

  \begin{subfigure}[t]{0.32\textwidth}
    \centering
    \includegraphics[width=\linewidth]{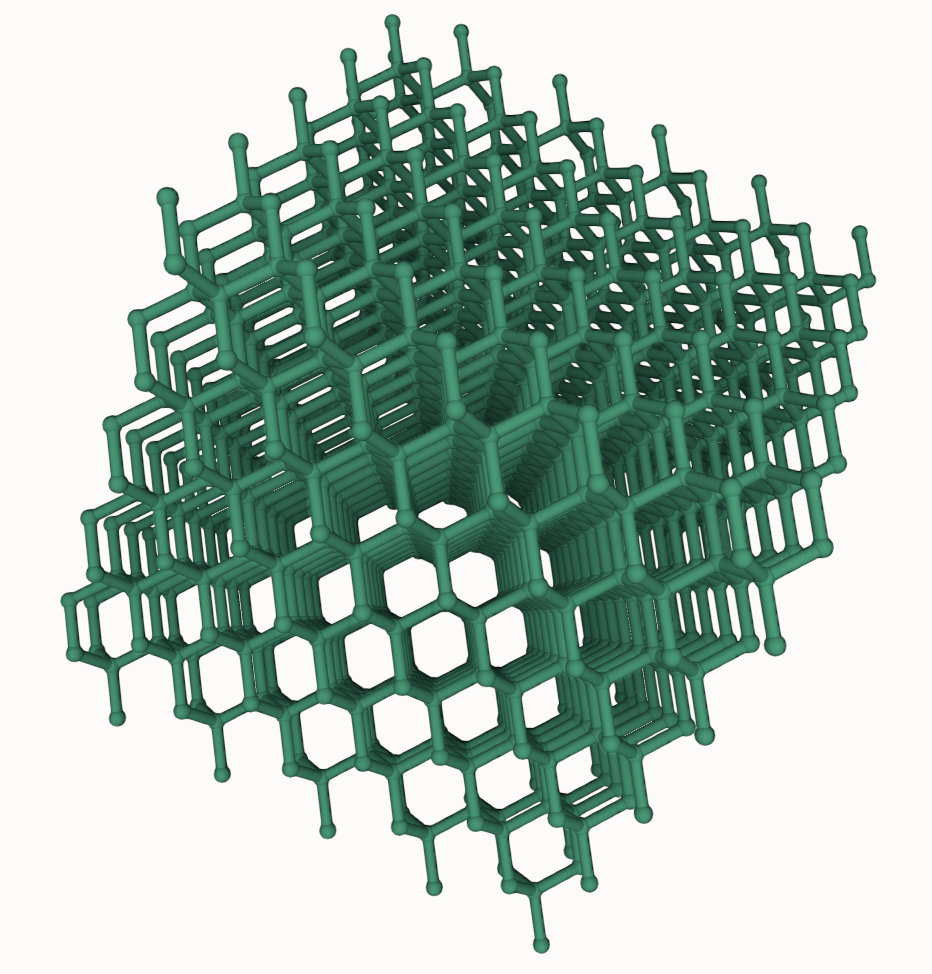}
    \caption{Carbon}
    \label{fig:carbonmol}
  \end{subfigure}\hfill
  \begin{subfigure}[t]{0.32\textwidth}
    \centering
    \includegraphics[width=\linewidth]{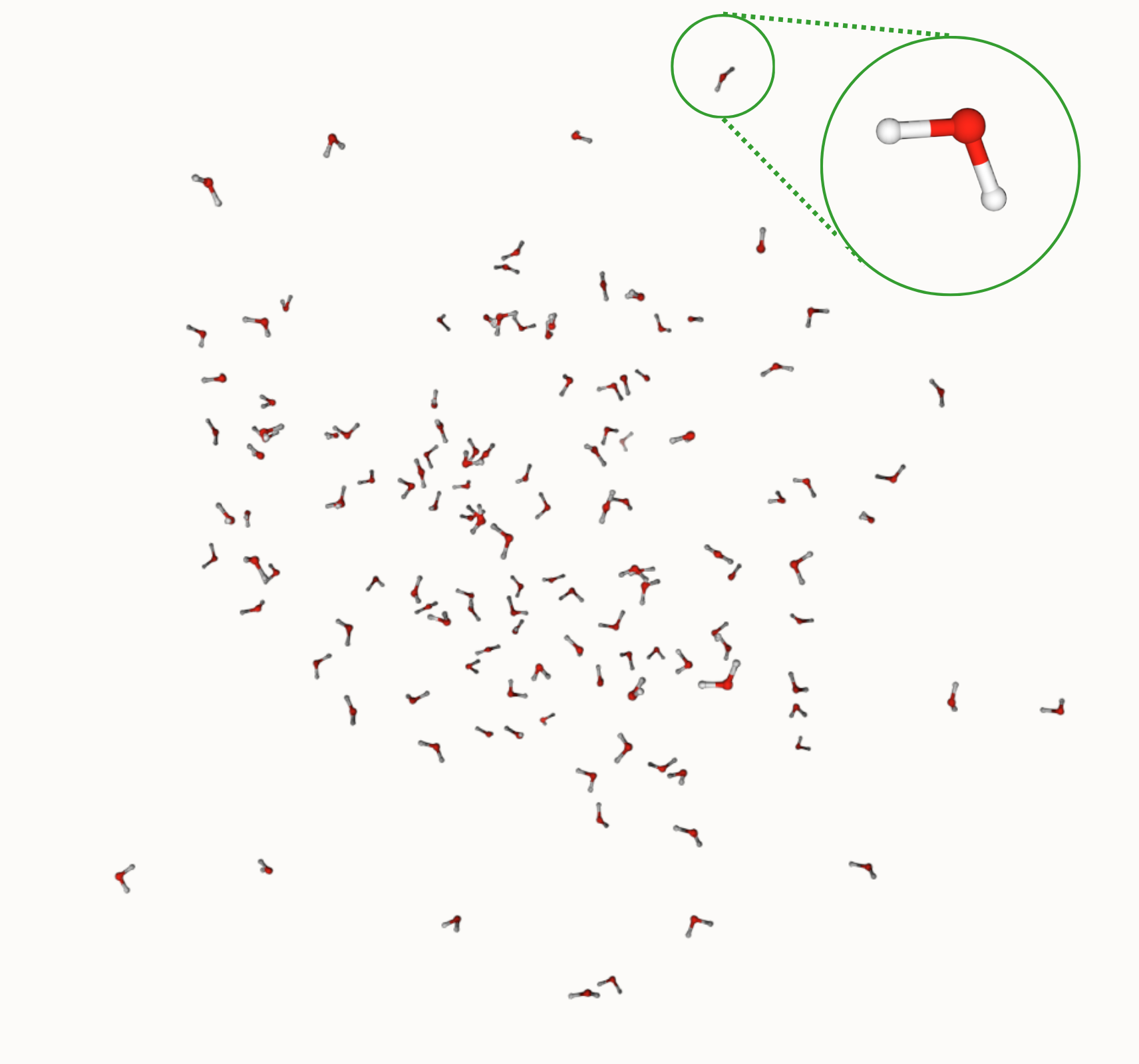}
    \caption{Water box}
    \label{fig:waterbox}
  \end{subfigure}\hfill
  \begin{subfigure}[t]{0.32\textwidth}
    \centering
    \includegraphics[width=\linewidth]{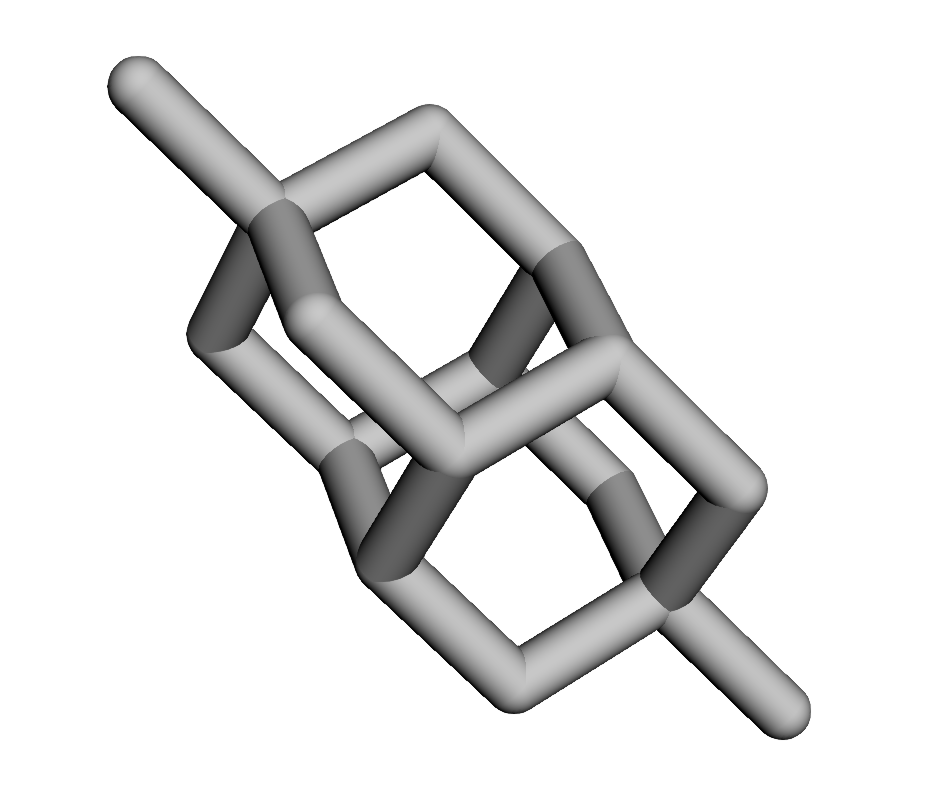}
    \caption{Diamond carbon unit cell}
    \label{fig:diamondcell}
  \end{subfigure}

  \caption{Structures used in this work.}
  \label{fig:structures}
\end{figure}

Three classes of molecular systems are employed. First, a \emph{single-species carbon cluster} (\texttt{carbon.xyz}) is used for kernel and block-level micro-profiling with a cutoff of $r_{\max}=6.0\,\text{\AA}$, which yields a moderately dense local graph (approximately $\sim 8$ neighbours per atom for the sizes considered) and reproducible shapes across repeated runs. Second, a \emph{diamond-carbon supercell} serves to scale atom and edge counts systematically; increasing the replication factor provides a direct lever to load the GPU while preserving chemistry. Third, a \emph{periodic water box} is used for MD accuracy and stability checks because it is sensitive to the model’s short-range interaction fidelity under periodic boundary conditions. Representative structures are shown in Fig.~\ref{fig:structures}.

\subsection{Configuration for Block-Level Experiments}
For controlled ablations and single-block timing, we instantiate a canonical model configuration mirroring the OFF medium architecture while keeping dataset-dependent quantities explicit. Table~\ref{tab:block-config} lists the exact settings used unless otherwise stated:

\begin{table}[h]
\centering
\small
\caption{Canonical configuration used for training or benchmarking individual blocks.}
\label{tab:block-config}
\begin{tabular}{llp{6.7cm}}
\toprule
\textbf{Key} & \textbf{Value} & \textbf{Notes} \\
\midrule
Atomic numbers & dataset-dependent & Elements present in the target system; used to build the $Z$ table and embeddings. \\
Cutoff $r_{\max}$ & $6.0\,\text{\AA}$ & Matches OFF24(M) for inference; exploratory block tests also use $r_{\max}\in[3.0,6.0]\,\text{\AA}$. \\
Hidden irreps & $128\times 0e + 128\times 1o$ & Medium setting; equivariant messages up to $L_{\max}=1$. \\
Correlation order & 3 & Induces up-to 4-body features per layer in the product basis. \\
\# interactions & 2 & Two message-passing layers as in OFF medium. \\
Radial basis & $n_{\text{Bessel}}=8$ & Standard Bessel basis size for radial features. \\
Cutoff envelope & $p=5$ & Polynomial cutoff smoothness. \\
Readout MLP irreps & $16\times 0e$ & Invariant head hidden size. \\
Warm-up / timing & 100 / 100 calls & Discard warm-up for reporting; shape-stable calls thereafter. \\
Batching & single configuration & Throughput reported per configuration to align with ASE/MD usage. \\
\bottomrule
\end{tabular}
\end{table}

\subsection{Architecture Decomposition}
For reference and to enable kernel-level replication, Table~\ref{tab:mace-arch-breakdown} in Appendix~\ref{ap:mab} provides a breakdown of the \texttt{ScaleShiftMACE} components used in this work, including the principal tensor shapes and the number of learnable weights associated with each linear mapping or MLP. This matches the medium-size, two-layer, $L_{\max}=1$ configuration described above and is the basis for the stated $\sim\!9.1$M parameter count.

\section{Experimental Setup}
\label{sec:experiments}

\subsection{Profiling and Per-block Benchmarks}
\label{sec:profiling}

We first profiled end-to-end MACE inference and its constituent blocks to map computational cost, identify bottlenecks, and determine whether execution was predominantly GPU- or CPU-bound. PyTorch’s profiler was configured to record both host and device activities. Traces were subsequently examined with \href{https://perfetto.dev/docs/tracing-101}{Perfetto}, which provided timeline visualisation, kernel clustering, and correlation between host ranges and device kernels. This workflow allowed us to attribute device time to specific MACE submodules and to confirm which operators dominate inference under steady-state conditions.

\paragraph{Instrumentation and set-up.}
All measurements were taken in steady state, following a warm-up of 100 iterations and a timed window of 100 iterations. Unless noted, the batch size was fixed to $B=1$ to reflect ASE-style single-configuration inference; selected experiments varied $B\in{1,4,8}$ to assess batching effects. The model was \emph{MACE-OFF24(M)} with $r_{\max}=6.0,$ $\mathrm{\AA}$, 128 channels, and $L_{\max}=1$. We wrapped major submodules (e.g., \text{interactions[...].conv\_tp}, \text{linear\_up}, product-basis and readout blocks) in profiler record ranges so that the exported trace contained explicit begin/end markers for each block. Perfetto analysis thus reported, per block and per iteration, host-side time (Python/dispatcher) and device-side time (sum of enclosed CUDA kernels), together with launch gaps and any observable host--device synchronisation.

\paragraph{Dataset choice and transferability.}
For micro-profiling we employed compact yet graph-dense systems (e.g., a single-species carbon cluster and a diamond supercell at \(r_{\max}=6.0\,\text{\AA}\)) to “load” tensor-product and product-basis operators without incurring the overheads of large dataset orchestration. The working hypothesis is that relative cost rankings observed on these controlled systems transfer to larger instances with similar local coordination.

\paragraph{Attribution and metrics.}
Let \(\mathcal{B}\) denote a profiled block. The average device time per call is
\[
\bar{T}^{\mathrm{dev}}(\mathcal{B}) \;=\; \frac{1}{N_{\mathrm{timed}}}\sum_{i=1}^{N_{\mathrm{timed}}}\!\!\Big(\sum_{k \in K_i(\mathcal{B})} t_k\Big),
\]
where \(K_i(\mathcal{B})\) is the set of CUDA kernels attributed to \(\mathcal{B}\) on iteration \(i\), and \(t_k\) their durations. Host time \(\bar{T}^{\mathrm{host}}(\mathcal{B})\) is defined analogously from CPU events inside the same range. We classify runs as GPU-bound when device time accounts for \(\geq 80\%\) of end-to-end iteration time; otherwise they are considered CPU/dispatch-bound.

\paragraph{Per-block low-precision experiments.}
To isolate numerical and performance effects of reduced precision, we exercised individual blocks with controlled inputs that preserve realistic tensor shapes. Two complementary procedures were used. First, we intercepted and cached the runtime tensors entering a target block during an FP64 forward pass, then replayed the block in alternative dtypes (FP32, BF16, FP16) to ensure identical shapes and sparsity patterns. Second, for stress tests we generated synthetic but shape-consistent dummy tensors (matching irreps and channel counts) to probe pure operator throughput. For each dtype, we measured latency and computed numerical discrepancies against the FP64 baseline with the maximum absolute error and the maximum relative error with a small stabiliser \(\epsilon\) to avoid division by near-zero references. For both procedures, a simple dummy loss function was applied to ensure well-defined backward passes during gradient checks. The loss was chosen as a mean-square surrogate,
\begin{equation}
    \mathcal{L}(x) \;=\; \bigl(x^{2}\bigr).\mathrm{sum}() \;+\; 0.1,
\end{equation}
where $x$ denotes the block output tensor. This form guarantees non-zero gradients across all elements and avoids trivial cancellation. 

\subsection{Mixed-Precision Inference and Traning}
\label{sec:mixedprecision-inference}
Building on the block-level profiling, we evaluated end-to-end inference under mixed-precision policies to quantify speed–accuracy trade-offs and backend effects. We compared the cuEquivariance (cuEq) and e3nn implementations across FP64, FP32, and autocast-enabled FP16/BF16 settings. In addition to changing the model’s default dtype, we tested selective casting in which only linear layers were run in FP16/BF16 while the remainder of the network stayed in FP32/FP64. Test structures were constructed as diamond supercells with a configurable replication factor to control atom/edge counts. Timing used CUDA events after an explicit warm-up; energies, per-atom energies, and forces were queried through the \texttt{MACECalculator} interface and synchronised at the device boundary. For each run, we recorded the backend, dtype policy, mean and dispersion of per-iteration latency, and numerical outputs (energies and average force magnitudes).

\paragraph{Toy low-precision training and sanity-check MD.}
To complement the inference study, we reuse the same experimental pipeline to train a pared-down MACE model on a small corpus of solvent molecules. The goal is not to reach state-of-the-art accuracy but to probe the \emph{sensitivity of training} to reduced precision. We consider three regimes: a full-precision FP64 baseline; mixed precision with FP32 and Linear layers in FP16/BF16. In the latter two, automatic mixed precision (AMP) is employed with master weights kept in FP32 and dynamic gradient scaling enabled to mitigate underflow. We validate on a held-out set of solvent molecules, reporting standard energy/force errors and monitoring training stability. This controlled setting isolates numerical effects of low precision while keeping data curation and orchestration overheads minimal.

\paragraph{Minimal Langevin MD using the trained model.}
As a downstream check of physical plausibility, we run a short Langevin dynamics simulation with the previous trained model as an ASE calculator. Velocities are initialised from a Maxwell–Boltzmann distribution at 1200\,K, and overall translation and rotation are removed to prevent spurious drift (\texttt{MaxwellBoltzmannDistribution}, \texttt{Stationary}, \texttt{ZeroRotation}). We then integrate with ASE’s \texttt{Langevin} thermostat using a 10\,fs timestep and a friction coefficient of 0.1, driving the system to the target temperature. Over the short trajectory, we monitor temperature convergence, potential-energy stability, and the absence of numerical pathologies (divergence or exploding forces). This sanity check complements the validation metrics by confirming that a model trained under reduced precision yields stable short-time dynamics or not.

\subsection{Molecular Dynamics Stability Protocols}
\label{sec:md-stability}
To assess dynamical robustness under reduced precision, we ran fully-coupled MD using the MACE-off (24) force field with the calculator attached in inference mode. It differs from the previous experiments as we have a proper set of weights on a state-of-the-art model trained in FP64. Simulations employed the ASE \texttt{NPT} integrator with a 1\,fs timestep, target temperature set by the input parameter, and external pressure at 1.013\,bar. Thermostat and barostat time scales followed the code defaults used here (temperature time constant 100\,fs; pressure factor defined as \texttt{ptime}$^{2}$ times an effective bulk modulus of 2.2\,GPa), with isotropic stress enforced. The initial configuration was a periodic water, with the cell made upper-triangular and the calculator bound prior to integration. Each precision arm (FP64, FP32, FP32+FP16, FP32+BF16 via layer-level casting) produced a single trajectory with identical initial conditions and synchronised output cadence. During the run we logged wall time, simulation time, temperature, density, pressure, potential energy, mean-squared displacement, centre-of-mass drift, and the Voigt components of the stress tensor at fixed intervals, and saved atomic coordinates periodically. Post hoc, we summarised stability by inspecting energy and temperature time series (relative to the FP64 reference sampled on the same time grid) and by verifying the absence of divergence or pathological drift.

\section{Chapter Summary}
\label{sec:chap-summary}

We fixed hardware to RTX~2080~Ti workstations and focused on software/method effects. Timings were taken in steady state after a 100-iteration warm-up over a 100-iteration window; default batch was \(B{=}1\) (ASE-like single-configuration inference) with spot checks at \(B\in\{1,4,8\}\). Runs sustained 80–100\% GPU utilisation, and latency/speedup were interpreted against compute/bandwidth regimes rather than VRAM limits. The stack was Linux, CUDA~12.6, PyTorch~2.8.0, MACE~v0.3.14, ASE, e3nn~v0.4.4 (baseline), and cuEquivariance~v0.4.0 (optimised). Reproducibility used strict seeding, deterministic kernels, TF32 disabled where appropriate, and cuBLAS workspace control. Datasets stressed tensor products: a carbon cluster (micro-profiling), a diamond-carbon supercell (scaling), and a periodic water box (MD stability). All inference/MD used the pre-trained \emph{MACE-OFF24(M)} (\(r_{\max}{=}6.0~\text{\AA}\), 128 channels, \(L_{\max}{=}1\)). Major submodules were bracketed with profiler ranges, and Perfetto attributed host/device time and launch gaps per block to enable like-for-like backend comparisons.

Our aim is to provide readers with a high-level, faithful computational map of MACE and to determine when lower-precision arithmetic meaningfully accelerates inference with a certain sense of stability. The first experiment (next chapter) tests whether MACE is GPU- or CPU-bound and whether our theoretical profiling of block-level cost aligns with practice. To make any further benchmarking credible --- and to expose the benefits of cuEquivariance --- the setup must be general, shape-stable, and consistently drive high GPU utilisation. The following section details the experimental design, reports results, and discusses their implications for precision settings, kernel bottlenecks, and backend choice.

\chapter{Results and Discussion}
\label{chapter:results_and_discussion}

\ifpdf
    \graphicspath{{Chapter3/Figs/Raster/}{Chapter3/Figs/PDF/}{Chapter3/Figs/}}
\else
    \graphicspath{{Chapter3/Figs/Vector/}{Chapter3/Figs/}}
\fi

\textit{Building on the background, hardware, and dataset choices established earlier, this chapter characterises the computational behaviour of MACE at inference time and identifies the levers for acceleration. Our goals are to (i) map end-to-end and per-block costs with PyTorch’s profiler and Perfetto, (ii) quantify the speedup of the cuEquivariance backend relative to the original e3nn implementation, and (iii) assess the accuracy–performance trade-offs of mixed-precision policies in both inference and short MD runs.}

\section{Profile of MACE}
\label{sec:pom}

\subsection{E3NN vs.\ cuEquivariance Speedup}
\label{sec:evcs}

\begin{figure}[h]
    \centering
    \includegraphics[width=1\linewidth]{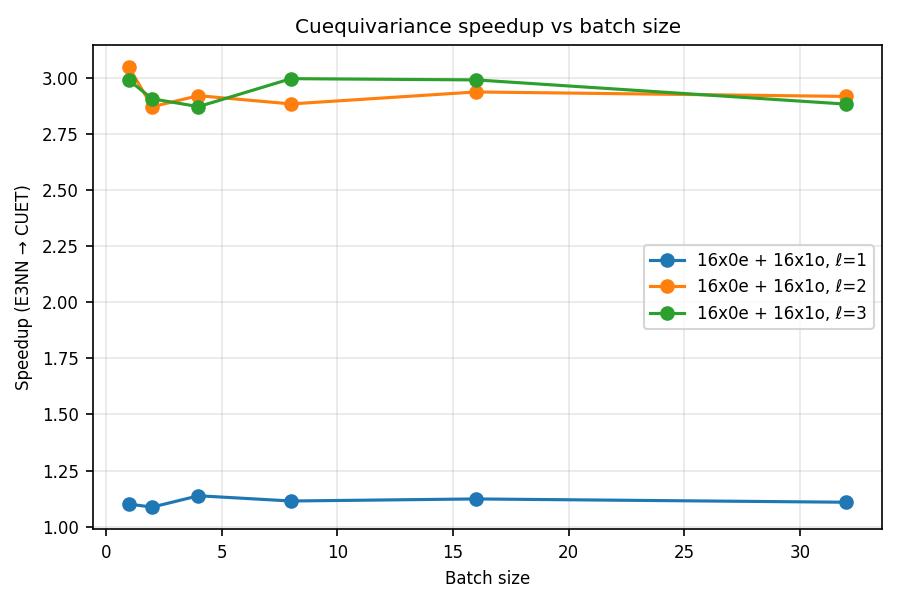}
    \caption[cuEquivariance vs E3NN speedup vs batch size]{\textbf{cuEquivariance yields near-constant $\sim$3$\times$ speedup for higher-order blocks.}
    Speedup (e3nn $\rightarrow$ cuEq) vs.\ batch size for three angular orders. The batch size does not have a high impact.
    For $\ell\in\{2,3\}$, speedup is $\approx$2.9--3.0$\times$ with weak dependence on batch size; for $\ell=1$ it is $\sim$1.1$\times$.
    Each point summarizes 100 forward passes under the same setup as Appendix~\ref{appendix:setup}.}
    \label{fig:bspeedup}
\end{figure}

\begin{figure}[h]
    \centering
    \includegraphics[width=1\linewidth]{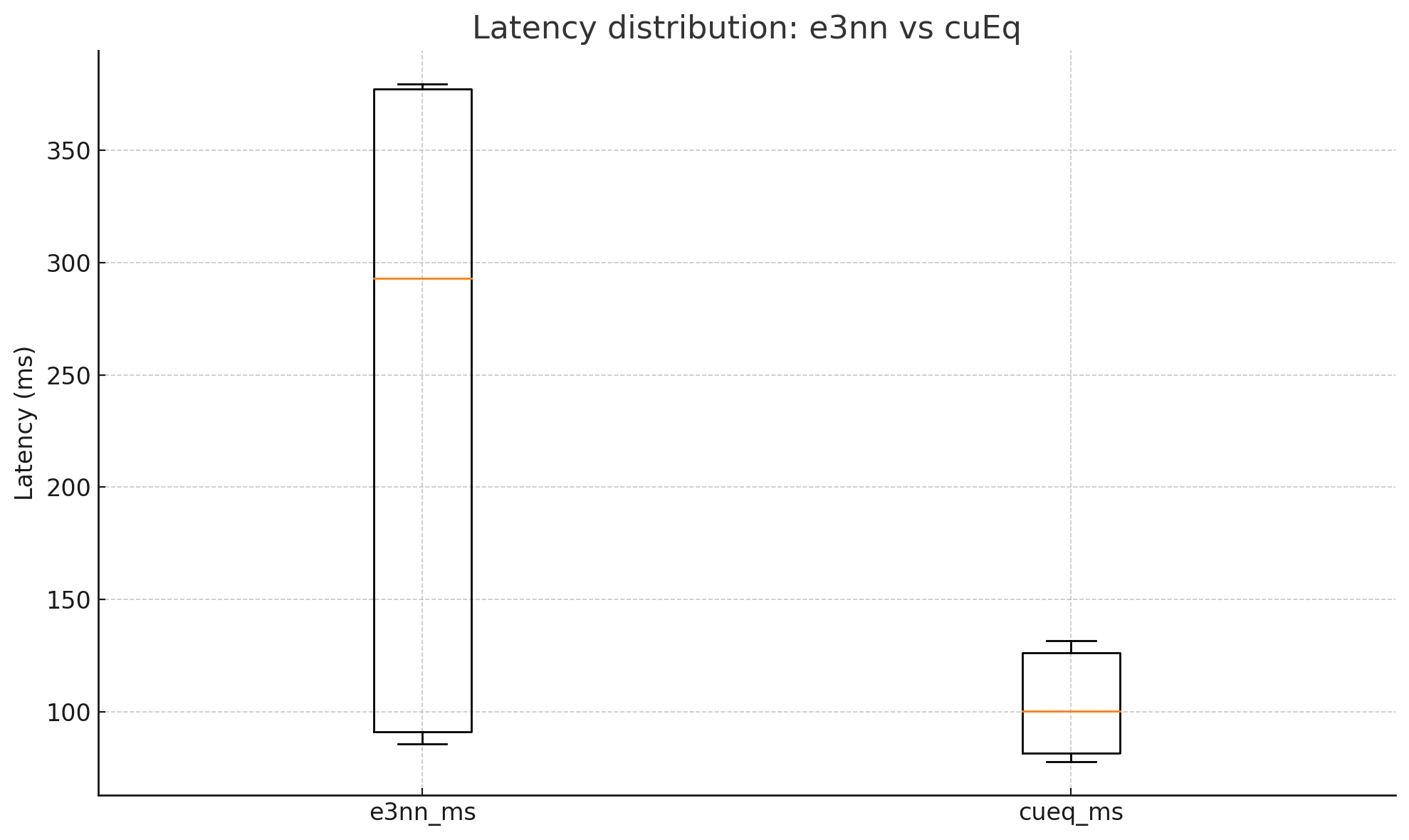}
    \caption[Latency distributions between cuEq and E3NN]{\textbf{Latency distributions: cuEq is faster and less variable.}
    Boxplots of per-step latency over 100 runs.
    Medians around $\sim$100\,ms (cuEq) vs.\ $\sim$300\,ms (e3nn) illustrate the \(\approx3\times\) gap and narrower dispersion with cuEq.}
    \label{fig:latdist}
\end{figure}

Switching MACE from the e3nn backend to cuEquivariance (cuEq) reduces end-to-end \emph{inference latency} by \textbf{2.98$\times$} on our carbon benchmark: \textbf{122.9\,ms} (cuEq) vs.\ \textbf{366.1\,ms} (e3nn), averaged over 100 forward passes.
The distribution of latencies is also markedly tighter under cuEq (Fig.~\ref{fig:latdist}), indicating more stable step times.
Across batch sizes, speedup is \textbf{$\approx$3$\times$} for higher angular orders ($\ell\!\in\!\{2,3\}$) and only \textbf{$\sim$1.1$\times$} for $\ell=1$ (Fig.~\ref{fig:bspeedup}).

Practically, a near-\(3\times\) reduction in per-step latency translates linearly to shorter wall-clock time for MD runs and screening workloads using this model configuration.
The larger gains for $\ell\ge 2$ align with prior reports that cuEq accelerates tensor products and related equivariant primitives via fused and custom CUDA kernels with benefits increasing as angular order (and thus tensor-product cost) grows.
The muted improvement at $\ell=1$ suggests those settings are closer to memory/launch-overhead limits where cuEq’s kernel fusion yields less benefit.
Overall, these results support using cuEq as the default backend for production inference with mid/high-order MACE blocks, reserving e3nn only where later sections show it offers numerical advantages.

\subsection{cuEquivariance inference profiling: warm-up vs.\ steady state}
\label{sec:cueq-prof}

\begin{figure}[h]
    \centering
    \includegraphics[width=\linewidth]{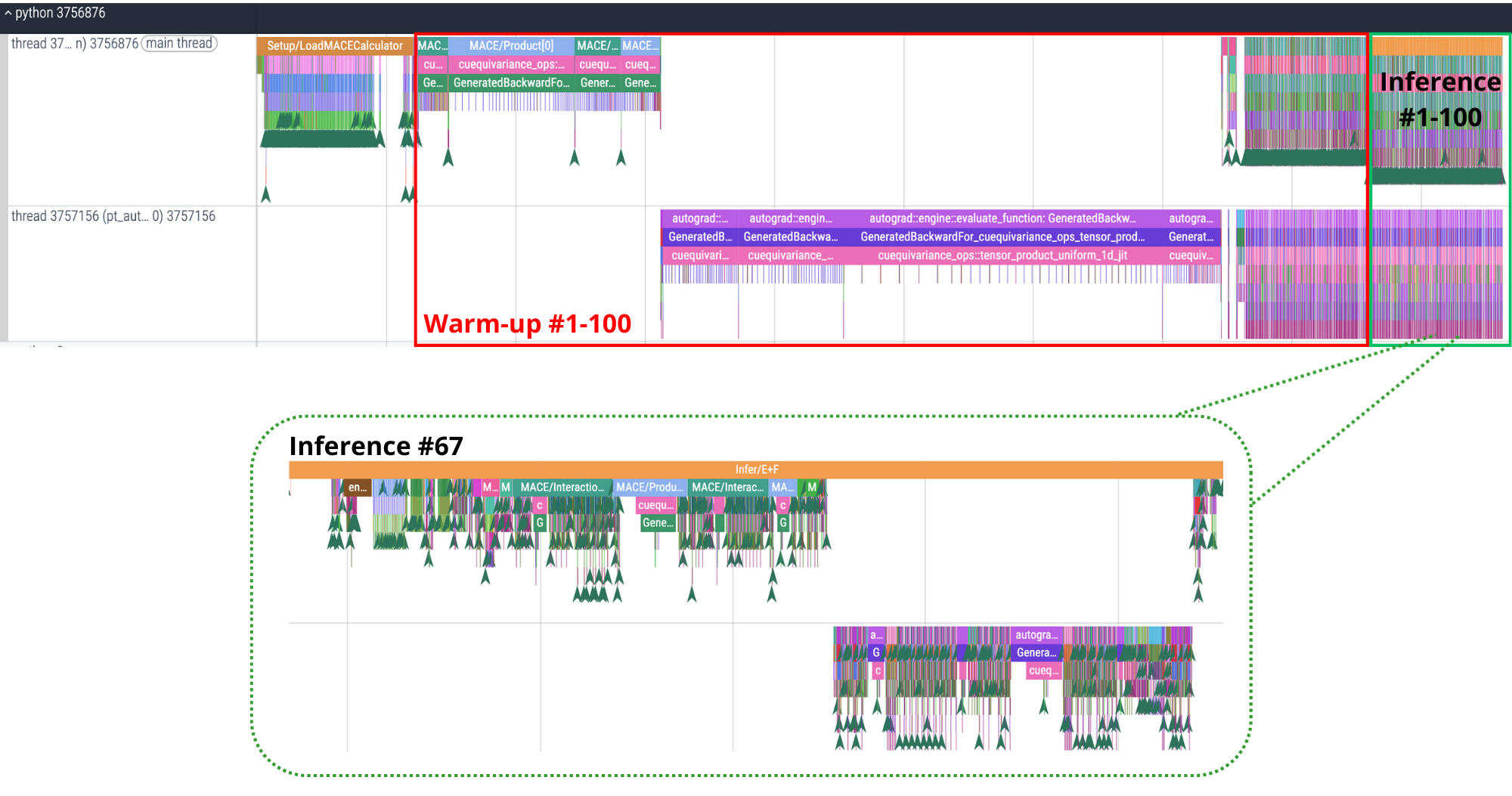}
    \caption[MACE's inference profiling mapping]{\textbf{Warm-up amortizes one-off initialization; steady-state latency is stable.}
    PyTorch trace of 100 warm-ups followed by 100 inference calls for MACE with cuEquivariance.
    Top lane: forward pass; bottom lane: autograd backward used to obtain forces.
    The initial segment (setup + first call) takes $\sim$30\,s, after which step times stabilize around $\sim$48\,ms (zoom: a representative inference).}
    \label{fig:warmup}
\end{figure}

The first end-to-end call (load + initial forward/backward for $E{+}F$) incurs a large one-off cost of \textbf{$\sim$30\,s} as you can see in the red box. After a short warm-up (100 calls), step times converge to a stable \textbf{$\sim$48\,ms} per inference, with low dispersion across repeated runs (Fig.~\ref{fig:warmup}) as shown in the green circled zoom.

The long first call reflects backend initialization and code generation/caching for cuEquivariance operators, memory pool growth, and autograd graph setup. Since this overhead is \emph{amortized}, benchmarks and production MD should exclude first-call timing, include a brief warm-up with the same tensor shapes as the target workload, and keep inference workers long-lived to retain cached kernels.
The stable $\sim$48\,ms steady-state is the relevant figure for wall-clock planning and aligns with the latency reductions reported in \S\ref{sec:evcs}. Note that changing angular order, species set, or batch/shape may trigger new specializations and transient spikes; fixing shapes or pre-touching common configurations mitigates this.

\subsubsection{Profiling overview }
\label{sec:profiling-overview}

Profiling shows a counterintuitive picture: even with high GPU utilisation, end-to-end time is often limited by \emph{pipeline overhead} --- synchronisations, dtype/device casts and copies, and many small kernel launches --- rather than pure math. The heavy math remains the cuEquivariance tensor products and GEMM-style ops, but a large share of wall time is lost outside those kernels. Accuracy is unaffected by pipeline fixes; it is only at risk when we change numerics (e.g., BF16/TF32) or the math itself.

Summing the block-level ranges yields an average \emph{forward} wall-clock of \(\approx 31.9~\mathrm{ms/step}\).
Of this, \(\approx 13.8~\mathrm{ms}\) is CUDA kernel execution and \(\approx 18.1~\mathrm{ms}\) is host-side overhead (launch/dispatch, indexing, layout ops).
Synchronization/idling is small (\(\sim 1.1~\mathrm{ms/step}\)), primarily from \texttt{cudaStreamSynchronize} and host-to-device \texttt{memcpy}.
Independently, operator-level self-times report \(3.331~\mathrm{s}\) Self CPU and \(1.299~\mathrm{s}\) Self CUDA across the 100 profiled steps, indicating a larger CPU fraction --- differences from wall time reflect overlap and profiler accounting (Table~\ref{tab:profiling_overview}).

A substantial fraction of time is spent in host–device syncs and $aten::to/\_to\_copy/copy\_$ transfers; cuEq TP/GEMM kernels account for the main compute but are not the only bottleneck. Therefore, optimising tensor-product kernels is necessary but insufficient; reducing syncs, unifying dtype/device, and cutting kernel-launch count can yield accuracy-neutral speedups of similar magnitude.
A safe way, would be removing \texttt{.item()} hot paths, lower log frequency, pin host memory, use non-blocking H2D moves, cast once in \texttt{collate}, consider \texttt{torch.compile}/CUDA Graphs. However, the risk is that lower precision (BF16/TF32), low-precision accumulators/weights, structured sparsity --- require validation (MAE/RMSE, force cosine, MD stability).

The ratios indicate the workload is \textbf{dispatcher/launch-bound more than math-bound} (CPU/launch bounded rather than GPU bounded): many short kernels and Python/ATen dispatch dominate step time.
Effective accelerations therefore should:
(i) \emph{reduce kernel count/fuse ops} (cuEquivariance operators, §\ref{sec:evcs}); 
(ii) \emph{increase work per launch} (larger batch or micro-batching; shape bucketing);
(iii) \emph{cut Python overhead} (\texttt{torch.compile}, \texttt{torch.inference\_mode}, moving hot paths to fused CUDA kernels);
(iv) \emph{stabilise shapes} to enable persistent caching and CUDA Graph capture.
These implications align with the near-\(3\times\) cuEq speedups observed in §\ref{sec:evcs}, especially at higher \(\ell\) where tensor-product fusion pays off.

\begin{table}[h]
\centering
\begin{tabular}{l|l }
\toprule
\textbf{Metric} & \textbf{Value}  \\
\midrule
Wall-clock per step & \(\approx 31.9~\mathrm{ms}\) \\
CUDA kernel time (\%) & \(\approx 43\%\) \(\sim 13.8~\mathrm{ms}\) of \(\sim 31.9~\mathrm{ms}\) \\
CPU-side overhead (\%) & \(\approx 57\%\) \\
Sync / idle & \(\approx 1.1~\mathrm{ms/step}\) \\
Self times (100 steps) & CPU: \(3.331~\mathrm{s}\), CUDA: \(1.299~\mathrm{s}\) \\
Peak GPU memory & forward-only, \(O(\mathrm{GB})\)  \\
Peak CPU memory & \(\lesssim 0.1~\mathrm{GB}\)\\
\bottomrule
\end{tabular}
\caption[End-to-end forward-only breakdown of MACE profiling results]{End-to-end forward-only breakdown (batch \(=1\), 100 profiled steps). The decomposition indicates a launch/dispatch-dominated regime, suggesting operator fusion, batching, and graph capture as the primary levers for speedup.}
\label{tab:profiling_overview}
\end{table}

This analysis suggests two distinct routes to acceleration: 

\begin{enumerate} 
    \item \textbf{Systems-level optimisation (no accuracy trade-off).} Reduce host overhead and launch latency by engineering the input/compute pipeline: fewer kernel launches, better stream concurrency, careful synchronisation, epilogue fusion, and process/thread parallelism. While this should lower inference time without affecting accuracy, it is cumbersome, trial-heavy, and largely outside the core ML scope. 
    \item \textbf{Model/block and precision optimisation (potential accuracy trade-off).} Target the computational hot spots inside MACE --- especially tensor-product blocks --- and use lower precision on the cuEquivariance backend to make inference cheaper. This path promises practical speedups but requires explicit validation of numerical stability and accuracy. 
\end{enumerate}

Given the scope of this work, we prioritise the second route: hands-on development to characterise the impact of reduced precision in cuEquivariance and to quantify stability/accuracy effects, with systems-level tuning left as future engineering work.

\subsubsection{Block-Level Cost}

\begin{figure}[H]
  \centering
  \begin{subfigure}{0.48\textwidth}
    \centering
    \includegraphics[width=\linewidth]{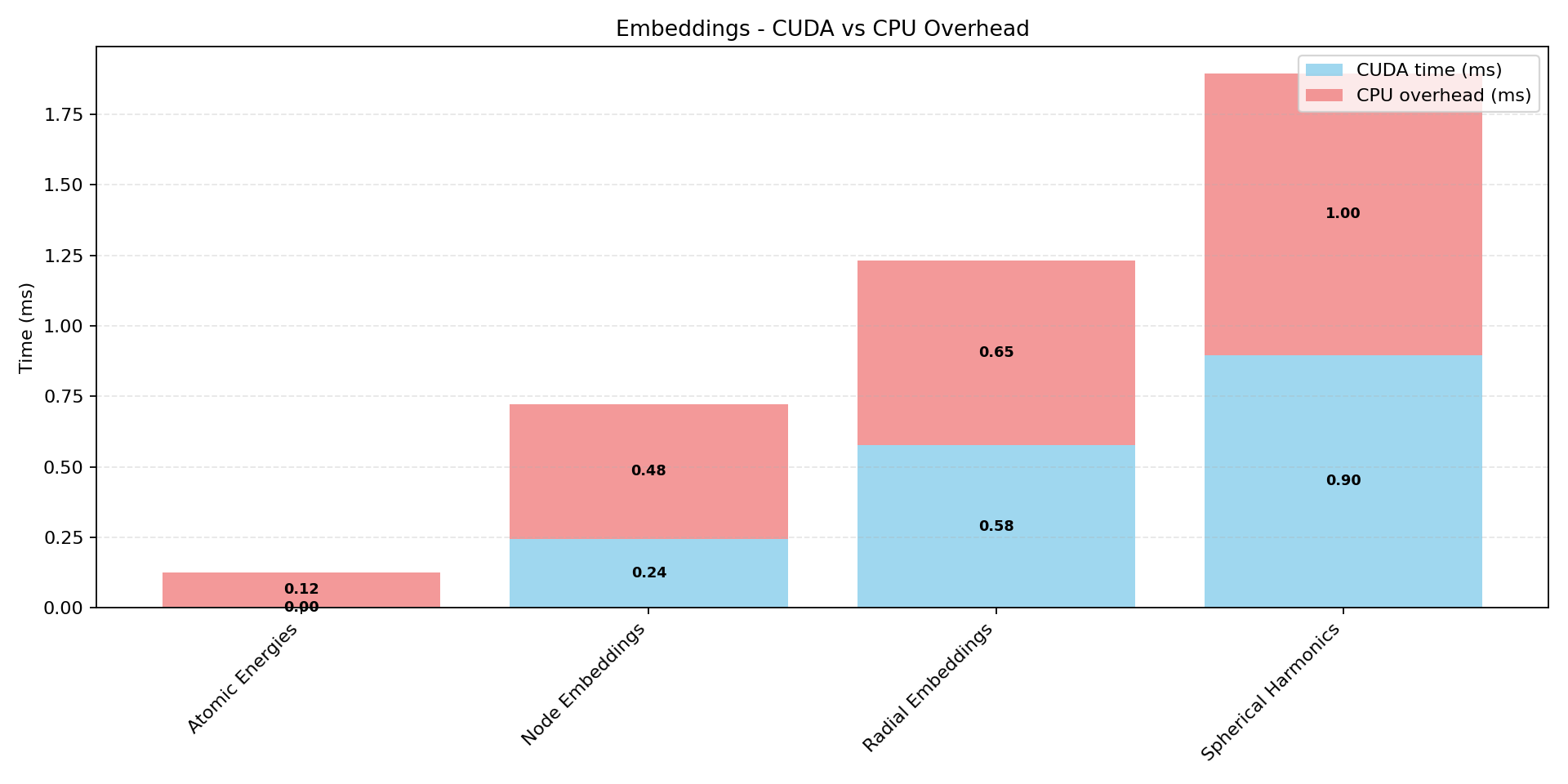}
    \caption{\textbf{Embeddings are moderate and launch-heavy.} Spherical harmonics and radial pieces incur many short kernels and layout ops.}
    \label{fig:mace_embeddings_cuda_cpu}
  \end{subfigure}\hfill
  \begin{subfigure}{0.48\textwidth}
    \centering
    \includegraphics[width=\linewidth]{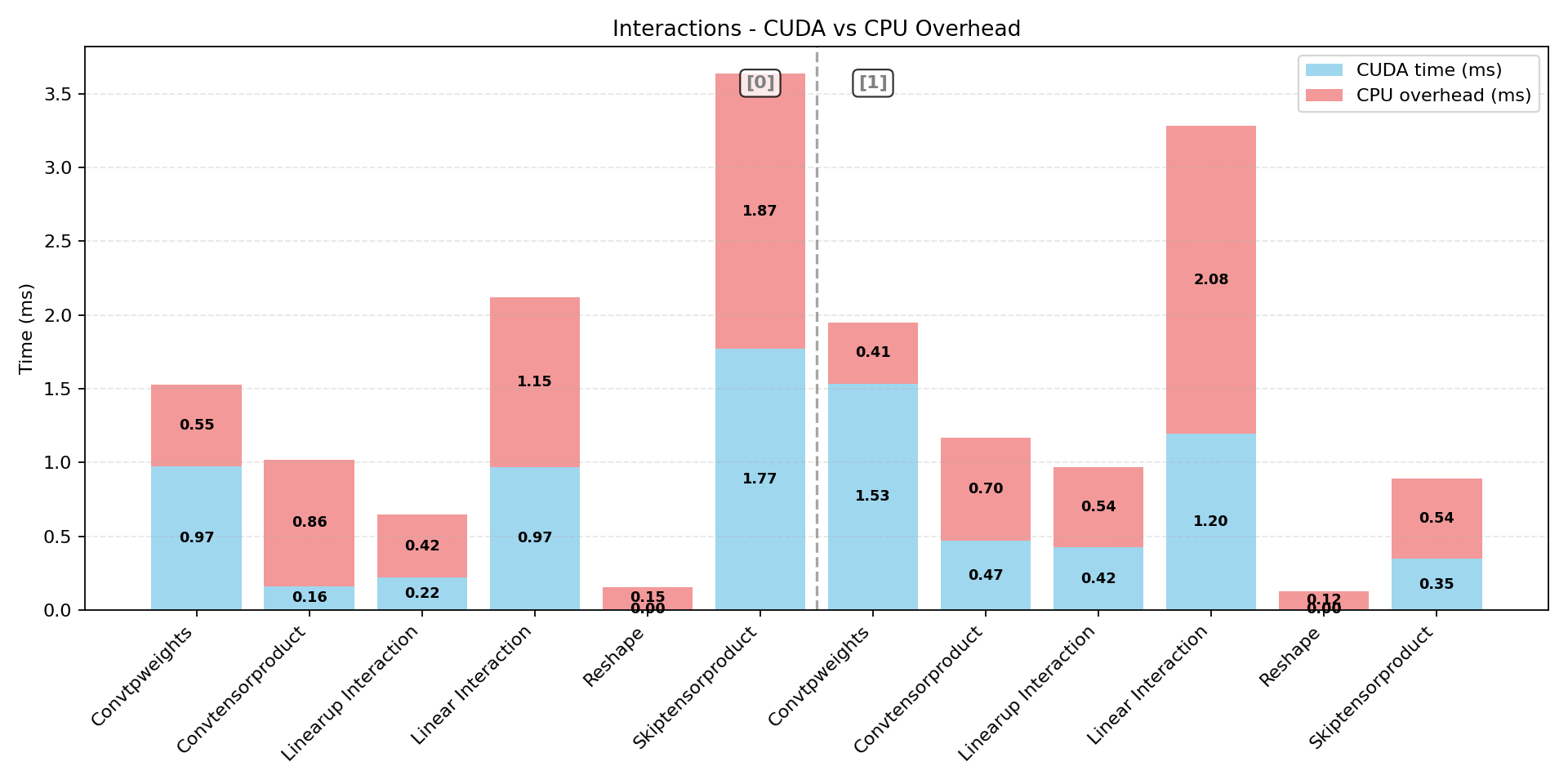}
    \caption{\textbf{Interaction stack is dominated by Linear Interaction and TP chains.} CUDA math is substantial, but CPU dispatch remains larger.}
    \label{fig:mace_interaction_cuda_cpu}
  \end{subfigure}


  \begin{subfigure}{0.48\textwidth}
    \centering
    \includegraphics[width=\linewidth]{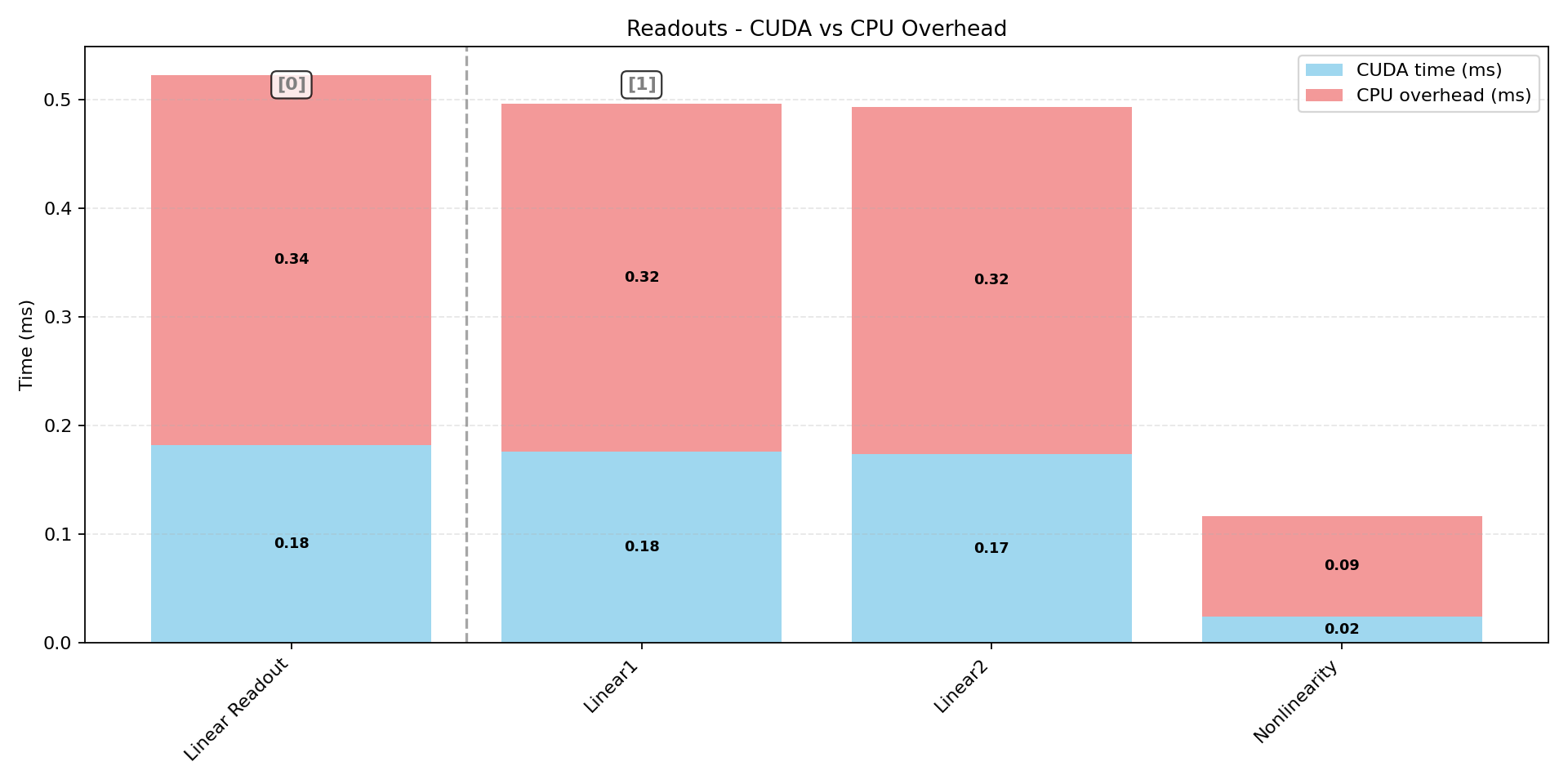}
    \caption{\textbf{Readouts are small but CPU-skewed.} Multiple tiny GEMMs and activations make launch overhead a sizable fraction.}
    \label{fig:mace_readout_cuda_cpu}
  \end{subfigure}\hfill
  \begin{subfigure}{0.48\textwidth}
    \centering
    \includegraphics[width=\linewidth]{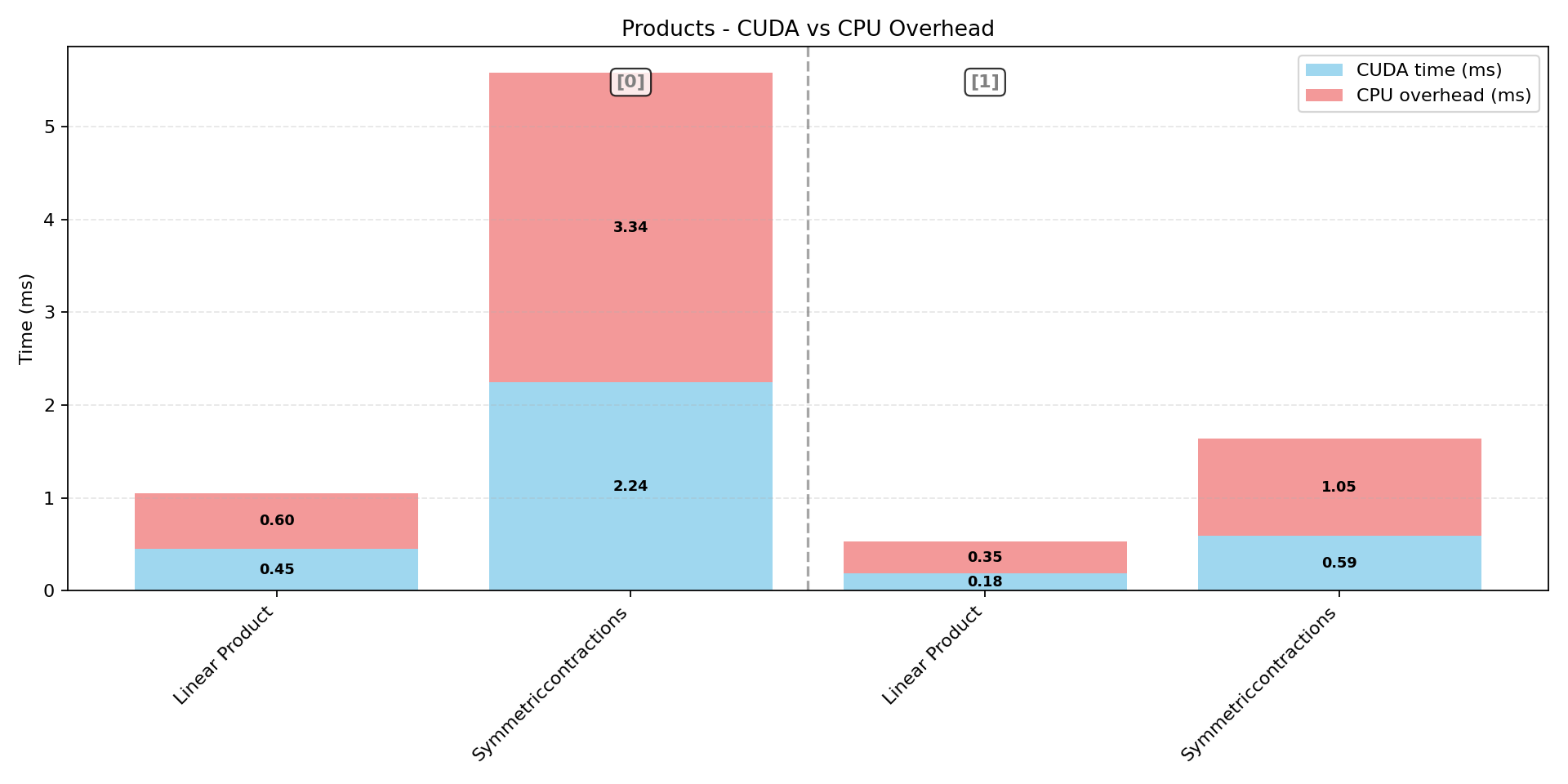}
    \caption{\textbf{Product[0]—SymmetricContractions is the top hotspot.} Highest wall time with large host overhead and FP64 \texttt{dgemm} in the CUDA portion.}
    \label{fig:mace_product_cuda_cpu}
  \end{subfigure}


  \begin{subfigure}{0.48\textwidth}
    \centering
    \includegraphics[width=\linewidth]{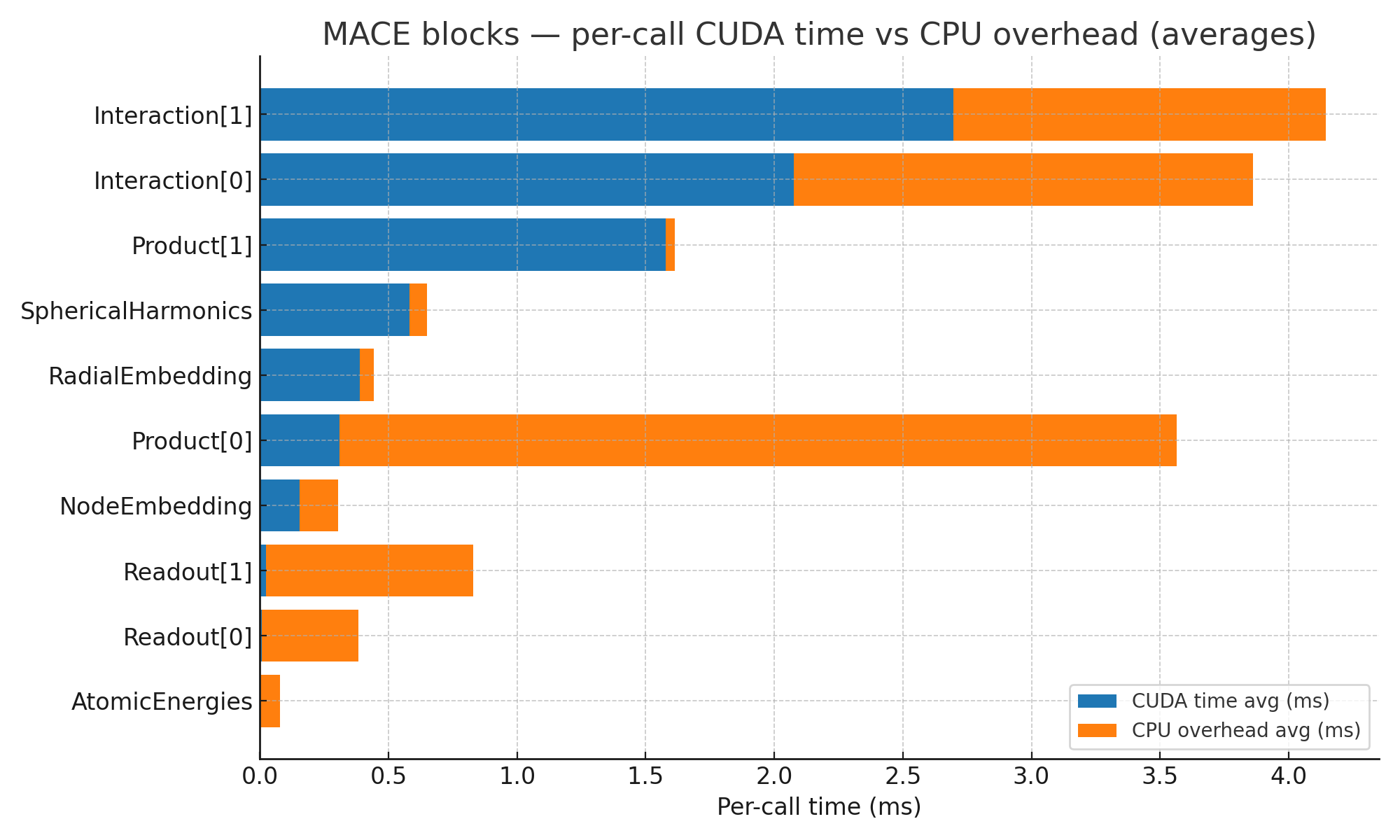}
    \caption{\textbf{Summary across mother blocks.} The CPU (orange) share exceeds the CUDA (blue) share in most blocks. Confirming that the interactions, products and spherical harmonics are CUDA heavy.}
    \label{fig:mace_block_cuda_cpu_overhead}
  \end{subfigure}

  \caption[Block-level profiling at FP64]{\textbf{Block-level profiling at FP64.} Panels show CUDA (blue) vs.\ CPU (pink or orange) time per block during forward inference. The question addressed is which blocks dominate wall time and whether the workload is compute- or launch-bounded. Original profiler outputs are provided in Appendix~\ref{appendix:result_profiling_inference}.}
  \label{fig:four-panel}
\end{figure}

Decomposing the forward pass by MACE blocks shows that computation is dominated by the \emph{Product} and \emph{Interaction} stacks, with large host-side overhead across both as seen in Figure~\ref{fig:mace_interaction_cuda_cpu}. The single largest contributor is \textbf{Product[0] --- SymmetricContractions} at \(5.58~\mathrm{ms/step}\) (\(2.24\) CUDA \(+\) \(3.34\) CPU). A second contraction stage, \textbf{Product[1]}, contributes \(1.64~\mathrm{ms}\) (\(0.59\) CUDA \(+\) \(1.05\) CPU). Within \emph{Interaction[0]}, \texttt{SkipTensorProduct} accounts for \(3.64~\mathrm{ms}\) (\(1.77\) CUDA \(+\) \(1.87\) CPU) and \texttt{Linear Interaction} adds \(2.12~\mathrm{ms}\) (\(0.97\) CUDA \(+\) \(1.15\) CPU); \texttt{ConvTPWeights} and \texttt{ConvTensorProduct} contribute \(1.52~\mathrm{ms}\) and \(1.02~\mathrm{ms}\), respectively. In \emph{Interaction[1]}, \texttt{Linear Interaction} is hottest at \(3.28~\mathrm{ms}\) (\(1.20\) CUDA \(+\) \(2.08\) CPU) with additional cost from \texttt{ConvTPWeights} (\(1.94~\mathrm{ms}\)), \texttt{ConvTensorProduct} (\(1.17~\mathrm{ms}\)), and \texttt{LinearUp}/\texttt{SkipTensorProduct} (each \(\lesssim 1~\mathrm{ms}\)). The \emph{Embedding} stack is moderate --- \textbf{Spherical Harmonics} \(1.90~\mathrm{ms}\), \textbf{Radial Embeddings} \(1.23~\mathrm{ms}\), \textbf{Node Embeddings} \(0.72~\mathrm{ms}\) --- and \emph{Readouts} are small but CPU-heavy (each \(\approx 0.5~\mathrm{ms}\)). 

Across panels the blue CUDA bars correspond to cuEquivariance tensor products and GEMM-style matmuls (\texttt{aten::mm/bmm/matmul}, cuBLAS \texttt{dgemm}), whereas the pink bars arise from Python/ATen dispatch, indexing/gather-scatter, and frequent layout/elementwise epilogues. The net effect is a \emph{massively CPU/launch-bounded} profile at FP64.

The ranking identifies \textbf{SymmetricContractions[0]} as the primary target for acceleration, followed by \textbf{Linear Interaction} and the \textbf{(Skip/Conv)TensorProduct} chain. Their absolute CUDA times confirm where heavy math resides, but their even larger host components show that many short kernels and dispatcher work dominate wall time. This explains why the cuEquivariance backend, delivers the near-\(3\times\) end-to-end gains reported earlier: it reduces both math cost and launch pressure. At the same time, the embedding and readout stacks present secondary opportunities where fusing segmented-polynomial pieces and minimizing layout churn can remove dozens of tiny launches for accuracy-neutral wins. Because FP64 \texttt{dgemm} appears throughout the hottest paths, lowering precision (FP32/TF32/BF16 with appropriate accumulators) should further shrink the blue portions. Overall, these block-level results support a two-pronged strategy used in the rest of this section: keep cuEquivariance and reduce launch count to attack the pink bars, and then evaluate safe precision downgrades to reduce the blue bars --- prioritizing \texttt{SymmetricContractions}, \texttt{Linear Interaction}, \texttt{ConvTensorProduct} where the payoff is greatest.

\section[Proof of Concept]{Proof of Concept - Testing lower precision on MACE's individual blocks}

\label{sec:poc-blocks}

\begin{figure}[h]
  \centering
  \includegraphics[width=\linewidth]{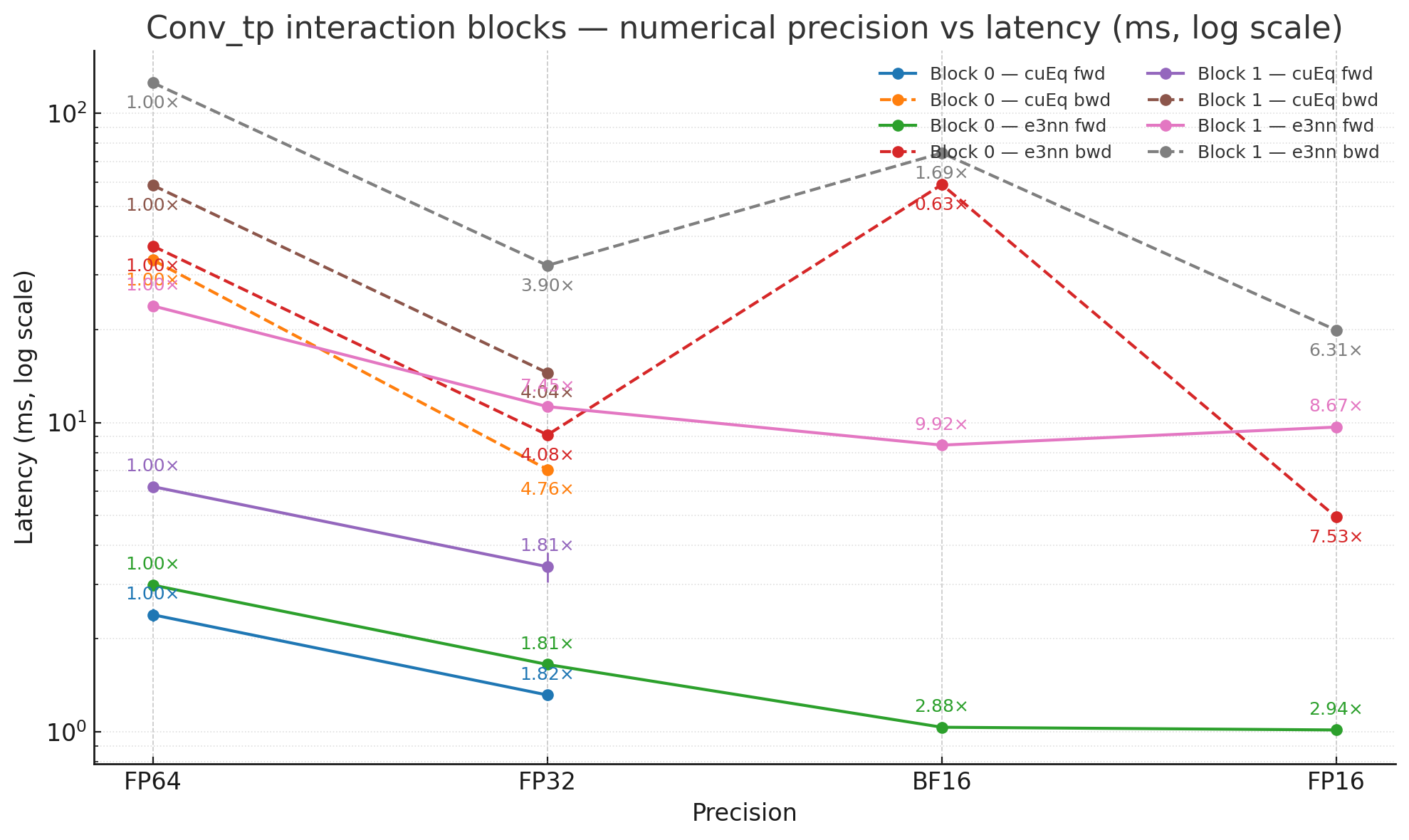}
  \caption[Interaction conv\_tp blocks]{\textbf{Interaction conv\_tp blocks: FP16 on cuEq gives the best forward latency.} Forward times drop steadily from FP64\(\rightarrow\)FP32 and reach their minimum at FP16 under e3nn; backward is less monotonic, especially for e3nn, reflecting different kernel selections and launch sensitivity.}
  \label{fig:conv-tp-latency}
\end{figure}

\begin{figure}[h]
  \centering
  \includegraphics[width=\linewidth]{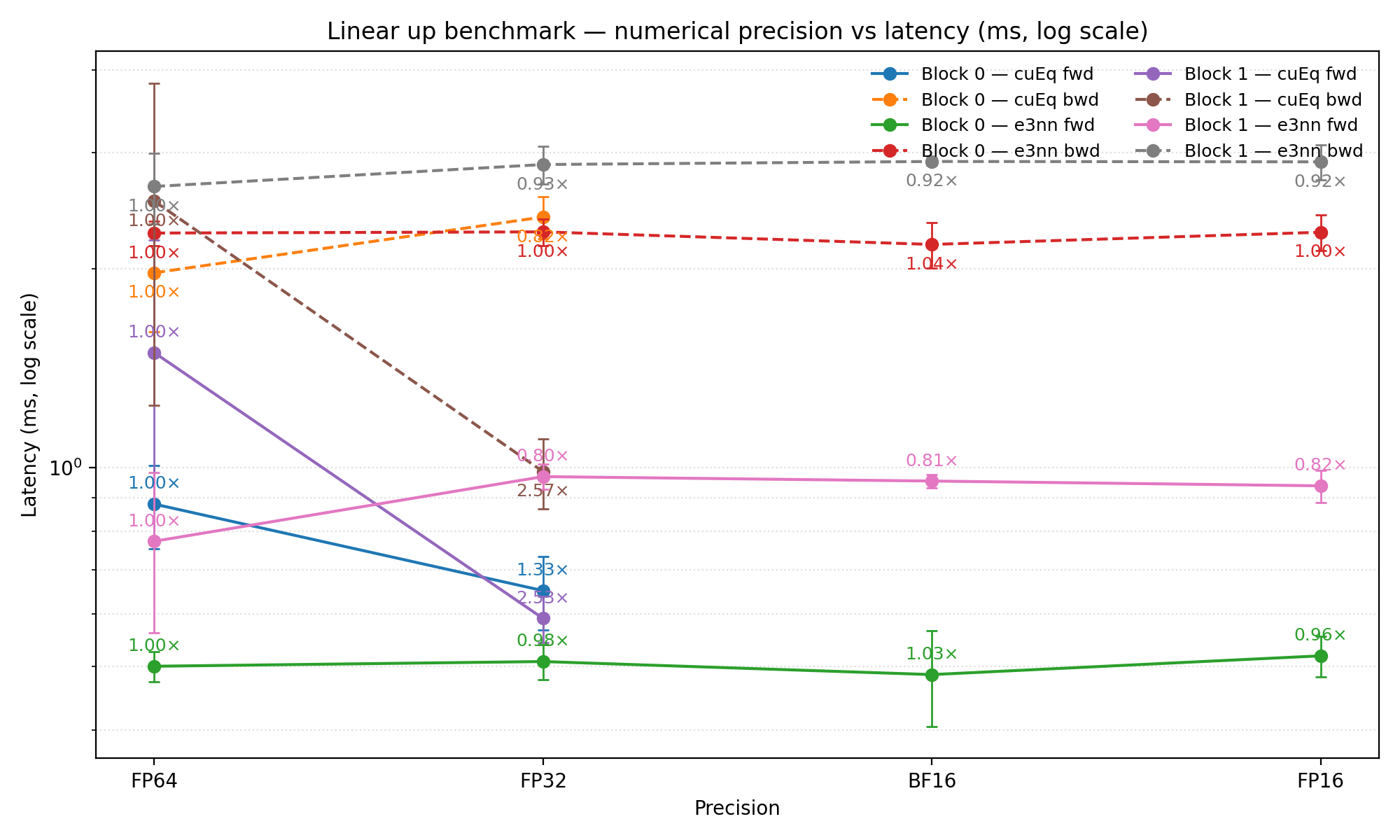}
  \caption[Linear up block]{\textbf{Linear up is launch-limited; precision offers modest gains.} Latencies are already sub-millisecond and dominated by host/launch overhead, so FP16/ BF16 provide little advantage, with cuEq improving at FP32 and largely saturating thereafter.}
  \label{fig:linear-up-latency}
\end{figure}

\begin{figure}[h]
  \centering
  \includegraphics[width=\linewidth]{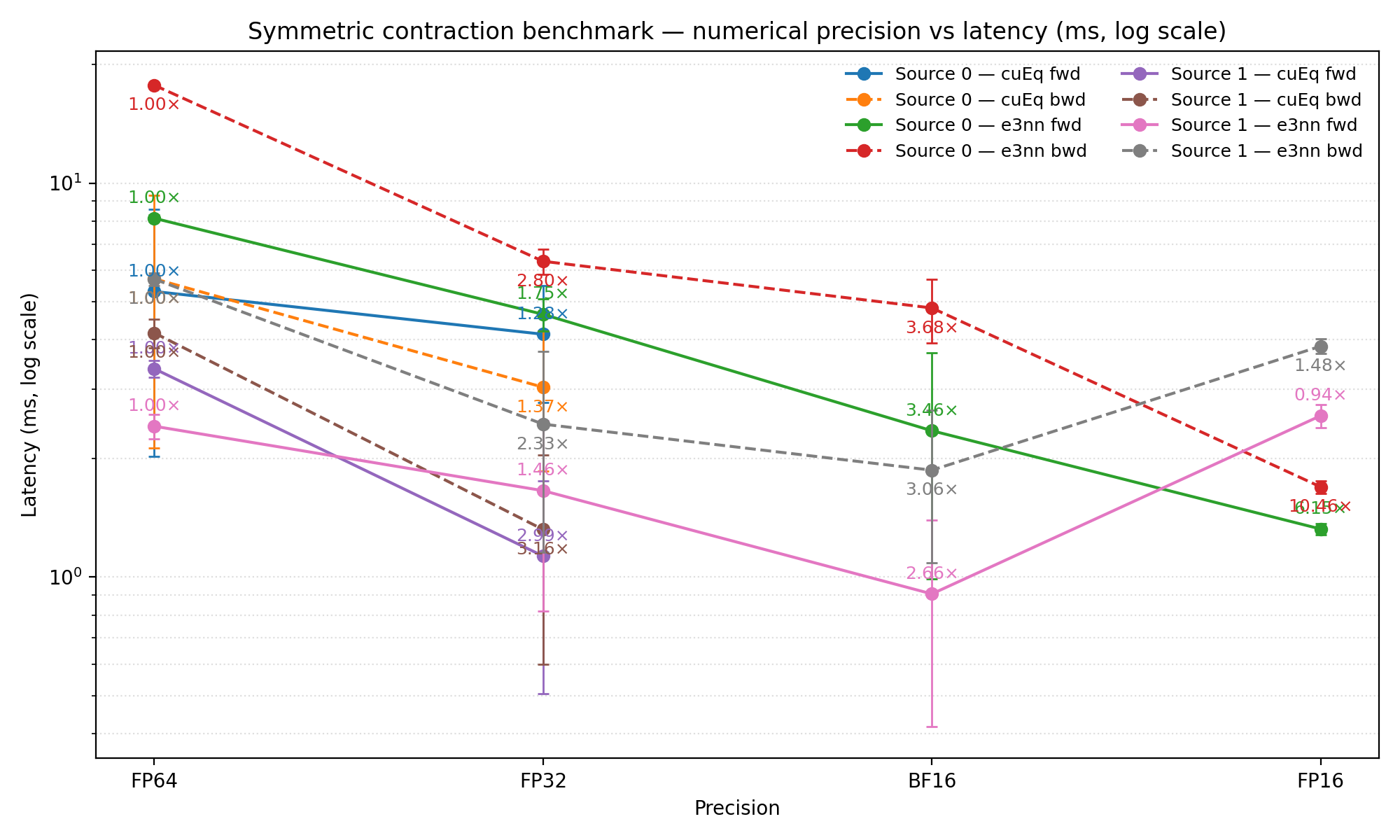}
  \caption[Symmetric contraction block]{\textbf{Symmetric contraction is compute-dense; lower precision helps most.} e3nn sees large FP32\(\rightarrow\)FP16 speedups in forward and backward; cuEq gains are strongest at FP32 and may plateau at FP16 for specific sources, consistent with kernel-choice effects.}
  \label{fig:symm-contr-latency}
\end{figure}

We microbenchmarked the heaviest stacks such as \emph{Interaction conv\_tp} (Figure~\ref{fig:conv-tp-latency}) , \emph{Linear-up} (Figure~\ref{fig:linear-up-latency}) and \emph{Symmetric contraction} (Figure~\ref{fig:symm-contr-latency}) across backends and precisions. By design, \emph{cuEquivariance (cuEq) was evaluated at FP64 and FP32 only}; BF16/FP16 on cuEq were not computed in this round. Instead, we first probed \emph{e3nn} at lower precisions to check feasibility. The plots show that e3nn forward latency scales cleanly as precision is reduced: FP64$\rightarrow$FP32 yields a clear drop, and BF16/FP16 usually reduce it further for compute-dense paths (especially \texttt{conv\_tp} and Symmetric contraction), while backward is less monotonic for some blocks. Linear-up is largely launch-limited, so all curves move little regardless of precision or backend. Importantly, tensor discrepancies remain consistent with numerical precision theory: \emph{forward} max absolute and relative errors are typically in the $10^{-6}$/$10^{-7}$ range at FP32 and rise to $10^{-3}$/$10^{-2}$ at BF16/FP16, which matches the expected loss of mantissa bits when moving from 53 (FP64) to 24 (FP32) to 10/7 (FP16/BF16). Backward errors are somewhat larger but follow the same order-of-magnitude shift.

These block-level experiments serve a specific purpose: before spending effort on kernel work for cuEq at BF16/FP16, we verified on e3nn that lower precision does in fact unlock meaningful latency gains on the math-bound blocks and that the resulting tensor errors remain within theory-aligned bounds. This outcome \emph{motivates} our next step: test \emph{cuEq BF16/FP16} beginning with the linear layers (with FP32 accumulators), where kernel fusion and tensor-core paths should compound benefits beyond the FP32 cuEq baseline, while the empirically small numerical drift observed at FP32 suggests tractable accuracy control. We do not expect dramatic wins on launch-bound components such as Linear-up, but contraction- and \texttt{conv\_tp}-heavy paths are well-positioned to benefit. End-to-end adoption will still require the force-centric checks laid out earlier (relative force error, force cosine, short NVE/NVT/NPT), yet the e3nn results provide the needed feasibility signal to proceed with cuEq BF16/FP16 trials in a targeted, accuracy-aware manner.


\section{Inference impact of mixed precision}

Some persistent buffers (e.g., radial/SH coefficients, polynomial tables) may remain FP64 even when the default dtype is set to FP32. The dtype pairs below therefore represent practical, partially mixed-precision paths rather than fully re-trained low-precision models.

Relative to the FP64 baseline ($495.05\!\pm\!4.96$\,ms), FP32-based inference achieves $3.01$–$3.26\times$ speedups with very small \emph{energy} deviations (\,$|\Delta E|\!\approx\!78$\,eV, $\sim\!0.0074\%$ or $\sim\!74$\,ppm) that are nearly constant across FP32/FP16/BF16 variants. By contrast, the reported \emph{force} magnitude increases from $9.97\times10^{-3}$ to $1.56\times10^{-2}$\,eV/\AA, an absolute shift of $5.63\times10^{-3}$\,eV/\AA\ (about $56\%$ relative). 

For this configuration, mixed-precision cuEq delivers $\sim\!3\times$ end-to-end latency reduction at essentially unchanged total energy but with a sizeable change in the reported force magnitude. The nearly identical $\Delta E$ across FP32/FP16/BF16 suggests a deterministic accumulation/rounding offset relative to FP64 rather than random noise. Overall, FP32/BF16 offers the best speed among tested settings ($3.26\times$) with energy offsets at the $\sim\!74$\,ppm level; the acceptability of the force shift is evaluated in the MD experiments that follow.

\begin{table}[H]
\centering
\resizebox{\textwidth}{!}{%
\begin{tabular}{lcccccccccc}
\toprule
\textbf{Backend} & \textbf{Dtype} & \textbf{Time (ms)} & \textbf{Speedup} & \textbf{Energy (eV)} & $\mathbf{|\Delta E|}$  & $\mathbf{\Delta E_{\%}}$ & \textbf{Forces (eV/\AA)} & $\mathbf{|\Delta F|}$ \textbf{(eV/\AA)} & $\mathbf{\Delta F_{\%}}$ \\
\midrule
CUEQ & FP64/FP64 & $495.05 \pm 4.96$ & $1.00\times$ & $-1055426.492 \pm 2.33{\times}10^{-10}$ & $0$ & $0$ & $9.97{\times}10^{-3} \pm 3.56{\times}10^{-18}$ & $0$ & $0$ \\
CUEQ & FP32/FP32 & $164.71 \pm 6.07$ & $3.01\times$ & $-1055504.875 \pm 0$ & $78.383$ & $0.00743\%$ & $1.56{\times}10^{-2} \pm 3.56{\times}10^{-9}$ & $5.63{\times}10^{-3}$ & $56.47\%$ \\
CUEQ & FP32/FP16 & $158.27 \pm 5.82$ & $3.13\times$ & $-1055504.500 \pm 0$ & $78.008$ & $0.00739\%$ & $1.56{\times}10^{-2} \pm 8.02{\times}10^{-8}$ & $5.63{\times}10^{-3}$ & $56.47\%$ \\
CUEQ & FP32/BF16 & $151.65 \pm 5.38$ & $3.26\times$ & $-1055504.625 \pm 0$ & $78.133$ & $0.00740\%$ & $1.56{\times}10^{-2} \pm 1.05{\times}10^{-8}$ & $5.63{\times}10^{-3}$ & $56.47\%$ \\
\bottomrule
\end{tabular}
}
\caption[Mixed-precision inference on cuEquivariance]{Mixed-precision inference on cuEquivariance (supercell size 8, 100 warm-up + 100 iterations, GPU utilisation $\approx$55\%). Errors are absolute/relative deviations from the FP64 baseline; speedup is the ratio of FP64 time to the reported time.}
\label{tab:mixed_precision_inference}
\end{table}

\section{Molecular dynamics precision outcome and its impact}

\subsection{Experiment 1 — Mixed-precision stability via Langevin MD with MACE-OFF24}
\label{subsec:mixed_precision_md}

\begin{figure}[H]
  \centering
  \includegraphics[width=\linewidth]{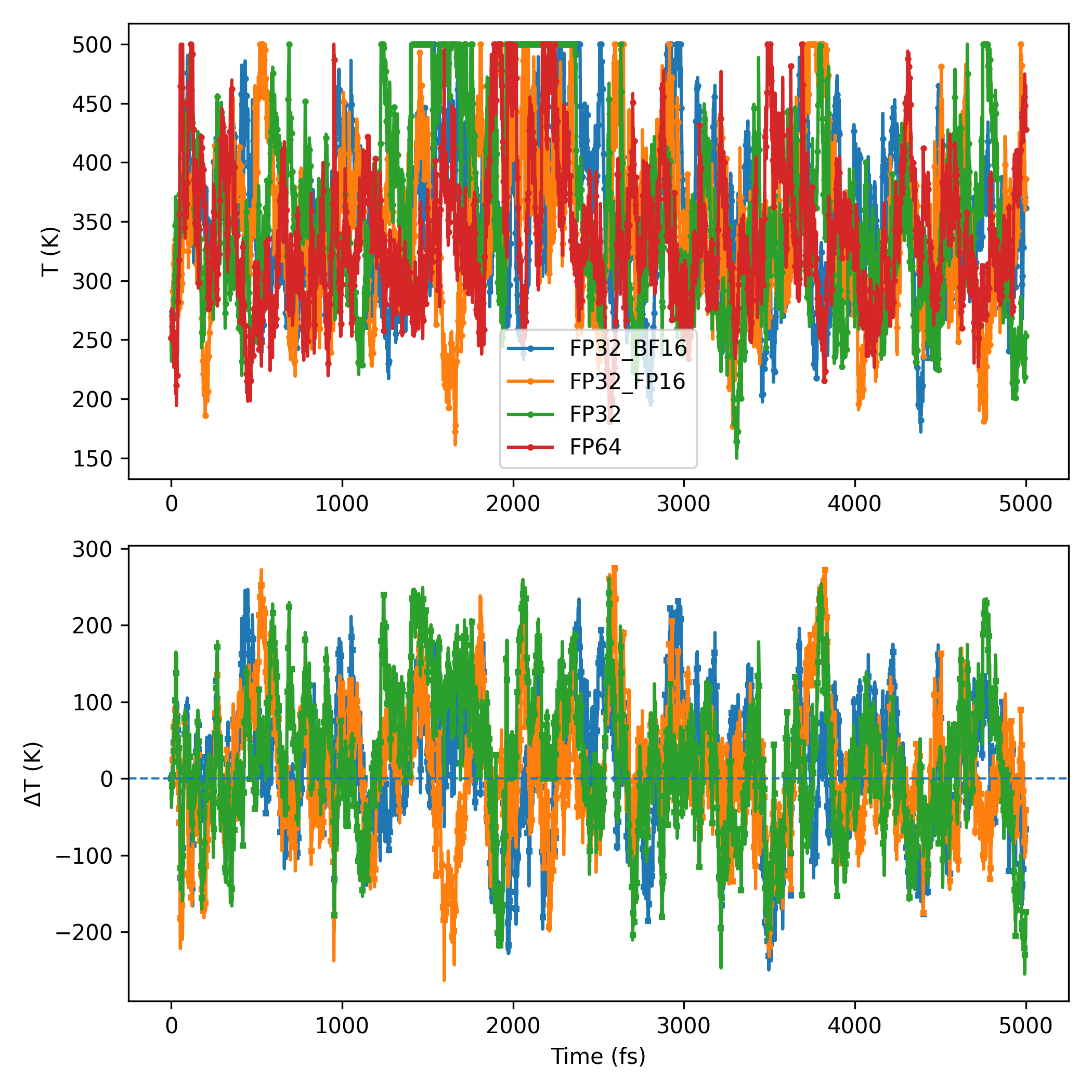}
  \caption[Temperature traces and deviations]{\textbf{Temperature traces and deviations.} Top: $T(t)$ for FP64, FP32, FP32\_FP16, and FP32\_BF16 under Langevin NVT ($\Delta t=1~\mathrm{fs}$, $\gamma=0.1$). Bottom: $\Delta T(t)$ relative to FP64. All runs remain stable; FP32\_BF16 tracks FP64 most closely, whereas FP32 and FP32\_FP16 show broader excursions.}
  \label{fig:temp_cleaned}
\end{figure}

\begin{figure}[H]
  \centering
  \includegraphics[width=\linewidth]{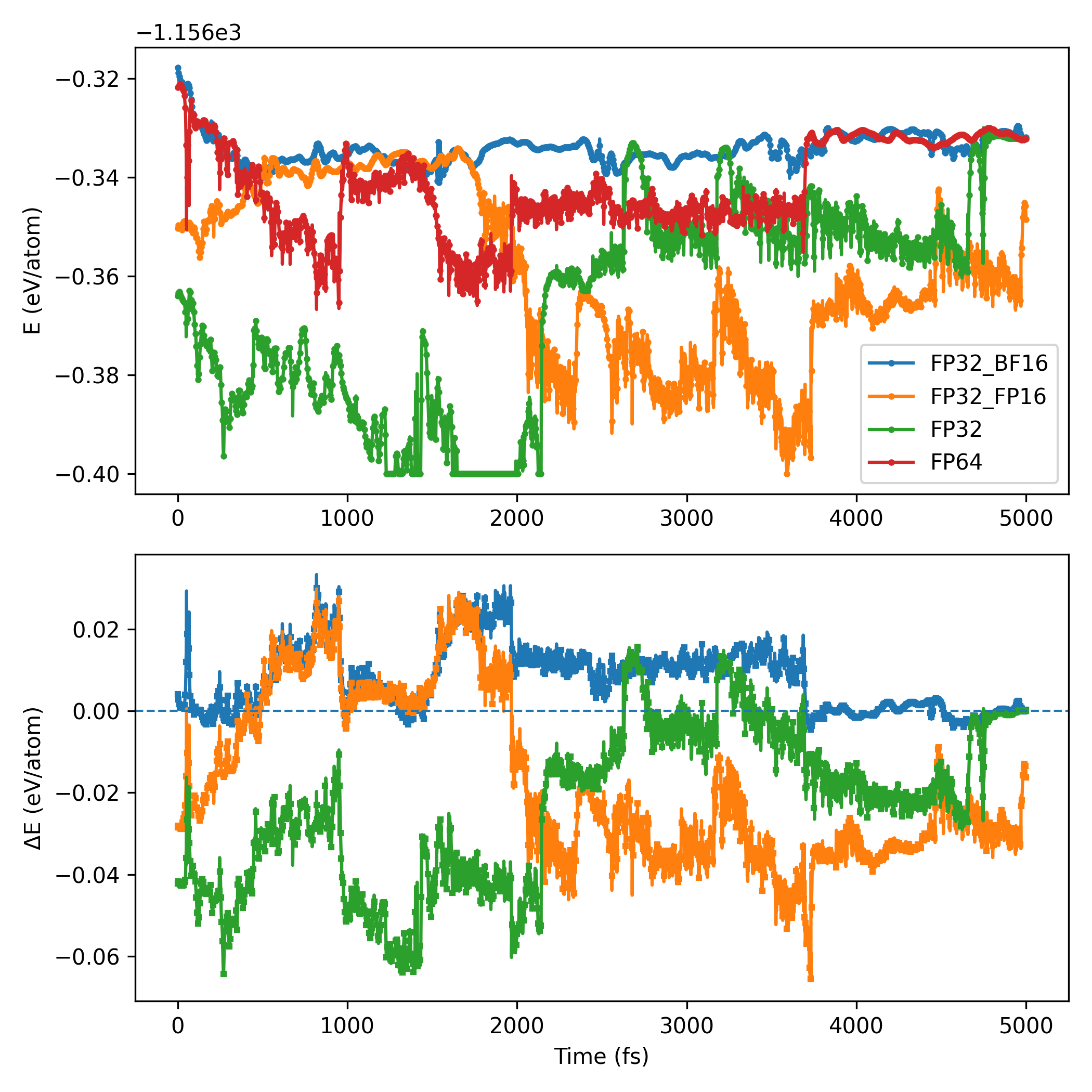}
  \caption[Potential energy per atom and deviations]{\textbf{Potential energy per atom and deviations.} Top: $E_{\mathrm{atom}}(t)$; bottom: $\Delta E_{\mathrm{atom}}(t)$ vs FP64. FP32\_BF16 aligns best with FP64; FP32 shows the largest negative bias and step-like drifts; FP32\_FP16 is intermediate.}
  \label{fig:energy_cleaned}
\end{figure}

All reduced-precision trajectories remain stable and fluctuate around the target thermal envelope (Fig.~\ref{fig:temp_cleaned}). FP32\_BF16 tracks FP64 most closely across the entire window, whereas FP32\_FP16 and pure FP32 exhibit broader temperature excursions. In the energy traces (Fig.~\ref{fig:energy_cleaned}), FP32\_BF16 again aligns best with FP64; FP32 shows step-like negative bias and the largest deviations, while FP32\_FP16 sits between the two. Quantitatively (Table~\ref{tab:mp_metrics}), energy RMSE and bias confirm this ordering: FP32 has the largest error and bias ($E_{\rm RMSE}=35.3~\mathrm{meV/atom}$, $E_{\rm bias}=-25.2~\mathrm{meV/atom}$), FP32\_FP16 is intermediate ($26.3$ and $-16.9$), and FP32\_BF16 is closest to FP64 ($11.1$ and $+7.9$). Temperature deviations follow the same pattern, with FP32\_BF16 showing the smallest RMSE ($\sim 89~\mathrm{K}$) and tighter maxima ($\sim 315~\mathrm{K}$). Force peaks remain comparable to FP64 for FP32\_FP16 and are slightly softened for FP32\_BF16.

On these timescales, mixed precision can be used without catastrophic loss of stability, but the choice of format matters. FP32\_BF16 offers the best accuracy–speed trade-off among reduced-precision modes, consistent with numerical precision theory: BF16 preserves an FP32-like exponent and mitigates dynamic-range issues in higher-order tensor products and reductions, while FP16’s smaller exponent is more brittle and pure FP32 accumulates noticeable bias. This ranking motivates using BF16 for linear layers for throughput-critical paths while keeping FP32 accumulators, and it supports proceeding with cuEquivariance BF16/FP16 trials in a controlled, force-centric validation loop.

\begin{table}[H]
  \centering
  \caption{Mixed-precision MD metrics versus FP64 baseline.}
  \label{tab:mp_metrics}
  \begin{tabular}{lrrrrrr}
    \toprule
    Variant & $E_{\rm RMSE}$ & $E_{\rm bias}$ & $T_{\rm RMSE}$ & $T_{\rm bias}$ & $|\Delta T|_{\max}$ & max\(|F|\) \\
            & (meV/atom) & (meV/atom) & (K) & (K) & (K) & ratio \\
    \midrule
    FP64        & 0.000 & 0.000  & 0.00   & 0.00  & 0.0   & 1.000 \\
    FP32        & 35.311 & $-25.230$ & 1368.72 & 295.24 & 15861.7 & 5.493 \\
    FP32\_BF16  & \textbf{11.130} & \textbf{+7.885} & \textbf{89.06} & \textbf{13.11} & \textbf{315.0} & \textbf{0.137} \\
    FP32\_FP16  & 26.262 & $-16.946$ & 92.67  & 9.59  & 608.4 & 1.036 \\
    \bottomrule
  \end{tabular}
\end{table}

\subsection{Experiment 2 - NPT water box and compare it to precision }

All precisions thermalise to \(\sim 299~\mathrm{K}\) within a few hundred steps and then track each other closely; mean temperature differences relative to FP64 are \(|\Delta T|/T_{64}<0.03\%\). Densities converge to \(1.073\text{–}1.077~\mathrm{g\,cm^{-3}}\); the largest shift is BF16 at \(+0.29\%\) vs FP64, well within the per-run standard deviations (\(\sim 0.013\text{–}0.016\)). Potential energies show a constant offset of \(\approx 0.9~\mathrm{eV}\) on a \(\sim 2.7{\times}10^{5}~\mathrm{eV}\) scale (i.e., \(<4\) ppm, sub-meV/atom) with parallel post-equilibration trends, indicating no precision-induced drift. MSD varies modestly (FP16 \(\sim\!{+}10\%\), BF16 \(\sim\!{-}5\%\) vs FP64) but remains small relative to run-to-run spread. The drift statistic settles to \(0.07\pm 0.02\) for all modes; relative differences (\(\le 8\%\)) are inside one standard deviation. Performance scales as expected: FP32 halves wall time vs FP64 (\(\sim\!2.0\times\)), and BF16/FP16 are \(\sim\!4.3\times\) faster than FP64 (MeSteps/day \(2.44\text{–}2.47\) vs \(0.56\)).

For this system, casting only Linear layers to BF16/FP16 within an FP32 cuEq model preserves the NPT ensemble within noise while delivering substantial speedups. Relative deviations versus FP64 are tiny for temperature (\(<0.03\%\)), small for density (\(<0.29\%\)), ppm-level for energy, and within uncertainty for the drift metric, supporting the use of \textbf{BF16 (preferred)} or FP16 for Linear layers with FP32 accumulators, FP32 as a conservative default, and FP64 for reference checks. Reporting MeSteps/day provides a hardware-agnostic throughput measure for fixed step counts.
This mixed-precision policy achieves $\approx\!2\times$ (FP32) to $\approx\!4.3\times$ (BF16/FP16) faster MD without compromising stability.
 

\begin{figure}[H]
  \centering
  \includegraphics[width=0.75\linewidth]{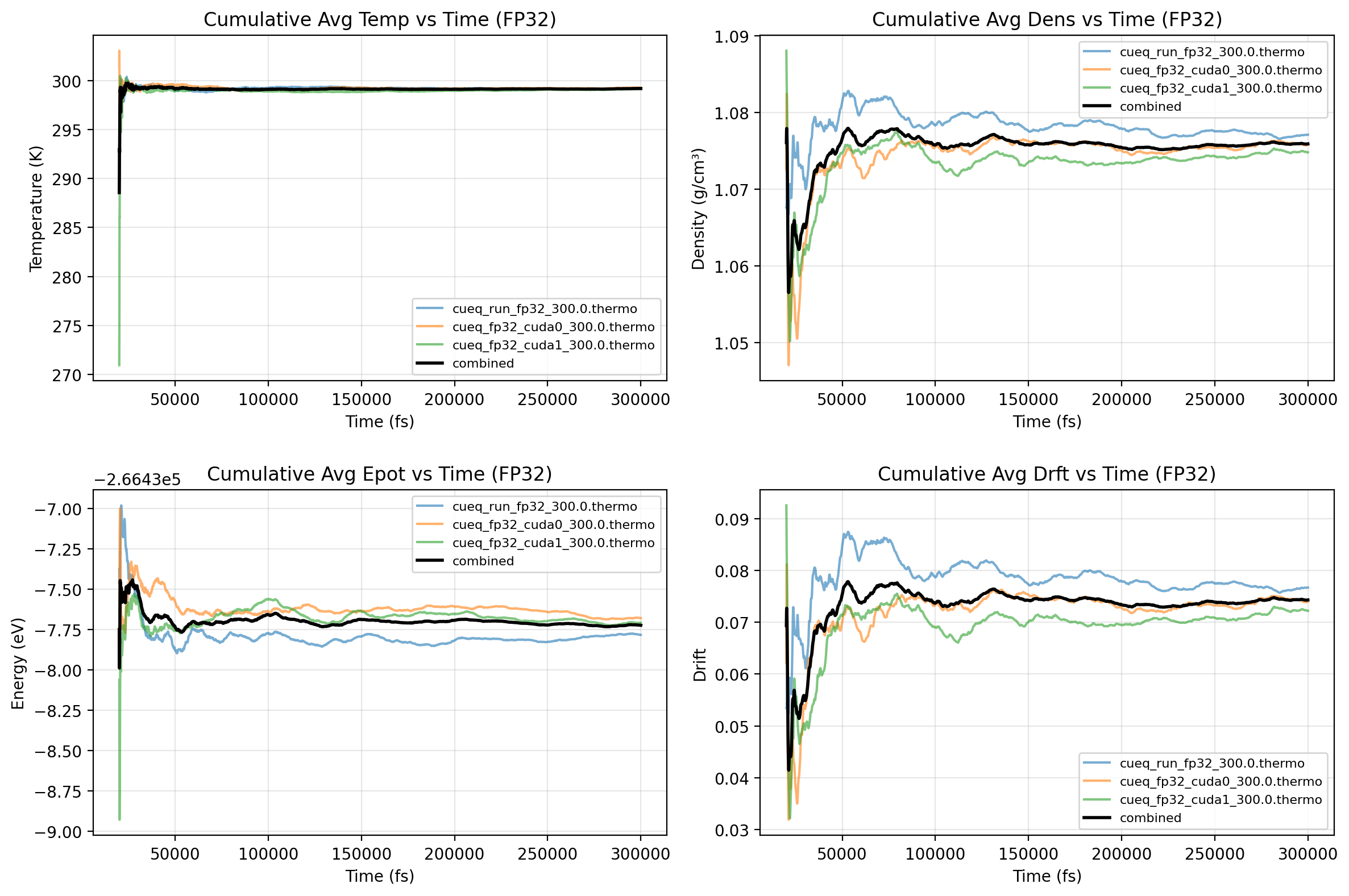}
  \caption{FP32 plots summary (3 seeds averaged).}
  \label{fig:fp32_summary}
\end{figure}

\begin{figure}[H]
  \centering
  \includegraphics[width=0.75\linewidth]{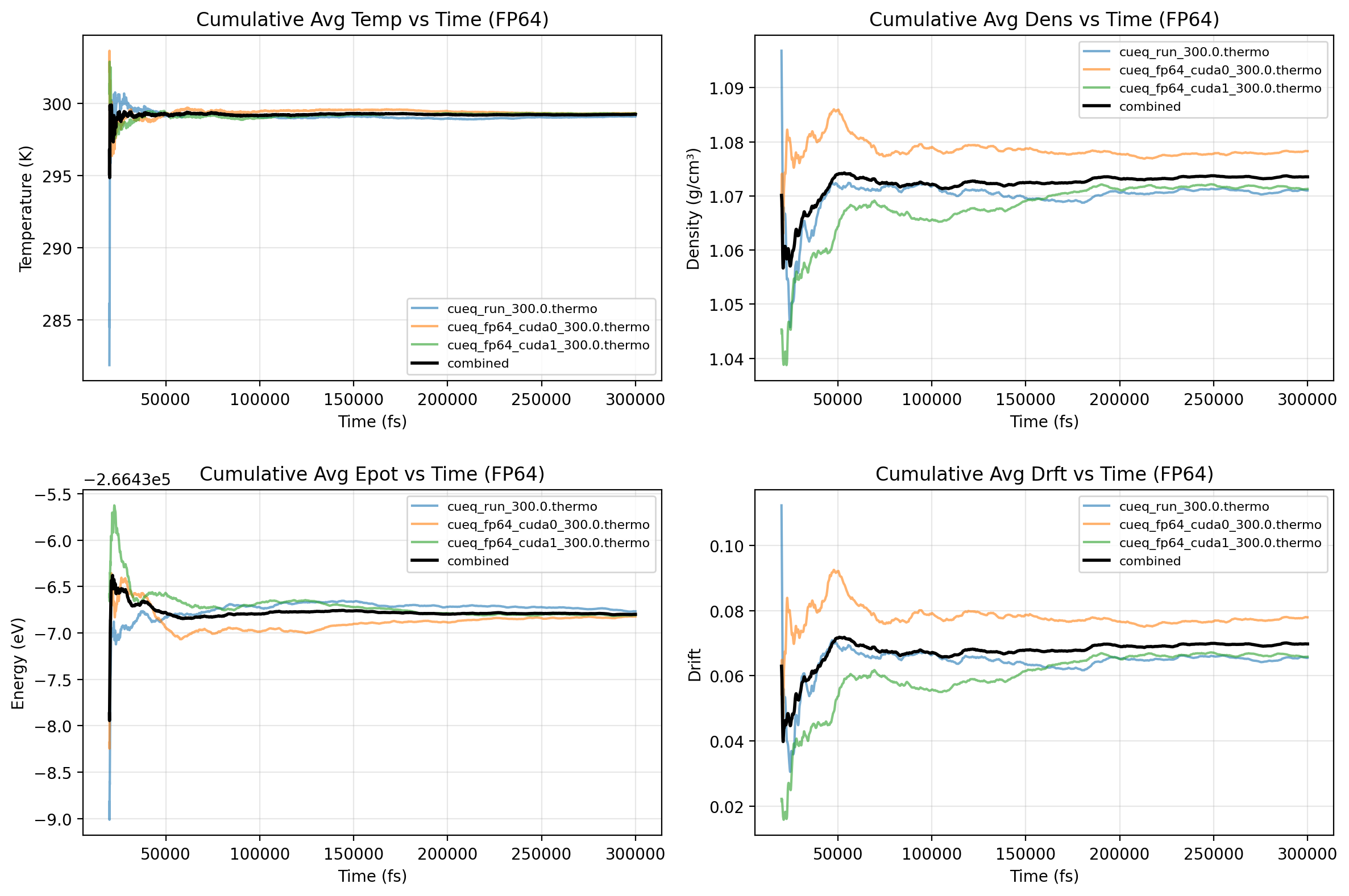}
  \caption{FP64 plots summary (3 seeds averaged).}
  \label{fig:fp64_summary}
\end{figure}

\begin{figure}[H]
  \centering
  \includegraphics[width=0.75\linewidth]{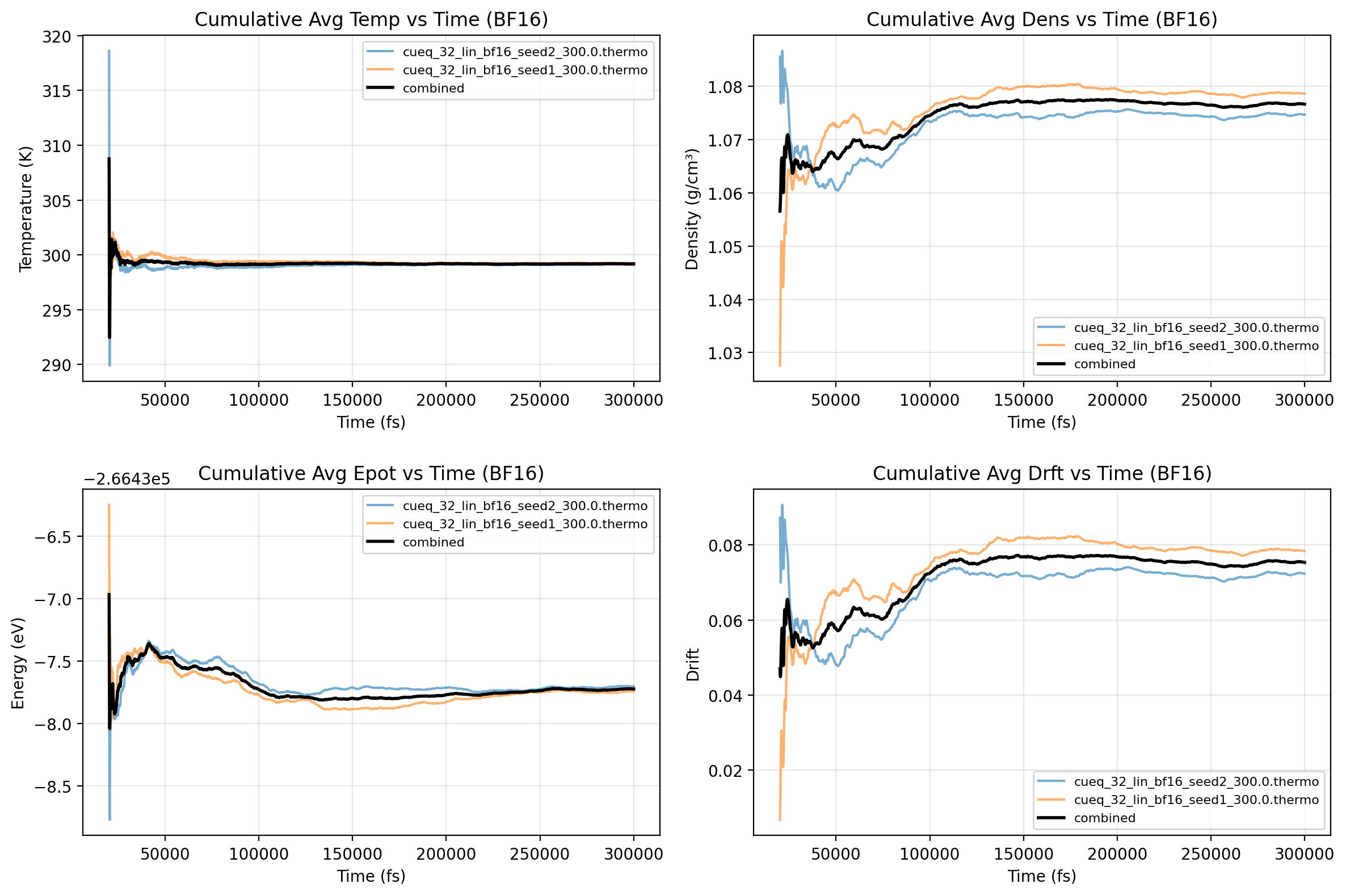}
  \caption{FP32 with Linear Layers in BF16 — plots summary (3 seeds averaged).}
  \label{fig:fp32_linbf16_summary}
\end{figure}

\begin{figure}[H]
  \centering
  \includegraphics[width=0.75\linewidth]{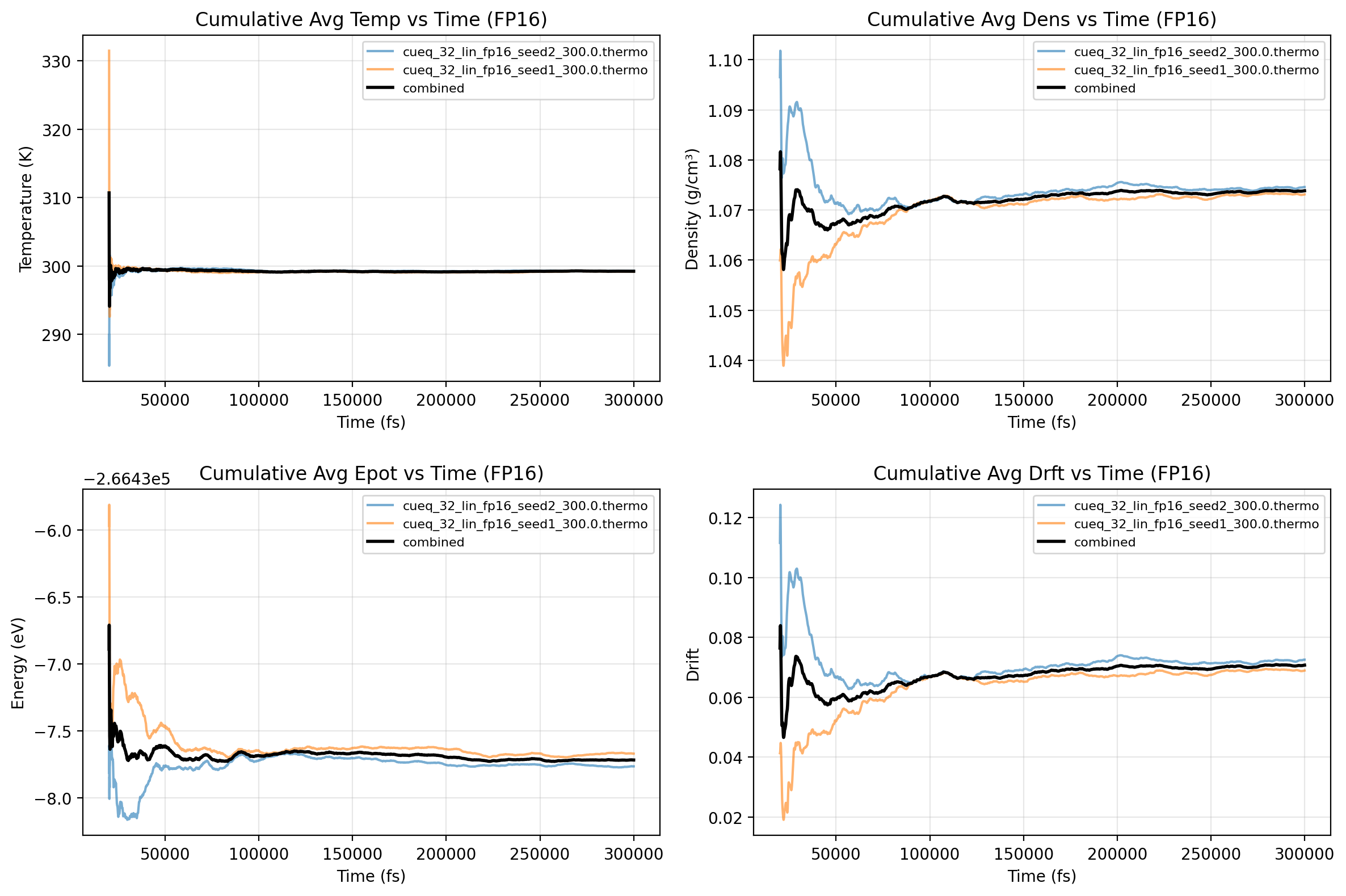}
  \caption{FP32 with Linear Layers in FP16 — plots summary (3 seeds averaged).}
  \label{fig:fp32_linfp16_summary}
\end{figure}

\begin{figure}[H]
    \centering
    \includegraphics[width=\linewidth]{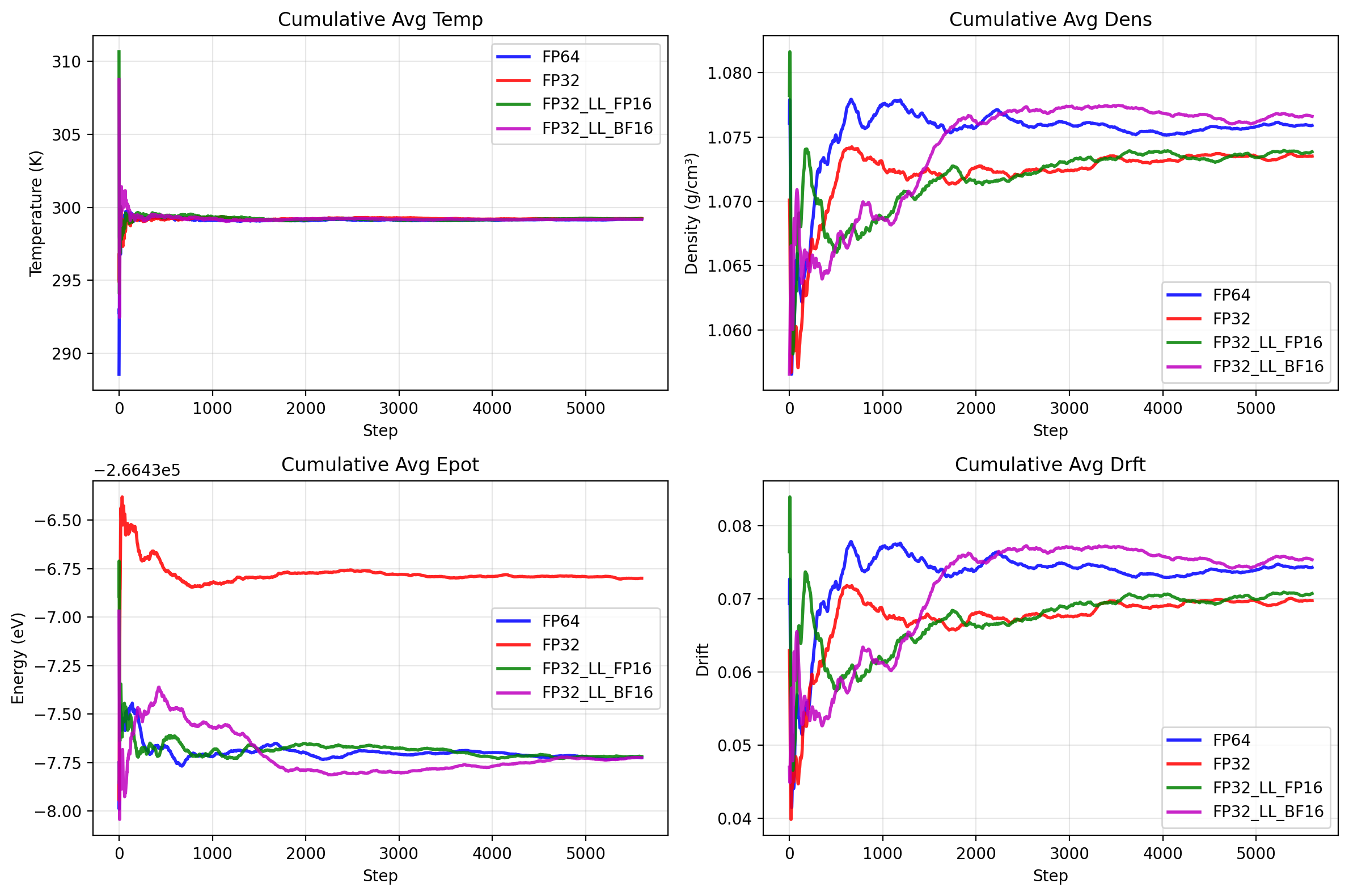}
    \caption[Direct comparison across precisions]{\textbf{Direct comparison across precisions.} Seed-averaged (3 of them traces for FP64, FP32, and FP32 with Linear layers cast to BF16 or FP16, shown on identical axes for visual comparison.}
    \label{fig:precision_4way}
\end{figure}

\begin{table}[h]
\centering
\resizebox{\textwidth}{!}{
\begin{tabular}{lcccc}
\toprule
\textbf{Metric} & \textbf{FP64} & \textbf{FP32} & \textbf{BF16} & \textbf{FP16} \\
\midrule
Temperature (K)       & 299.241 $\pm$ 7.18                  & 299.198 $\pm$ 7.24                  & 299.177 $\pm$ 8.76                    & 299.211 $\pm$ 9.04 \\
Density (g/cm$^3$)    & 1.073520 $\pm$ 1.31$\times 10^{-2}$ & 1.075910 $\pm$ 1.31$\times 10^{-2}$ & 1.076609 $\pm$ 1.63$\times 10^{-2}$   & 1.073862 $\pm$ 1.61$\times 10^{-2}$ \\
Potential Energy (eV) & -266436.8 $\pm$ 5.14$\times 10^{-1}$& -266437.7 $\pm$ 5.40$\times 10^{-1}$& -266437.7 $\pm$ 6.77$\times 10^{-1}$  & -266437.7 $\pm$ 6.49$\times 10^{-1}$ \\
MSD (\AA$^2$)         & 185.851 $\pm$ 9.13$\times 10^{1}$   & 183.556 $\pm$ 9.48$\times 10^{1}$   & 176.995 $\pm$ 8.19$\times 10^{1}$     & 204.417 $\pm$ 9.67$\times 10^{1}$ \\
Drift                 & 0.06977 $\pm$ 2.42$\times 10^{-2}$  & 0.07431 $\pm$ 2.46$\times 10^{-2}$  & 0.07535 $\pm$ 3.06$\times 10^{-2}$    & 0.07074 $\pm$ 2.92$\times 10^{-2}$ \\
Wall time (wt) (h)         & 12.65 $\pm$ 3.45                    & 6.22 $\pm$ 1.67                      & 2.92 $\pm$ 1.47                        & 2.95 $\pm$ 1.48 \\
\midrule
Speedup $\uparrow$ & - & 2.03 & 4.33 & 4.28\\
\toprule
MeSteps/day & 0.56 & 1.16 & 2.47 & 2.44 \\
\midrule
Speedup $\uparrow$ & - & 2.07 & 4.42 & 4.36 \\
\bottomrule
\end{tabular}
}
\caption[Final means per numerical precision]{Final means per numerical precision (ASE-style units: K, g/cm$^3$, eV, \AA$^2$). BF16/FP16 wall time converted from seconds to hours. Drift shown as reported; unit depends on simulation setup. MeSteps are the Mega Steps per day}
\label{tab:md_final_means}
\end{table}

\begin{table}[h]
\centering
\resizebox{\textwidth}{!}{
\begin{tabular}{lccc}
\toprule
\textbf{Metric} & \textbf{$|\Delta|$ FP32 vs FP64} & \textbf{$|\Delta|$ BF16 vs FP64} & \textbf{$|\Delta|$ FP16 vs FP64} \\
\midrule
Temperature (K)       & $0.043$                     & $0.064$                     & $0.030$ \\
Density (g/cm$^3$)    & $2.39\times10^{-3}$         & $3.09\times10^{-3}$         & $3.42\times10^{-4}$ \\
Potential Energy (eV) & $0.900$                     & $0.900$                     & $0.900$ \\
MSD (\AA$^2$)         & $2.295$                     & $8.856$                     & $18.566$ \\
Drift                 & $4.54\times10^{-3}$         & $5.58\times10^{-3}$         & $9.70\times10^{-4}$ \\
Wall time (h)         & $6.43$                      & $9.73$                      & $9.70$ \\
MeSteps/day           & $0.60$                      & $1.91$                      & $1.88$ \\
\bottomrule
\end{tabular}}
\caption[Absolute differences of the seed-averaged means]{Absolute differences of the seed-averaged means in Table~\ref{tab:md_final_means} with respect to the FP64 baseline. 
For performance metrics (wall time, MeSteps/day) these are absolute \emph{differences}, not errors.}
\label{tab:md_abs_errors}
\end{table}

Across all thermodynamic observables the mixed-precision runs are effectively indistinguishable from FP64 once run-to-run noise is considered. Temperatures differ from FP64 by at most \(0.064~\mathrm{K}\) on a \(299~\mathrm{K}\) baseline (\(\lesssim 0.02\%\)), and densities by at most \(3.1\times10^{-3}~\mathrm{g\,cm^{-3}}\) on a \(\sim 1.074~\mathrm{g\,cm^{-3}}\) baseline (\(\approx 0.29\%\)). Both gaps are smaller than the reported standard deviations, so they are not statistically meaningful for this box size. Potential energy shows a uniform \(\sim 0.90~\mathrm{eV}\) offset across FP32/BF16/FP16; on a total scale of \(2.66\times10^{5}~\mathrm{eV}\) that is \(3\text{--}4\) ppm (sub-meV/atom), and the constancy of the offset suggests deterministic rounding/accumulation differences rather than drift.

Mean-squared displacement (MSD) shifts are modest relative to their uncertainties (FP32 \(-1.2\%\), BF16 \(-4.8\%\), FP16 \(+10.0\%\) vs.\ FP64, while \(\sigma/\text{mean}\approx 0.4\text{--}0.5\)), and the drift metric differs by only a few \(10^{-3}\) in absolute value (\(\le 8\%\) relative) with overlapping error bars --- again consistent with no precision-induced degradation of long-time behaviour.

On performance, FP32 about halves wall time versus FP64, and casting only the Linear layers to BF16 or FP16 on cuEq yields a further \(\sim 4.3\times\) reduction relative to FP64 (MeSteps/day rising from \(0.56\) to \(\sim 2.45\)). Because BF16 retains the larger exponent, it offers a slightly safer dynamic range than FP16 while delivering similar throughput. Practically, use FP32 as the default for production NPT and enable BF16 (or FP16) for Linear layers with FP32 accumulation when targeting maximum throughput.

\begin{figure}[H]
    \centering
    \includegraphics[width=\linewidth]{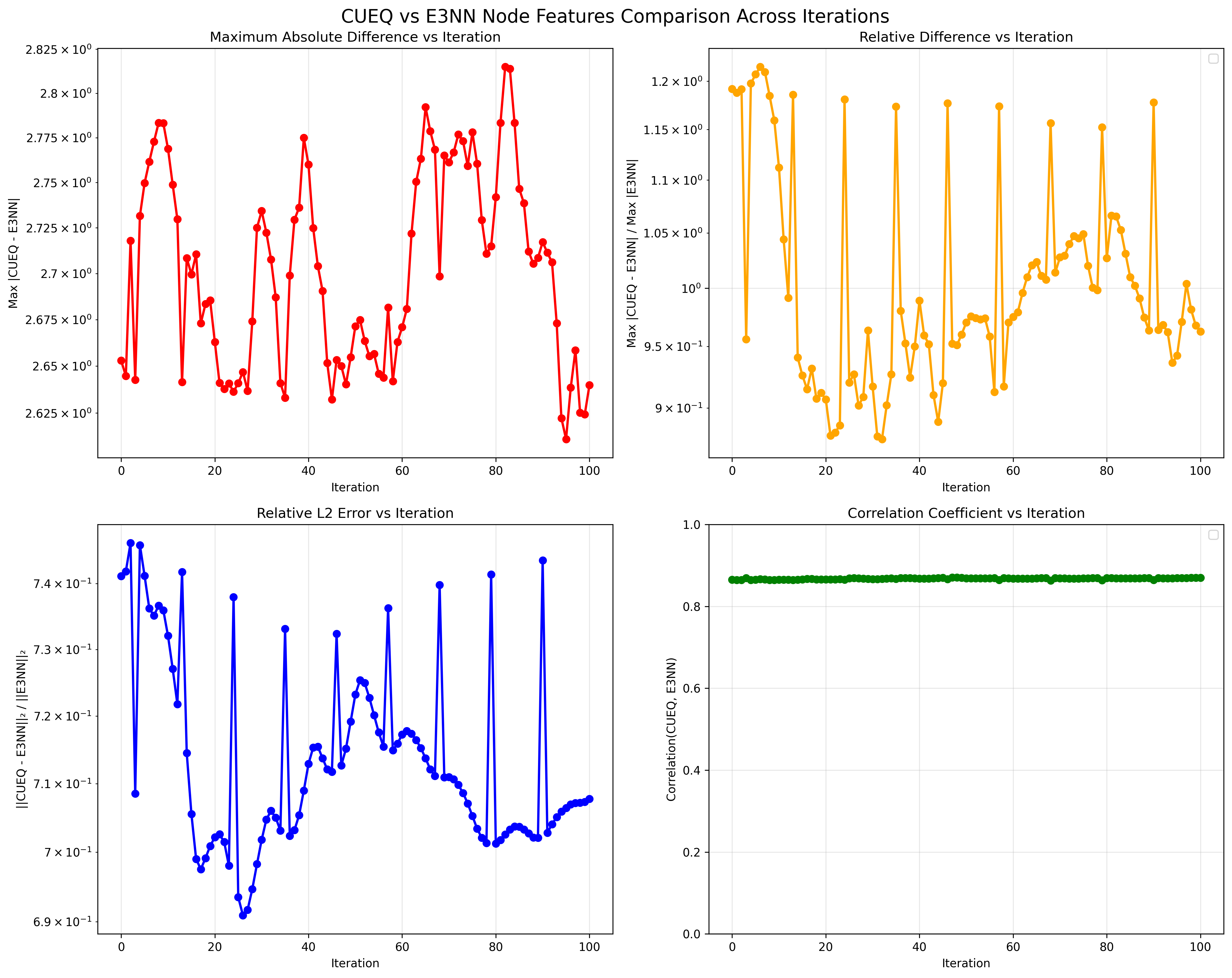}
    \caption[cuEq vs.\ e3nn node features disagree when backends are mixed]{\textbf{cuEq vs.\ e3nn node features disagree when backends are mixed.}
    Over 100 steps, the last readout features produced by a cuEq model differ systematically from those produced by an otherwise-identical model where only the final \texttt{Linear} is replaced with \texttt{e3nn.o3.Linear}. 
    The maximum absolute difference remains $\mathcal{O}(1)$–$\mathcal{O}(3)$, the relative $L_2$ error stays around $\sim\!0.70$, and the correlation coefficient plateaus at $\sim\!0.88$–$0.90$, indicating a persistent mismatch rather than rounding noise.}
    \label{fig:disrepencies}
\end{figure}

Figure~\ref{fig:disrepencies} shows that swapping only the final \texttt{Linear} for the e3nn implementation inside an otherwise cuEq model produces large, stationary discrepancies in node features: the relative $L_2$ error hovers near $0.70$, the per-step maxima do not decay, and the correlation with the cuEq output never exceeds $\sim\!0.90$.

This behaviour is consistent with a representation/layout incompatibility, not precision noise. cuEq’s blocks emit features in the configured \texttt{ir\_mul} layout, whereas \texttt{e3nn.o3.Linear} expects e3nn’s ordering; applying e3nn’s linear layer to cuEq-ordered features permutes channels and alters the physics. Mixing backends within the same readout stack also breaks implicit assumptions about how weights are mapped when converting models. In scripted modules, AMP/autocast boundaries may be ignored or partially fused, further obscuring dtype handling. The practical guidance is to avoid interleaving backends unless explicit adapters are inserted to convert between layouts and weight orderings; otherwise, comparisons of cuEq vs e3nn must be performed with \emph{end-to-end} homogeneous backends.

\section{Training at lower precision: RMSE, structure, and MD behaviour}

\begin{figure}[H]
    \centering
    \begin{subfigure}{0.48\linewidth}
        \centering
        \includegraphics[width=\linewidth]{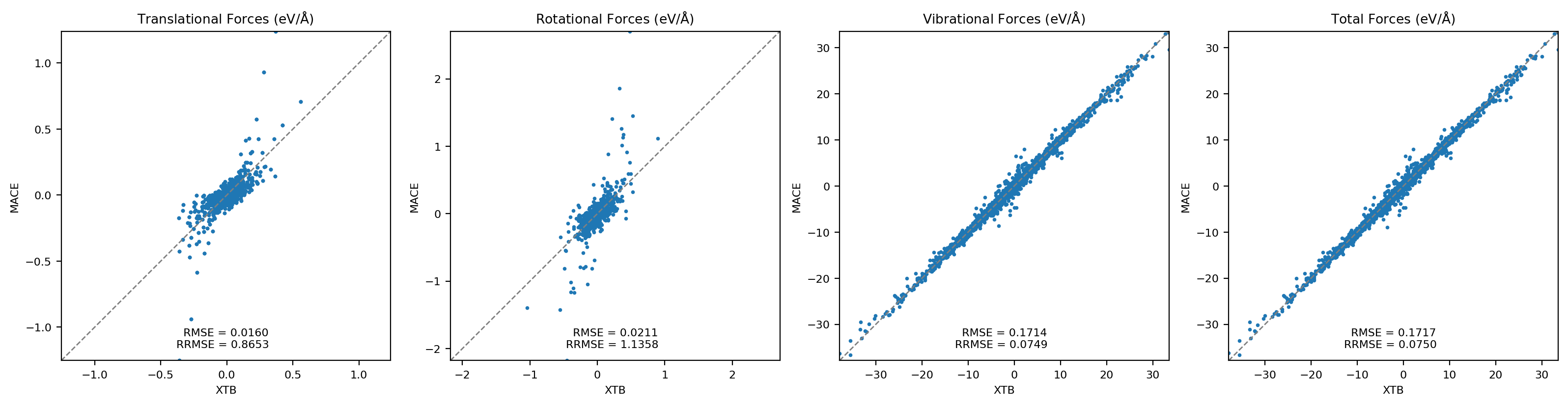}
        \caption{Intra and Inter decomposition fp64}
        \label{fig:intra_inter_fp64}
    \end{subfigure}
    \hfill
    \begin{subfigure}{0.48\linewidth}
        \centering
        \includegraphics[width=\linewidth]{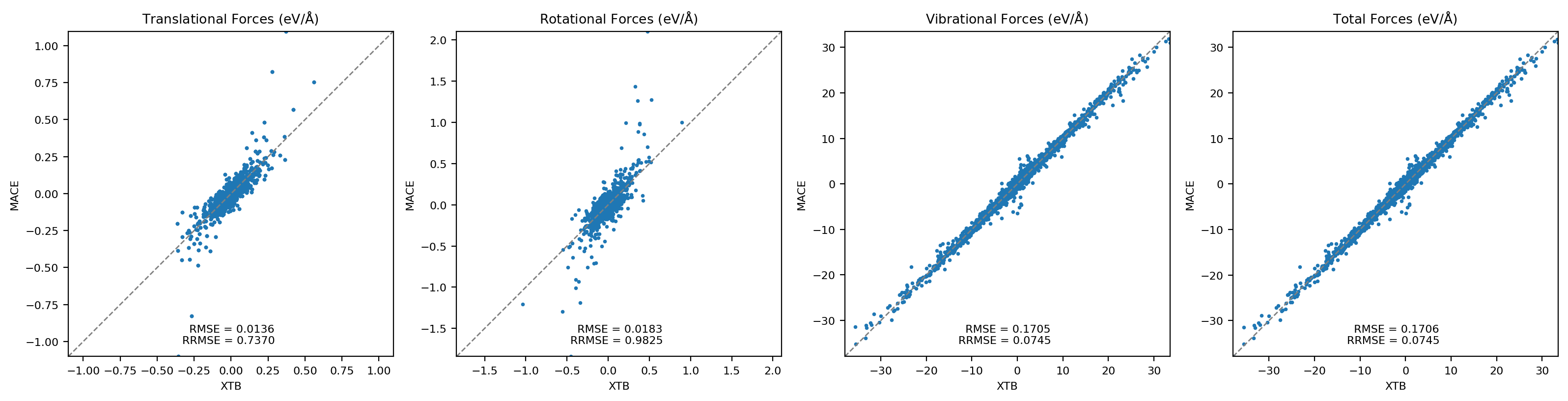}
        \caption{Intra and Inter decomposition fp32}
        \label{fig:intra_inter_fp32}
    \end{subfigure}
    
    \vspace{0.5cm}
    \begin{subfigure}{0.48\linewidth}
        \centering
        \includegraphics[width=\linewidth]{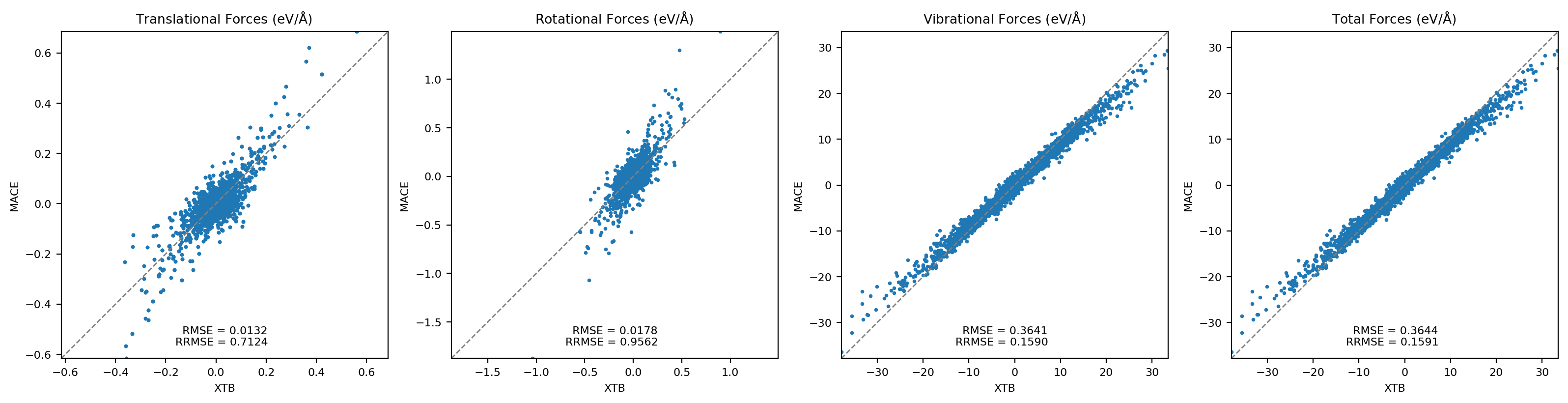}
        \caption{Intra and Inter decomposition fp32 and fp16}
        \label{fig:intra_inter_fp16}
    \end{subfigure}
    \hfill
    \begin{subfigure}{0.48\linewidth}
        \centering
        \includegraphics[width=\linewidth]{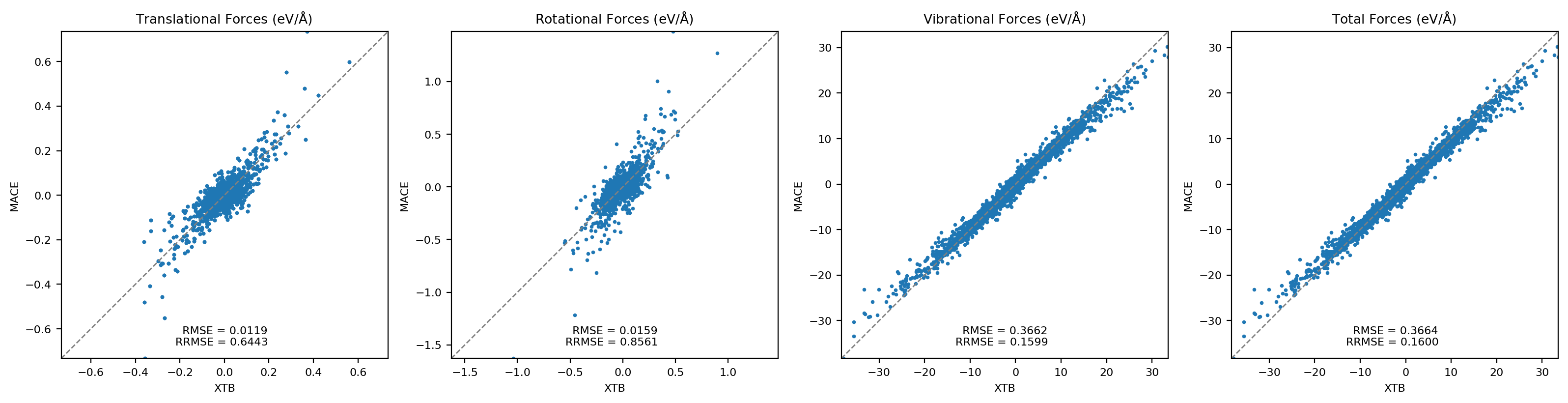}
        \caption{Intra and Inter decomposition fp32 and bf16}
        \label{fig:intra_inter_bf16}
    \end{subfigure}

    \caption[Comparison of intra- and inter-decomposition]{Comparison of intra- and inter-decomposition under different numerical precision settings.}
    \label{fig:intra_inter_comparison}
\end{figure}

\begin{figure}[H]
    \centering
    \includegraphics[width=1\linewidth]{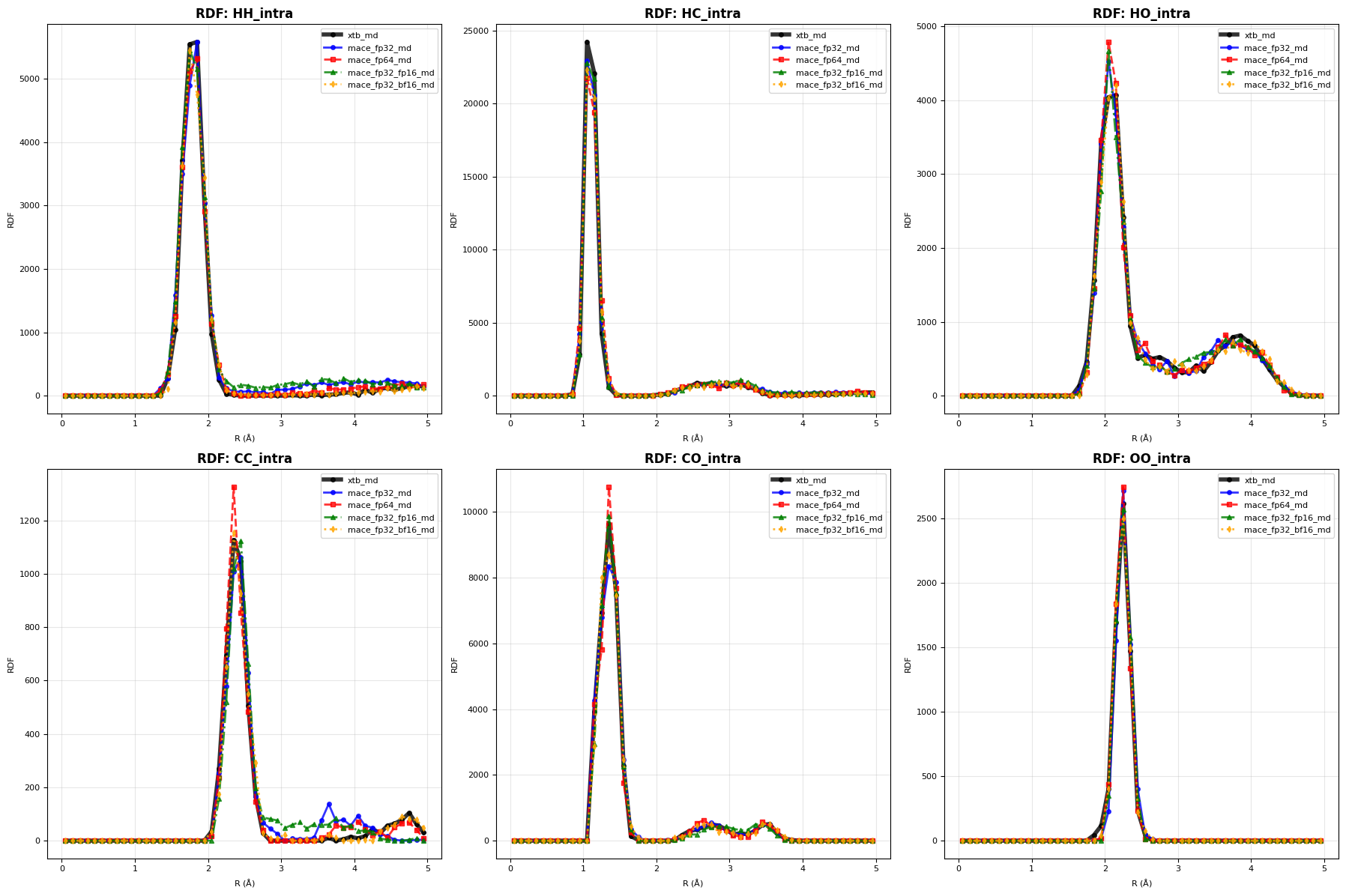}
    \caption[Partial radial distribution function]{Partial radial distribution function 
$g_{HO}(r)$ (H$\cdots$O) comparing
\texttt{xtb\_md} (black), \texttt{mace\_fp64} (red), \texttt{mace\_fp32} (blue), 
\texttt{mace\_fp32\_linear\_fp16} (green) and \texttt{mace\_fp32\_linear\_bf16} (orange). 
The peak at $\sim 2.0\,\text{\AA}$ is characteristic of the intermolecular hydrogen bond 
network in liquid water.}
    \label{fig:rdf_ho}
\end{figure}

Table~\ref{tab:error_table} shows that \emph{energy} RMSE is largely insensitive to dtype across splits (6–8\,meV/atom), with FP32 and FP64 nearly indistinguishable and BF16 sometimes lowest. By contrast, \emph{forces} degrade when Linear layers are in half precision: on the \emph{test} set, FP16/BF16 raise force RMSE from \(171.7\to 220~\mathrm{meV/\AA}\) (rel.\ RMSE \(7.50\%\to 9.6\%\)), while pure FP32 matches FP64 (\(170.6~\mathrm{meV/\AA}\), \(7.45\%\)). The unusually \emph{low} train force RMSE for FP32\(\rightarrow\)FP16 (60.5\,meV/\AA) does not carry to validation/test, indicating overfitting or a train/eval mismatch rather than a genuine advantage. The intra/inter decomposition in Fig.~\ref{fig:intra_inter_comparison} confirms that inter-molecular components are the hardest to learn: absolute errors are smaller but relative errors are larger for the inter terms, independent of dtype. RDFs in Fig.~\ref{fig:rdf_ho} and the multi-pair panel overlay almost perfectly across all precisions and with xTB: peak positions and widths are preserved (e.g., the hydrogen-bond peak near \(r\!\approx\!2.0\,\text{\AA}\)), indicating that structural sampling is robust to these precision choices. Short solvent MD runs are stable for all precisions; wall time improves from FP64 to FP32 and further when Linear layers are FP16/BF16, consistent with earlier performance sections.

Lower-precision \emph{training} introduces stochastic rounding and reduced mantissa that the model can absorb on the training split but that harms \emph{force} generalization when Linear weights are held in FP16/BF16. Energies remain accurate because they average errors and have higher signal-to-noise, while inter-molecular forces expose dynamic-range limits most acutely. The structural observables (RDFs) and MD stability do not degrade, so the precision sensitivity is primarily a force-accuracy issue rather than a breakdown of the sampled ensemble. Practically, these results argue for training in FP32 (or FP64 master weights with AMP) and reserving half precision for \emph{inference-time} speedups in Linear layers with FP32 accumulation. If lower-precision training is revisited, it should use loss scaling and FP32 master weights, and be validated with force-cosine, per-species inter/intra errors, and short NVT/NPT stability checks. 





\begin{table}[H]
\centering
\caption{Error table on Train, Validation, and Test sets under different numerical precisions under cuEq backend}
\label{tab:error_table}
\begin{tabular}{lllrrr}
\hline
\textbf{Dataset} & \textbf{Default} & \textbf{Linear} & \textbf{RMSE E} & \textbf{RMSE F} & \textbf{Rel.\ F RMSE} \\
 &  \textbf{dtype}& \textbf{dtype}& (meV/atom) & (meV/\AA) & (\%) \\
\hline
\multirow{5}{*}{Train} 
 &  fp64 & fp64 & 6.5 & 104.3 & 4.79 \\
 &  fp32 & fp32 & 6.8 & 106.7 & 4.90 \\
 &  fp32 & fp16 & 6.2 & 60.5  & 2.77 \\
 &  fp32 & bf16 & 5.7 & 202.7 & 7.84 \\
\hline
\multirow{4}{*}{Validation} 
 &  fp64 & fp64 & 6.6 & 161.2 & 6.23 \\
 &  fp32 & fp32 & 8.6 & 172.1 & 6.65 \\
 &  fp32 & fp16 & 6.3 & 208.2 & 8.05 \\
 &  fp32 & bf16 & 6.0 & 202.7 & 7.84 \\
\hline
\multirow{4}{*}{Test} 
 &  fp64 & fp64 & 7.2 & 171.7 & 7.50 \\
 &  fp32 & fp32 & 7.8 & 170.6 & 7.45 \\
 &  fp32 & fp16 & 7.6 & 220.3 & 9.62 \\
 &  fp32 & bf16 & 6.7 & 220.0 & 9.61 \\
\hline
\end{tabular}
\end{table}

\begin{figure}[H]
    \centering
    \begin{subfigure}{0.48\linewidth}
        \centering
        \includegraphics[width=\linewidth]{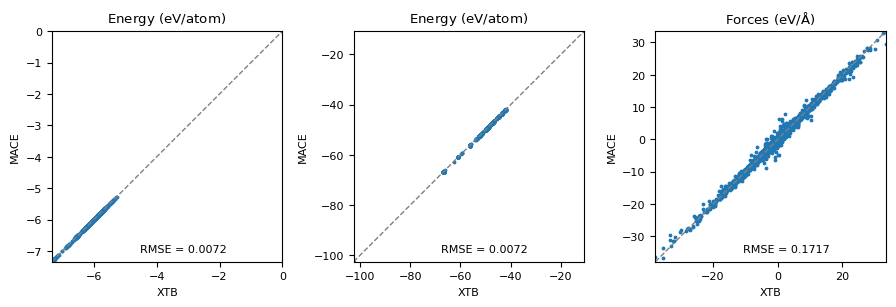}
        \caption{Testing RMSE for test data fp64}
        \label{fig:rmse_fp64}
    \end{subfigure}
    \hfill
    \begin{subfigure}{0.48\linewidth}
        \centering
        \includegraphics[width=\linewidth]{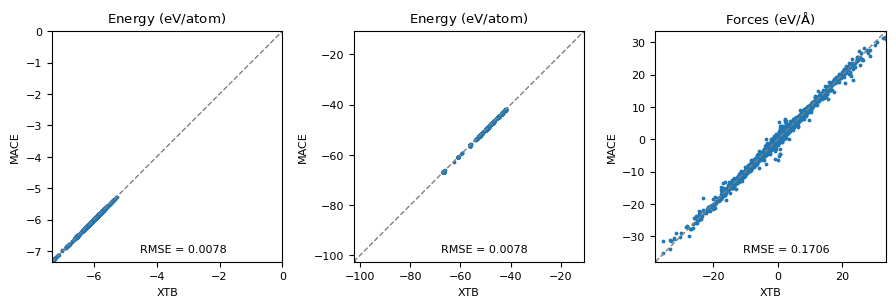}
        \caption{Testing RMSE for test data fp32}
        \label{fig:rmse_fp32}
    \end{subfigure}
    
    \vspace{0.5cm}
    \begin{subfigure}{0.48\linewidth}
        \centering
        \includegraphics[width=\linewidth]{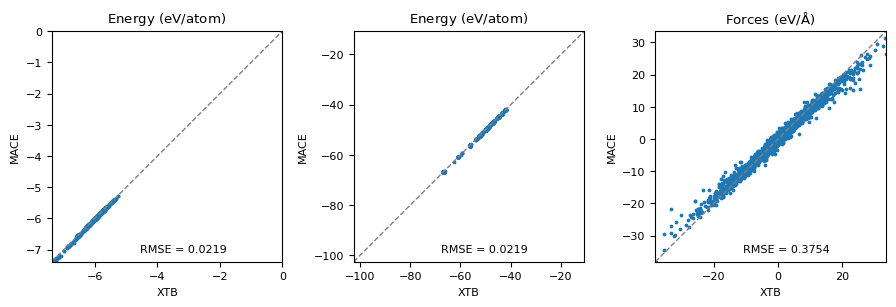}
        \caption{Testing RMSE for test data fp32 and fp16 linear}
        \label{fig:rmse_fp16}
    \end{subfigure}
    \hfill
    \begin{subfigure}{0.48\linewidth}
        \centering
        \includegraphics[width=\linewidth]{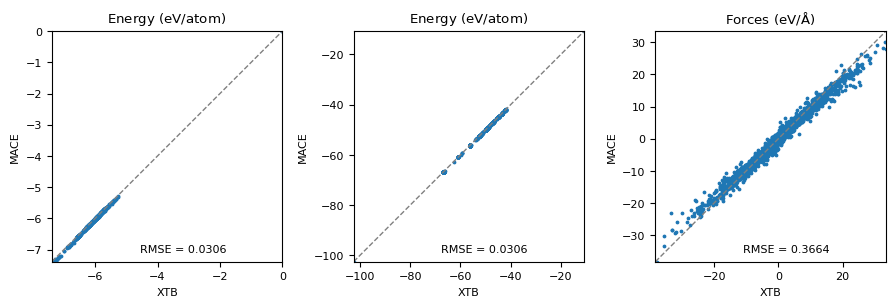}
        \caption{Testing RMSE for test data fp32 and bf16 linear}
        \label{fig:rmse_bf16}
    \end{subfigure}

    \caption{Comparison of RMSE across different numerical precision settings.}
    \label{fig:rmse_comparison}
\end{figure}





\section{Chapter's summary}

We profiled MACE end-to-end and per block, compared the e3nn and cuEquivariance (cuEq) backends, and quantified mixed-precision accuracy–throughput trade-offs. Switching to cuEq cuts inference latency by \(\sim\)3\(\times\) (122.9 vs.\ 366.1\,ms) with tighter variability; after a one-off \(\sim\)30\,s warm-up, steady-state steps are \(\sim\)48\,ms. Profiler totals show a dispatcher/launch-bound regime (CPU \(\approx\)57\% vs.\ CUDA \(\approx\)43\%), with cost concentrated in \emph{SymmetricContractions}, \emph{Linear Interaction}, and \emph{(Skip/Conv)TensorProducts}. Thus, operator fusion, batching/shape bucketing, and graph capture are the main levers, while precision downgrades shrink the remaining math.

Mixed-precision inference on cuEq yields 3.0–3.3\(\times\) speedups with a constant \(\sim\)0.9\,eV energy offset (ppm-level) and preserved MD ensembles; FP32+BF16 gives the best stability among reduced-precision modes. In NPT water, all precisions reproduce FP64 within noise (temperature \(<0.03\%\), density \(<0.29\%\)); throughput improves \(\sim\)2\(\times\) (FP32) and \(\sim\)4.3\(\times\)

\nomenclature[z-MACE]{MACE}{Message Passing Atomic Convolutional Encoder (equivariant ML force field)}
\nomenclature[z-e3nn]{e3nn}{PyTorch library for $E(3)$-/SO(3)-equivariant neural networks}
\nomenclature[z-cuEq]{cuEq}{NVIDIA cuEquivariance library (fused SO(3) tensor ops on GPU)}
\nomenclature[z-GPU]{GPU}{Graphics Processing Unit}
\nomenclature[z-CPU]{CPU}{Central Processing Unit}
\nomenclature[z-CUDA]{CUDA}{Compute Unified Device Architecture (NVIDIA GPU platform)}
\nomenclature[z-cuBLAS]{cuBLAS}{CUDA Basic Linear Algebra Subprograms}
\nomenclature[z-GEMM]{GEMM}{General Matrix–Matrix Multiply}
\nomenclature[z-AMP]{AMP}{Automatic Mixed Precision}
\nomenclature[z-DDP]{DDP}{Distributed Data Parallel}
\nomenclature[z-ASE]{ASE}{Atomic Simulation Environment}
\nomenclature[z-MD]{MD}{Molecular Dynamics}
\nomenclature[z-NVT]{NVT}{Canonical ensemble (constant $N,V,T$; often Langevin or Nosé–Hoover)}
\nomenclature[z-NPT]{NPT}{Isothermal–isobaric ensemble (constant $N,P,T$)}
\nomenclature[z-RDF]{RDF}{Radial Distribution Function}
\nomenclature[z-SH]{SH}{Spherical Harmonics}
\nomenclature[z-CG]{CG}{Clebsch–Gordan coefficients}
\nomenclature[z-TP]{TP}{Tensor product (SO(3) equivariant)}
\nomenclature[z-MLP]{MLP}{Multi-Layer Perceptron}
\nomenclature[z-FP32]{FP32}{IEEE-754 single precision (32-bit float)}
\nomenclature[z-FP16]{FP16}{IEEE-754 half precision (16-bit float)}
\nomenclature[z-BF16]{BF16}{bfloat16 (8-bit exponent, 7-bit mantissa)}
\nomenclature[z-TF32]{TF32}{TensorFloat-32 (FP32 range with 10-bit mantissa for Tensor Cores)}
\nomenclature[z-FLOP]{FLOP}{Floating-point operation}
\nomenclature[z-TFLOPS]{TFLOPS}{Tera–FLOPs per second}
\nomenclature[z-SM]{SM}{Streaming Multiprocessor (GPU core complex)}
\nomenclature[z-PBC]{PBC}{Periodic Boundary Conditions}
\nomenclature[z-RMSE]{RMSE}{Root Mean Squared Error}
\nomenclature[z-MAE]{MAE}{Mean Absolute Error}
\nomenclature[z-MIG]{MIG}{Multi-Instance GPU (A100/H100 partitioning)}

\nomenclature[a-B]{$B$}{Batch size (number of graphs/configurations per step)}
\nomenclature[a-N]{$N$}{Number of atoms (nodes) in a configuration}
\nomenclature[a-E]{$E$}{Total energy (eV) or per-atom energy as specified}
\nomenclature[a-F]{$\mathbf{F}$}{Forces (eV/\AA) on atoms}
\nomenclature[a-r]{$r$}{Interatomic distance (\,\AA)}
\nomenclature[a-rmax]{$r_{\max}$}{Cutoff radius for neighbor list (\,\AA)}
\nomenclature[a-Lmax]{$L_{\max}$}{Maximum equivariance order used in messages}
\nomenclature[a-Z]{$Z$}{Atomic number; also element embedding index}
\nomenclature[a-C]{$C$}{Channel count (feature multiplicity)}
\nomenclature[a-nb]{$n_{\text{Bessel}}$}{Size of radial (Bessel) basis}
\nomenclature[a-t]{$t$}{Time (simulation steps or seconds, per context)}
\nomenclature[a-T]{$T$}{Temperature (K)}
\nomenclature[a-p]{$p$}{Pressure (bar) or polynomial cutoff exponent (per context)}
\nomenclature[a-kB]{$k_{\mathrm{B}}$}{Boltzmann constant}
\nomenclature[a-Lcal]{$\mathcal{L}$}{Loss function}

\nomenclature[g-ell]{$\ell$}{Spherical harmonic / irreducible representation degree}
\nomenclature[g-theta]{$\theta$}{Angle; also used in cosine-similarity of forces}
\nomenclature[g-DeltaE]{$\Delta E$}{Energy difference vs.\ baseline (eV)}
\nomenclature[g-DeltaF]{$\Delta F$}{Force difference vs.\ baseline (eV/\AA)}
\nomenclature[g-eps]{$\epsilon$}{Machine epsilon; also small tolerance}
\nomenclature[g-sigma]{$\sigma$}{Standard deviation}
\nomenclature[g-gamma]{$\gamma$}{Friction coefficient in Langevin thermostat}

\nomenclature[x-AA]{$\text{\AA}$}{Angstrom (unit of length)}
\nomenclature[x-eV]{eV}{Electronvolt (unit of energy)}
\nomenclature[x-norm]{$\lVert\cdot\rVert$}{Vector or tensor norm (context-dependent)}
\nomenclature[x-otimes]{$\otimes$}{Kronecker/tensor product}
\nomenclature[x-odot]{$\odot$}{Hadamard (elementwise) product}
\nomenclature[x-R3]{$\mathbb{R}^3$}{Three-dimensional Euclidean space}
\nomenclature[x-SO3]{$\mathrm{SO}(3)$}{Special orthogonal group in 3D (rotations)}
\nomenclature[x-graph]{$\mathcal{G}=(\mathcal{V},\mathcal{E})$}{Atomic graph: vertices (atoms) and edges (neighbors)}

\nomenclature[r-top]{$(\cdot)^{\top}$}{Transpose}
\nomenclature[r-k]{$(\cdot)^{(k)}$}{Quantity at layer/iteration $k$}
\nomenclature[r-tc]{$(\cdot)^{\mathrm{TC}}$}{Computed on Tensor Cores (mixed precision path)}

\nomenclature[s-i]{$i,j$}{Atom (node) or index labels}
\nomenclature[s-atom]{\text{atom}}{Per-atom quantity (e.g.\ $E_{\text{atom}}$)}
\nomenclature[s-max]{\text{max}}{Maximal value/index (e.g.\ $L_{\max}$, $r_{\max}$)}
\nomenclature[s-neigh]{\text{nbr}}{Neighbor index/list}
\nomenclature[s-bessel]{\text{Bessel}}{Bessel basis index}
\nomenclature[s-sh]{\text{SH}}{Spherical harmonic index}


\chapter{Conclusion}
\label{chapter:conclusion}

\ifpdf
    \graphicspath{{Chapter4/Figs/Raster/}{Chapter4/Figs/PDF/}{Chapter4/Figs/}}
\else
    \graphicspath{{Chapter4/Figs/Vector/}{Chapter4/Figs/}}
\fi

\section[Work summary]{A very limited scope awaiting for kernel improvements}

This thesis mapped where MACE spends time, quantified the benefit of switching from e3nn to cuEquivariance (cuEq), and tested how reduced numerical precision affects accuracy and stability. We found that cuEq lowers inference latency by \(\sim 3\times\) and that the workload is primarily dispatcher/launch-bound rather than math-bound. Mixed-precision \emph{inference} --- keeping FP32 accumulators while casting only Linear layers to FP16/BF16 --- preserves thermodynamic behaviour in NVT/NPT MD within noise and yields substantial throughput gains, whereas training with half-precision Linear weights harms force generalisation. We also showed that mixing e3nn and cuEq modules without explicit layout/weight adapters produces large, stationary feature mismatches, so backends must be used homogeneously or bridged carefully. The main limitations were hardware (no TF32 support on our GPU) and the current cuEq kernel set, which lacks native FP16/BF16 implementations for several operators; these constraints narrowed the scope of precision ablations.

\section{Future Work}

A natural next step is to repeat all experiments on Ampere/Hopper GPUs (A100/H100/H200) to leverage TF32/BF16 tensor cores and re-benchmark speed/accuracy. On the systems side, reducing launch overhead via pinned memory and non-blocking transfers, shape bucketing/prefetching, \texttt{torch.compile}, and CUDA Graphs should unlock accuracy-neutral speedups. On the kernel side, implementing FP16/BF16 paths with FP32 accumulation for cuEq tensor products, symmetric contractions, and linears would enable end-to-end low-precision trials beyond bespoke Linear-only casts. Finally, adding explicit cuEq\(\leftrightarrow\)e3nn adapters and broadening validation to larger, more diverse datasets and longer MD will turn these findings into robust, general recommendations.


\begin{spacing}{0.9}



\bibliographystyle{plainnat}
\cleardoublepage
\bibliography{References/references} 



\end{spacing}

\begin{appendices} 

\chapter{Hardware Specification} 

\section*{NVIDIA A100 Tensor Core GPU Characteristics}
\label{appendix:nvidia}

\begin{table}[ht]
\centering
\caption{NVIDIA A100 80GB SXM Specifications}
\label{tab:a100nvidia}
\begin{tabular}{l|l}
\toprule
\textbf{Specification} & \textbf{A100 80GB SXM} \\
\midrule
FP64 & 9.7 TFLOPS \\
\hline
FP64 Tensor Core & 19.5 TFLOPS \\
\hline
FP32 & 19.5 TFLOPS \\
\hline
Tensor Float 32 (TF32) & 312 TFLOPS* \\
\hline
BFLOAT16 Tensor Core & 624 TFLOPS* \\
\hline
FP16 Tensor Core & 624 TFLOPS* \\
\hline
INT8 Tensor Core & 1248 TOPS* \\
\hline
GPU Memory & 80GB HBM2e \\
\hline
GPU Memory Bandwidth & 2,039 GB/s \\
\hline
Max Thermal Design Power (TDP) & 400W*** \\
\hline
Multi-Instance GPU & Up to 7 MIGs @ 10GB \\
\hline
Form Factor & SXM \\
\hline
Interconnect & NVLink: 600 GB/s \\
 & PCIe Gen4: 64 GB/s \\
\hline
 
Server Options & \begin{tabular}[c]{@{}l@{}}
NVIDIA HGX A100 - Partner and\\
NVIDIA-Certified Systems with\\
4, 8, or 16 GPUs\\
NVIDIA DGX A100 with 8 GPUs
\end{tabular} \\
\end{tabular}
\vspace{0.5em}
\begin{flushleft}
\footnotesize * With sparsity \\
*** 400W TDP for standard configuration. HGX A100-80GB CTS (Custom Thermal Solution) \\
SKU can support TDPs up to 500W.
\end{flushleft}
\end{table}

\section*{NVIDIA TURING RTX 2080 Ti GPU Characteristics}
\label{appendix:rtx}

\begin{table}[h!]
\centering
\caption{NVIDIA GeForce RTX 2080 (Turing) — key specifications \citep{NVIDIA_2017}}
\renewcommand{\arraystretch}{1.12}
\begin{tabularx}{\textwidth}{@{}lX@{}}
\toprule
\textbf{GPU Feature} & \textbf{GeForce RTX 2080} \\
\midrule
Architecture & Turing \\
GPCs & 6 \\
TPCs & 23 \\
SMs & 46 \\
CUDA Cores / SM & 64 \\
CUDA Cores / GPU & 2944 \\
Tensor Cores / SM & 8 \\
Tensor Cores / GPU & 368 \\
RT Cores & 46 \\
GPU Base Clock (MHz) (Ref / FE) & 1515 / 1515 \\
GPU Boost Clock (MHz) (Ref / FE) & 1710 / 1800 \\
RTX-OPS (Tera-OPS) (Ref / FE) & 57 / 60 \\
Rays Cast (Giga Rays/s) (Ref / FE) & 8 / 8 \\
Peak FP32 TFLOPS (Ref / FE) & 10 / 10.6 \\
Peak INT32 TIPS (Ref / FE) & 10 / 10.6 \\
Peak FP16 TFLOPS (Ref / FE) & 20.1 / 21.2 \\
Peak FP16 Tensor TFLOPS (FP16 Accum) (Ref / FE) & 80.5 / 84.8 \\
Peak FP16 Tensor TFLOPS (FP32 Accum) (Ref / FE) & 40.3 / 42.4 \\
Peak INT8 Tensor TOPS (Ref / FE) & 161.1 / 169.6 \\
Peak INT4 Tensor TOPS (Ref / FE) & 322.2 / 339.1 \\
Frame Buffer (Size / Type) & 8192 MB GDDR6 \\
Memory Interface & 256-bit \\
Memory Clock (Data Rate) & 14 Gbps \\
Memory Bandwidth (GB/s) & 448 \\
ROPs & 64 \\
Texture Units & 184 \\
Texel Fill-rate (GTexels/s) (Ref / FE) & 314.6 / 331.2 \\
L2 Cache Size & 4096 KB \\
Register File Size / SM & 256 KB \\
Register File Size / GPU & 11776 KB \\
TDP (Ref / FE) & 215 / 225 W \\
Transistor Count & 13.6 Billion \\
Die Size & 545 mm$^2$ \\
Manufacturing Process & 12 nm FFN \\
\bottomrule
\end{tabularx}
\end{table}

\chapter{Dataset Setup} 

\section{Comparing E3NN with Cuequivariance setup}
\label{appendix:setup}

\sisetup{
  detect-weight=true,
  detect-inline-weight=math,
  table-number-alignment = center,
  table-figures-integer = 3,
  table-figures-decimal  = 3,
  round-mode=places,
  round-precision=3
}

\newcommand{\datasetname}{\texttt{carbon.xyz}}
\newcommand{\datasetpath}{\texttt{Experiments/Official MACE notebook/data/carbon.xyz}}
\newcommand{\numstructures}{\textit{N/A}} 
\newcommand{\elementslist}{C}             
\newcommand{\atomsperstruct}{\textit{N/A}}
\newcommand{\cutoff}{6.0}                 
\newcommand{\avgnn}{8}                    
\newcommand{\dtype}{float64}              
\newcommand{\radialtype}{bessel}
\newcommand{\numbessel}{8}
\newcommand{\numpoly}{6}
\newcommand{\numinteractions}{2}
\newcommand{\correlation}{3}
\newcommand{\defaultirreps}{\texttt{16x0e + 16x1o}}
\newcommand{\defaultell}{3}
\newcommand{\defaultbs}{32}
\newcommand{\numiters}{100}
\newcommand{\warmupiters}{100}

\begin{table}[h]
  \centering
  \caption[Dataset summary for \datasetname]{Dataset summary for \datasetname. Values reflecting the current benchmark configuration are shown in \emph{italics} where runtime-derived.}
  \label{tab:dataset-carbon}
  \begin{tabular}{@{}ll@{}}
    \toprule
    \textbf{Field} & \textbf{Value} \\
    \midrule
    File (path)                 & \datasetpath \\
    Chemical elements           & \elementslist \\
    Number of structures        & \numstructures \\
    Atoms per structure         & \atomsperstruct \\
    Distance cutoff $r_{\max}$ (\si{\angstrom}) & \cutoff \\
    Avg.\ neighbors per atom    & \avgnn \\
    Units (energy, length)      & eV, \si{\angstrom} \\
    Notes                       & Single-species carbon benchmark used for MACE inference profiling. \\
    \bottomrule
  \end{tabular}
\end{table}

\begin{table}[h]
  \centering
  \caption{Benchmark configuration (mirrors the script settings).}
  \label{tab:bench-config}
  \begin{tabular}{@{}ll@{}}
    \toprule
    \textbf{Parameter} & \textbf{Setting} \\
    \midrule
    Default dtype                         & \texttt{\dtype} \\
    Hidden irreps                         & \defaultirreps \\
    Max angular momentum $\ell_{\max}$    & \defaultell \\
    Batch size                            & \defaultbs \\
    Iterations (timed)                    & \numiters \\
    Warmup iterations                     & \warmupiters \\
    Interaction blocks                    & \numinteractions~(\texttt{RealAgnosticResidualInteractionBlock}) \\
    Correlation order                     & \correlation \\
    Radial basis / cutoff polynomial      & \radialtype{} / \numpoly \\
    \# Bessel functions                   & \numbessel \\
    Cutoff $r_{\max}$ (\si{\angstrom})    & \cutoff \\
    Avg.\ neighbors (assumed)             & \avgnn \\
    \bottomrule
  \end{tabular}
\end{table}

\begin{table}[h]
  \centering
  \caption[MACE inference timings on \datasetname{}]{MACE inference timings on \datasetname{}: E3NN vs cuEquivariance (CUET).
  Mean wall-time per call in milliseconds; speedup defined as E3NN/CUET. 
  All runs are on a \texttt{cuda} device with hidden irreps set to \texttt{16x0e + 16x1o}.}
  \label{tab:results-e3nn-cueq}

  \[
  \begin{array}{c c c c c}
    \hline
    \text{Batch Size} & \ell_{\max} & \text{E3NN (ms)} & \text{CUET (ms)} & \text{Speedup (E3NN$\rightarrow$CUET)} \\
    \hline
     1  & 1 &  85.699 &  77.871 & 1.101 \\
     2  & 1 &  86.412 &  79.479 & 1.087 \\
     4  & 1 &  89.485 &  78.651 & 1.138 \\
     8  & 1 &  90.484 &  81.194 & 1.114 \\
    16  & 1 &  91.211 &  81.181 & 1.124 \\
    32  & 1 &  91.135 &  82.184 & 1.109 \\
     1  & 2 & 292.600 &  96.096 & 3.045 \\
     2  & 2 & 292.791 & 102.010 & 2.870 \\
     4  & 2 & 292.914 & 100.341 & 2.919 \\
     8  & 2 & 293.133 & 101.692 & 2.883 \\
    16  & 2 & 293.080 &  99.826 & 2.936 \\
    32  & 2 & 292.903 & 100.448 & 2.916 \\
     1  & 3 & 379.503 & 127.033 & 2.987 \\
     2  & 3 & 378.398 & 130.243 & 2.905 \\
     4  & 3 & 378.025 & 131.693 & 2.871 \\
     8  & 3 & 377.639 & 126.088 & 2.995 \\
    16  & 3 & 377.394 & 126.238 & 2.990 \\
    32  & 3 & 377.401 & 130.965 & 2.882 \\
    \hline
  \end{array}
  \]

  \vspace{0.25em}
  \footnotesize
  \textbf{Notes.} CUET entries appear only when \texttt{cuequivariance} is available on \texttt{cuda}. 
  Speedup is reported only when both backends were timed under identical settings. Values rounded to 3 d.p.
\end{table}

Speedup definition for reference in your narrative:
\[
  \text{Speedup} = \frac{\text{E3NN mean ms}}{\text{CUET mean ms}}.
\]

\section{MACE architecture breakdown}
\label{ap:mab}

\begin{table}[t]
\centering
\small
\caption[ScaleShiftMACE architecture breakdown]{ScaleShiftMACE architecture breakdown. Shapes follow the model print; “Params” are learnable weights (\texttt{weight\_numel}) when available.}
\label{tab:mace-arch-breakdown}
\resizebox{\textwidth}{!}{%
\begin{tabular}{@{} l l l l r @{}} 
\toprule
\textbf{Block} & \textbf{Submodule} & \textbf{Op / Irreps} & \textbf{Key shape(s)} & \textbf{Params} \\
\midrule
Node Embedding & \texttt{LinearNodeEmbeddingBlock.linear} & Linear 
& $[1280:1\times(10,128)] \cdot [10:1\times(1,10)] \rightarrow [128:1\times(1,128)]$
& 1{,}280 \\
\addlinespace
Radial Embedding & BesselBasis / Cutoff & Bessel / Poly cutoff 
& $r_{\max}{=}6.0,\ n_{\text{basis}}{=}8$ 
& -- \\
\addlinespace
Sph.\ Harmonics & SphericalHarmonics & $Y_\ell$ 
& -- 
& -- \\
\addlinespace
Atomic Energies & AtomicEnergiesBlock & Table 
& per-$Z$ energies 
& -- \\
\addlinespace
\multicolumn{5}{@{}l}{\textbf{Interactions[0]} RealAgnosticInteractionBlock} \\
& \texttt{linear\_up} & Linear 
& $[16384:(128,128)] \rightarrow [128]$
& 16{,}384 \\
& \texttt{conv\_tp} & Tensor Product 
& $a{=}[4{\times}128],\ b{=}[1{\times}128] \rightarrow D{=}[16{\times}128]$
& -- \\
& \texttt{conv\_tp\_weights} & FCN $[8,64,64,64,512]$ 
& MLP for TP weights 
& \emph{est.} $\sim$42k \\
& \texttt{linear} & Linear 
& $[4{\times}(128,128)] \rightarrow [2048]$
& 65{,}536 \\
& \texttt{skip\_tp} & FullyConnectedTP 
& $[4{\times}(128,10,128)]$
& 655{,}360 \\
\addlinespace
\multicolumn{5}{@{}l}{\textbf{Interactions[1]} RealAgnosticResidualInteractionBlock} \\
& \texttt{linear\_up} & Linear 
& $[2{\times}(128,128)] \rightarrow [512]$
& 32{,}768 \\
& \texttt{conv\_tp} & Tensor Product 
& $a{=}[10{\times}128],\ b{=}[4{\times}128] \rightarrow D{=}[40{\times}128]$
& -- \\
& \texttt{conv\_tp\_weights} & FCN $[8,64,64,64,1280]$ 
& MLP for TP weights 
& \emph{est.} $\sim$92k \\
& \texttt{linear} & Linear 
& $[10{\times}(128,128)] \rightarrow [2048]$
& 163{,}840 \\
& \texttt{skip\_tp} & FullyConnectedTP 
& $[1{\times}(128,10,128)]$
& 163{,}840 \\
\addlinespace
\multicolumn{5}{@{}l}{\textbf{Products and Readouts}} \\
& \texttt{ProductBasis[0].linear} & Linear 
& $[2{\times}(128,128)] \rightarrow [512]$
& 32{,}768 \\
& \texttt{ProductBasis[1].linear} & Linear 
& $[1{\times}(128,128)] \rightarrow [128]$
& 16{,}384 \\
& \texttt{LinearReadoutBlock.linear} & Linear 
& $[128 \rightarrow 1]$
& 128 \\
& \texttt{NonLinearReadout} & MLP 
& $128 \rightarrow 16 \rightarrow 1$
& 2{,}064 \\
\addlinespace
Scale/Shift & ScaleShiftBlock & Affine 
& $E \mapsto sE + \Delta$
& -- \\
\bottomrule
\end{tabular}
}%
\end{table}

\section{Inference profiler setup for figuring out which blocks is computing expensive}
\label{appendix:setup_2}

\subsection{MACE-Off 24}

\begin{table}[h]
\centering
\caption{Key features of the MACE-OFF24(M) model.}
\begin{tabular}{|l|l|}
\hline
\textbf{Feature} & \textbf{Details} \\
\hline
Model variant & MACE-OFF24 (Medium, short-range ML potential) \\
\hline
Cutoff radius & 6.0 \AA \\
\hline
Training dataset & SPICE v2 (includes solvated PubChem molecules, \\
                 & amino acid--ligand pairs, water clusters up to $\sim$90 atoms) \\
\hline
Performance highlights & Accurate water density across $T$ (within 2\% of experiment); \\
                      & peptide folding (Ala$_3$, Ala$_{15}$) free energy surfaces \\
\hline
Applications & Organic molecular liquids, condensed-phase biomolecular systems \\
\hline
\end{tabular}
\end{table}

\subsection{Diamon Carbon Unit cell}

\begin{table}[h]
\centering
\caption{Atomistic setup for diamond supercell simulation.}
\begin{tabular}{|c|l|l|}
\hline
\textbf{Step} & \textbf{Description} & \textbf{Details / Reference} \\
\hline
1 & Build unit cell of diamond & Cubic diamond structure of carbon, lattice constant $a = 3.567$ \AA \\
\hline
2 & Define supercell size & Supercell size set to $N \times N \times N$, with $N = 3$ in this work \\
\hline
3 & Create supercell & Final supercell contains $8 \times N^3 = 216$ atoms (for $N=3$) \\
\hline
\end{tabular}
\end{table}

\subsection{Results for Profiling inference for blocks inspection}
\label{appendix:result_profiling_inference}

\begin{table}[h]
\centering
\scriptsize
\setlength{\tabcolsep}{3pt}
\renewcommand{\arraystretch}{1.1}
\begin{adjustbox}{max width=\textwidth}
\begin{tabular}{lrrrrrrrrrrrrrr}
\toprule
\textbf{Name} & \textbf{Self CPU \%} & \textbf{Self CPU} & \textbf{CPU total \%} & \textbf{CPU total} & \textbf{CPU time avg} & \textbf{Self CUDA} & \textbf{Self CUDA \%} & \textbf{CUDA total} & \textbf{CUDA time avg} & \textbf{CPU Mem} & \textbf{Self CPU Mem} & \textbf{CUDA Mem} & \textbf{Self CUDA Mem} & \textbf{\# of Calls} \\
\midrule
MACE/SphericalHarmonics & 0.00\% & 0.000us & 0.00\% & 0.000us & 0.000us & 91.464ms & 6.26\% & 91.464ms & 896.709us & 0 B & 0 B & 0 B & 0 B & 102 \\
MACE/RadialEmbedding    & 0.00\% & 0.000us & 0.00\% & 0.000us & 0.000us & 58.728ms & 4.02\% & 58.728ms & 575.768us & 0 B & 0 B & 0 B & 0 B & 102 \\
MACE/NodeEmbedding      & 0.00\% & 0.000us & 0.00\% & 0.000us & 0.000us & 24.921ms & 1.70\% & 24.921ms & 244.322us & 0 B & 0 B & 0 B & 0 B & 102 \\
MACE/RadialEmbedding    & 0.45\% & 20.607ms & 1.46\% & 66.804ms & 654.942us & 0.000us & 0.00\% & 13.572ms & 133.062us & 0 B & 0 B & 73.41 MB & -27.99 MB & 102 \\
MACE/SphericalHarmonics & 0.45\% & 20.519ms & 2.22\% & 101.859ms & 998.621us & 0.000us & 0.00\% & 6.213ms & 60.910us & 0 B & 0 B & 63.35 MB & -408.00 KB & 102 \\
MACE/NodeEmbedding      & 0.52\% & 23.722ms & 1.06\% & 48.737ms & 477.818us & 0.000us & 0.00\% & 968.993us & 9.500us & 0 B & 0 B & 1.59 MB & -1.79 MB & 102 \\
MACE/AtomicEnergies     & 0.27\% & 12.491ms & 0.55\% & 25.067ms & 122.876us & 0.000us & 0.00\% & 642.910us & 3.152us & 0 B & 0 B & 102.00 KB & 0 B & 204 \\
MACE/AtomicEnergies     & 0.00\% & 0.000us & 0.00\% & 0.000us & 0.000us & 642.910us & 0.04\% & 642.910us & 3.152us & 0 B & 0 B & 0 B & 0 B & 204 \\
\bottomrule
\end{tabular}
\end{adjustbox}
\caption[Profiler summary for MACE Embeddings]{Profiler summary for MACE Embeddings. Times/units are as reported by the profiler.}
\label{tab:mace-profiler}
\end{table}

\begin{table}[h]
\centering
\scriptsize
\setlength{\tabcolsep}{3pt}
\renewcommand{\arraystretch}{1.1}
\begin{adjustbox}{max width=\textwidth}
\begin{tabular}{lrrrrrrrrrrrrrr}
\toprule
\textbf{Name} & \textbf{Self CPU \%} & \textbf{Self CPU} & \textbf{CPU total \%} & \textbf{CPU total} & \textbf{CPU time avg} & \textbf{Self CUDA} & \textbf{Self CUDA \%} & \textbf{CUDA total} & \textbf{CUDA time avg} & \textbf{CPU Mem} & \textbf{Self CPU Mem} & \textbf{CUDA Mem} & \textbf{Self CUDA Mem} & \textbf{\# of Calls} \\
\midrule
MACE/Interaction[1]/Main & 0.28\% & 13.017ms & 10.15\% & 465.217ms & 4.561ms & 0.000us & 0.00\% & 309.471ms & 3.034ms & 5.58 KB & 816 B & 2.97 GB & -114.75 MB & 102 \\
MACE/Interaction[0]/Main & 0.35\% & 15.896ms & 11.54\% & 528.589ms & 5.182ms & 0.000us & 0.00\% & 238.041ms & 2.334ms & 5.58 KB & 816 B & 2.05 GB & -102.00 MB & 102 \\
MACE/Interaction[0]/SkipTensorProduct & 0.00\% & 0.000us & 0.00\% & 0.000us & 0.000us & 180.552ms & 12.35\% & 180.552ms & 1.770ms & 0 B & 0 B & 0 B & 0 B & 102 \\
MACE/Interaction[1]/ConvTPWeights & 0.00\% & 0.000us & 0.00\% & 0.000us & 0.000us & 156.587ms & 10.71\% & 156.587ms & 1.535ms & 0 B & 0 B & 0 B & 0 B & 102 \\
MACE/Interaction[1]/ConvTPWeights & 0.32\% & 14.560ms & 0.92\% & 42.319ms & 414.889us & 0.000us & 0.00\% & 146.769ms & 1.439ms & 4.78 KB & 4.78 KB & 2.94 GB & -1.15 GB & 102 \\
MACE/Interaction[1]/Linear\_Interaction & 0.00\% & 0.000us & 0.00\% & 0.000us & 0.000us & 122.091ms & 8.35\% & 122.091ms & 1.197ms & 0 B & 0 B & 0 B & 0 B & 102 \\
MACE/Interaction[0]/SkipTensorProduct & 1.13\% & 51.570ms & 4.16\% & 190.394ms & 1.867ms & 0.000us & 0.00\% & 109.916ms & 1.078ms & 0 B & 0 B & 637.70 MB & -93.11 MB & 102 \\
MACE/Interaction[0]/ConvTPWeights & 0.00\% & 0.000us & 0.00\% & 0.000us & 0.000us & 99.112ms & 6.78\% & 99.112ms & 971.689us & 0 B & 0 B & 0 B & 0 B & 102 \\
MACE/Interaction[0]/Linear\_Interaction & 0.00\% & 0.000us & 0.00\% & 0.000us & 0.000us & 98.881ms & 6.76\% & 98.881ms & 969.417us & 0 B & 0 B & 0 B & 0 B & 102 \\
MACE/Interaction[1]/Linear\_Interaction & 1.33\% & 61.061ms & 4.64\% & 212.526ms & 2.084ms & 0.000us & 0.00\% & 89.208ms & 874.588us & 0 B & 0 B & 26.00 MB & -165.75 MB & 102 \\
MACE/Interaction[0]/ConvTPWeights & 0.42\% & 19.035ms & 1.23\% & 56.486ms & 553.782us & 0.000us & 0.00\% & 81.766ms & 801.624us & 4.78 KB & 4.78 KB & 1.43 GB & -1.15 GB & 102 \\
MACE/Interaction[1]/ConvTensorProduct & 0.00\% & 0.000us & 0.00\% & 0.000us & 0.000us & 47.832ms & 3.27\% & 47.832ms & 468.943us & 0 B & 0 B & 0 B & 0 B & 102 \\
MACE/Interaction[1]/ConvTensorProduct & 0.22\% & 10.267ms & 1.56\% & 71.339ms & 699.406us & 0.000us & 0.00\% & 47.687ms & 467.521us & 0 B & 0 B & 63.75 MB & 0 B & 102 \\
MACE/Interaction[1]/LinearUp\_Interaction & 0.00\% & 0.000us & 0.00\% & 0.000us & 0.000us & 43.221ms & 2.96\% & 43.221ms & 423.732us & 0 B & 0 B & 0 B & 0 B & 102 \\
MACE/Interaction[1]/SkipTensorProduct & 0.00\% & 0.000us & 0.00\% & 0.000us & 0.000us & 35.367ms & 2.42\% & 35.367ms & 346.732us & 0 B & 0 B & 0 B & 0 B & 102 \\
MACE/Interaction[0]/Linear\_Interaction & 0.84\% & 38.433ms & 2.56\% & 117.277ms & 1.150ms & 0.000us & 0.00\% & 33.501ms & 328.444us & 0 B & 0 B & 25.70 MB & -51.00 MB & 102 \\
MACE/Interaction[0]/LinearUp\_Interaction & 0.00\% & 0.000us & 0.00\% & 0.000us & 0.000us & 22.709ms & 1.55\% & 22.709ms & 222.636us & 0 B & 0 B & 0 B & 0 B & 102 \\
MACE/Interaction[0]/ConvTensorProduct & 0.00\% & 0.000us & 0.00\% & 0.000us & 0.000us & 16.503ms & 1.13\% & 16.503ms & 161.799us & 0 B & 0 B & 0 B & 0 B & 102 \\
MACE/Interaction[1]/SkipTensorProduct & 0.43\% & 19.837ms & 1.21\% & 55.576ms & 544.861us & 0.000us & 0.00\% & 12.605ms & 123.582us & 0 B & 0 B & 1.64 MB & -19.23 MB & 102 \\
MACE/Interaction[1]/LinearUp\_Interaction & 0.46\% & 21.001ms & 1.21\% & 55.588ms & 544.985us & 0.000us & 0.00\% & 12.579ms & 123.325us & 0 B & 0 B & 6.47 MB & -12.75 MB & 102 \\
MACE/Interaction[0]/ConvTensorProduct & 0.23\% & 10.715ms & 1.91\% & 87.273ms & 855.615us & 0.000us & 0.00\% & 10.232ms & 100.312us & 0 B & 0 B & 25.50 MB & 0 B & 102 \\
MACE/Interaction[0]/LinearUp\_Interaction & 0.44\% & 20.037ms & 0.94\% & 43.053ms & 422.092us & 0.000us & 0.00\% & 2.004ms & 19.645us & 0 B & 0 B & 1.59 MB & -3.24 MB & 102 \\
MACE/Interaction[1]/Reshape & 0.14\% & 6.450ms & 0.28\% & 12.742ms & 124.923us & 0.000us & 0.00\% & 419.169us & 4.109us & 0 B & 0 B & 25.50 MB & 0 B & 102 \\
MACE/Interaction[1]/Reshape & 0.00\% & 0.000us & 0.00\% & 0.000us & 0.000us & 419.169us & 0.03\% & 419.169us & 4.109us & 0 B & 0 B & 0 B & 0 B & 102 \\
MACE/Interaction[0]/Reshape & 0.17\% & 7.652ms & 0.34\% & 15.453ms & 151.503us & 0.000us & 0.00\% & 418.683us & 4.105us & 0 B & 0 B & 25.50 MB & 0 B & 102 \\
MACE/Interaction[0]/Reshape & 0.00\% & 0.000us & 0.00\% & 0.000us & 0.000us & 418.683us & 0.03\% & 418.683us & 4.105us & 0 B & 0 B & 0 B & 0 B & 102 \\
MACE/Interaction[1]/Main & 0.00\% & 0.000us & 0.00\% & 0.000us & 0.000us & 203.846us & 0.01\% & 203.846us & 1.998us & 0 B & 0 B & 0 B & 0 B & 102 \\
MACE/Interaction[0]/Main & 0.00\% & 0.000us & 0.00\% & 0.000us & 0.000us & 203.589us & 0.01\% & 203.589us & 1.996us & 0 B & 0 B & 0 B & 0 B & 102 \\
\bottomrule
\end{tabular}
\end{adjustbox}
\caption{Profiler summary for MACE Interaction blocks.}
\label{tab:mace-interaction}
\end{table}

\begin{table}[h]
\centering
\scriptsize
\setlength{\tabcolsep}{3pt}
\renewcommand{\arraystretch}{1.1}
\begin{adjustbox}{max width=\textwidth}
\begin{tabular}{lrrrrrrrrrrrrrr}
\toprule
\textbf{Name} & \textbf{Self CPU \%} & \textbf{Self CPU} & \textbf{CPU total \%} & \textbf{CPU total} & \textbf{CPU time avg} & \textbf{Self CUDA} & \textbf{Self CUDA \%} & \textbf{CUDA total} & \textbf{CUDA time avg} & \textbf{CPU Mem} & \textbf{Self CPU Mem} & \textbf{CUDA Mem} & \textbf{Self CUDA Mem} & \textbf{\# of Calls} \\
\midrule
MACE/Product[0]/SymmetricContractions & 0.00\% & 0.000us & 0.00\% & 0.000us & 0.000us & 228.542ms & 15.63\% & 228.542ms & 2.241ms & 0 B & 0 B & 0 B & 0 B & 102 \\
MACE/Product[1]/Main & 0.00\% & 0.000us & 0.00\% & 0.000us & 0.000us & 159.399ms & 10.90\% & 159.399ms & 1.563ms & 0 B & 0 B & 0 B & 0 B & 102 \\
MACE/Product[1]/SymmetricContractions & 0.00\% & 0.000us & 0.00\% & 0.000us & 0.000us & 60.079ms & 4.11\% & 60.079ms & 589.013us & 0 B & 0 B & 0 B & 0 B & 102 \\
MACE/Product[0]/Linear\_Product & 0.00\% & 0.000us & 0.00\% & 0.000us & 0.000us & 45.857ms & 3.14\% & 45.857ms & 449.583us & 0 B & 0 B & 0 B & 0 B & 102 \\
MACE/Product[0]/Main & 0.19\% & 8.527ms & 9.26\% & 424.326ms & 4.160ms & 0.000us & 0.00\% & 35.361ms & 346.681us & 0 B & 0 B & 35.41 MB & -6.42 MB & 102 \\
MACE/Product[0]/SymmetricContractions & 1.74\% & 79.737ms & 7.43\% & 340.484ms & 3.338ms & 0.000us & 0.00\% & 21.563ms & 211.405us & 0 B & 0 B & 35.26 MB & -28.89 MB & 102 \\
MACE/Product[1]/Linear\_Product & 0.00\% & 0.000us & 0.00\% & 0.000us & 0.000us & 18.653ms & 1.28\% & 18.653ms & 182.868us & 0 B & 0 B & 0 B & 0 B & 102 \\
MACE/Product[0]/Linear\_Product & 0.50\% & 23.066ms & 1.33\% & 61.034ms & 598.376us & 0.000us & 0.00\% & 12.603ms & 123.564us & 0 B & 0 B & 6.47 MB & -12.75 MB & 102 \\
MACE/Product[0]/Main & 0.00\% & 0.000us & 0.00\% & 0.000us & 0.000us & 11.438ms & 0.78\% & 11.438ms & 112.136us & 0 B & 0 B & 0 B & 0 B & 102 \\
MACE/Product[1]/Main & 0.17\% & 8.013ms & 3.57\% & 163.543ms & 1.603ms & 0.000us & 0.00\% & 11.045ms & 108.286us & 0 B & 0 B & 14.64 MB & -3.24 MB & 102 \\
MACE/Product[1]/SymmetricContractions & 0.44\% & 20.185ms & 2.34\% & 107.158ms & 1.051ms & 0.000us & 0.00\% & 7.602ms & 74.528us & 0 B & 0 B & 14.54 MB & -12.95 MB & 102 \\
MACE/Product[1]/Linear\_Product & 0.34\% & 15.409ms & 0.77\% & 35.333ms & 346.402us & 0.000us & 0.00\% & 2.024ms & 19.848us & 0 B & 0 B & 1.64 MB & -3.19 MB & 102 \\
\bottomrule
\end{tabular}
\end{adjustbox}
\caption{Profiler summary for MACE Product blocks.}
\label{tab:mace-product}
\end{table}

\begin{table}[h]
\centering
\scriptsize
\setlength{\tabcolsep}{3pt}
\renewcommand{\arraystretch}{1.1}
\begin{adjustbox}{max width=\textwidth}
\begin{tabular}{lrrrrrrrrrrrrrr}
\toprule
\textbf{Name} & \textbf{Self CPU \%} & \textbf{Self CPU} & \textbf{CPU total \%} & \textbf{CPU total} & \textbf{CPU time avg} & \textbf{Self CUDA} & \textbf{Self CUDA \%} & \textbf{CUDA total} & \textbf{CUDA time avg} & \textbf{CPU Mem} & \textbf{Self CPU Mem} & \textbf{CUDA Mem} & \textbf{Self CUDA Mem} & \textbf{\# of Calls} \\
\midrule
MACE/Readout[0]/Linear\_readout & 0.00\% & 0.000us & 0.00\% & 0.000us & 0.000us & 18.588ms & 1.27\% & 18.588ms & 182.239us & 0 B & 0 B & 0 B & 0 B & 102 \\
MACE/Readout[1]/Linear1         & 0.00\% & 0.000us & 0.00\% & 0.000us & 0.000us & 17.966ms & 1.23\% & 17.966ms & 176.140us & 0 B & 0 B & 0 B & 0 B & 102 \\
MACE/Readout[1]/Linear2         & 0.00\% & 0.000us & 0.00\% & 0.000us & 0.000us & 17.716ms & 1.21\% & 17.716ms & 173.685us & 0 B & 0 B & 0 B & 0 B & 102 \\
MACE/Readout[1]/Main            & 0.12\% & 5.452ms & 1.75\% & 80.119ms & 785.482us & 0.000us & 0.00\% & 2.854ms & 27.981us & 816 B & 0 B & 357.00 KB & -204.00 KB & 102 \\
MACE/Readout[1]/NonLinearity    & 0.00\% & 0.000us & 0.00\% & 0.000us & 0.000us & 2.496ms & 0.17\% & 2.496ms & 24.466us & 0 B & 0 B & 0 B & 0 B & 102 \\
MACE/Readout[1]/Linear1         & 0.31\% & 14.291ms & 0.71\% & 32.684ms & 320.432us & 0.000us & 0.00\% & 1.407ms & 13.792us & 0 B & 0 B & 255.00 KB & -408.00 KB & 102 \\
MACE/Readout[0]/Main            & 0.05\% & 2.467ms & 0.81\% & 37.169ms & 364.404us & 0.000us & 0.00\% & 837.461us & 8.210us & 0 B & 0 B & 102.00 KB & 0 B & 102 \\
MACE/Readout[0]/Linear\_readout & 0.32\% & 14.844ms & 0.76\% & 34.702ms & 340.214us & 0.000us & 0.00\% & 837.461us & 8.210us & 0 B & 0 B & 102.00 KB & -102.00 KB & 102 \\
MACE/Readout[1]/Linear2         & 0.31\% & 14.338ms & 0.71\% & 32.605ms & 319.660us & 0.000us & 0.00\% & 755.986us & 7.412us & 0 B & 0 B & 102.00 KB & -102.00 KB & 102 \\
MACE/Readout[1]/NonLinearity    & 0.10\% & 4.665ms & 0.20\% & 9.378ms & 91.942us & 0.000us & 0.00\% & 691.330us & 6.778us & 816 B & 816 B & 204.00 KB & -204.00 KB & 102 \\
\bottomrule
\end{tabular}
\end{adjustbox}
\caption{Profiler summary for MACE Readout blocks.}
\label{tab:mace-readout}
\end{table}

\end{appendices}

\printthesisindex 

\end{document}